\documentclass[twoside,11pt]{article}

\usepackage[abbrvbib, preprint]{jmlr2e}

\usepackage{jmlr2e}
\usepackage{xcolor}
\usepackage{amsmath}
\usepackage{amsfonts}
\usepackage{bm}
\usepackage{times}
\usepackage{latexsym}
\usepackage{xcolor}
\usepackage{tabularx}
\usepackage{booktabs}
\usepackage{caption}
\usepackage{longtable}
\usepackage{multirow}
\usepackage{graphicx}
\usepackage[capitalize]{cleveref}
\usepackage{comment}

\ShortHeadings{Graph Neural Networks for NLP: A Survey}{Lingfei Wu and Yu Chen, et al.}
\firstpageno{1}

\begin{document}

\title{Graph Neural Networks for Natural Language Processing: \\ A Survey}

\author{\name Lingfei Wu\thanks{Both authors contributed equally to this research.} \email lwu@email.wm.edu\\
        \addr Pinterest, USA
        \AND
        \name Yu Chen\footnotemark[1] \email hugochan2013@gmail.com\\ 
        \addr Rensselaer Polytechnic Institute, USA 
        \AND
        \name Kai Shen\thanks{This research is done when Kai Shen is an intern at JD.COM.} \email shenkai@zju.edu.cn\\
        \addr Zhejiang University, China
        \AND
        \name Xiaojie Guo \email xiaojie.guo@jd.com\\
        \addr IBM T.J. Watson Research Center, USA
        \AND
        \name Hanning Gao \email ghnqwerty@gmail.com\\
        \addr Central China Normal University, China
        \AND
        \name Shucheng Li\thanks{Shucheng Li is also with National Key Lab for Novel Software Technology, Nanjing University.} \email shuchengli@smail.nju.edu.cn\\
        \addr Nanjing University, China
        \AND
        \name Jian Pei \email jpei@cs.sfu.ca\\
        \addr Simon Fraser University, Canada
        \AND
        \name Bo Long \email bo.long@jd.com\\
        \addr JD.COM, China
}

\editor{***}

\maketitle

\begin{abstract}
Deep learning has become the dominant approach in coping with various tasks in Natural Language Processing (NLP). Although text inputs are typically represented as a sequence of tokens, there is a rich variety of NLP problems that can be best expressed with a graph structure. As a result, there is a surge of interests in developing new deep learning techniques on graphs for a large number of NLP tasks. 
In this survey, we present a comprehensive overview on \emph{Graph Neural Networks (GNNs) for Natural Language Processing}. We propose a new taxonomy of GNNs for NLP, which systematically organizes existing research of GNNs for NLP along three axes: graph construction, graph representation learning, and graph based encoder-decoder models. We further introduce a large number of NLP applications that are exploiting the power of GNNs and summarize the corresponding benchmark datasets, evaluation metrics, and open-source codes. Finally, we discuss various outstanding challenges for making the full use of GNNs for NLP as well as future research directions. To the best of our knowledge, this is the first comprehensive overview of Graph Neural Networks for Natural Language Processing.

\end{abstract}


\maketitle
\section{Introduction}


Deep learning has become the dominant approach in coping with various tasks in Natural Language Processing (NLP) today, especially when operated on large-scale text corpora. Conventionally, text sequences are considered as a bag of tokens such as BoW and TF-IDF in NLP tasks. With recent success of Word Embeddings techniques \citep{DBLP:conf/nips/MikolovSCCD13,pennington2014glove}, sentences are typically represented as a sequence of tokens in NLP tasks. Hence, popular deep learning techniques such as recurrent neural networks \citep{650093} and convolutional neural networks \citep{NIPS2012_c399862d} have been widely applied for modeling text sequence.

However, there is a rich variety of NLP problems that can be best expressed with a graph structure. For instance, the sentence structural information in text sequence (i.e. syntactic parsing trees like dependency and constituency parsing trees) can be exploited to augment original sequence data by incorporating the task-specific knowledge. Similarly, the semantic information in sequence data (i.e. semantic parsing graphs like Abstract Meaning Representation graphs and Information Extraction graphs) can be leveraged to enhance original sequence data as well. Therefore, these graph-structured data can encode complicated pairwise relationships between entity tokens for learning more informative representations. 

Unfortunately, deep learning techniques that were disruptive for Euclidean data (e.g, images) or sequence data (e.g, text) are not immediately applicable to graph-structured data, due to the complexity of graph data such as irregular structure and varying size of node neighbors. As a result, this gap has driven a tide in research for deep learning on graphs, especially in development of graph neural networks (GNNs) \citep{GNNBook2022,kipf2016semi, defferrard2016convolutional, hamilton2017inductive}.

\begin{figure}[h]
\centering
\includegraphics[width=\textwidth]{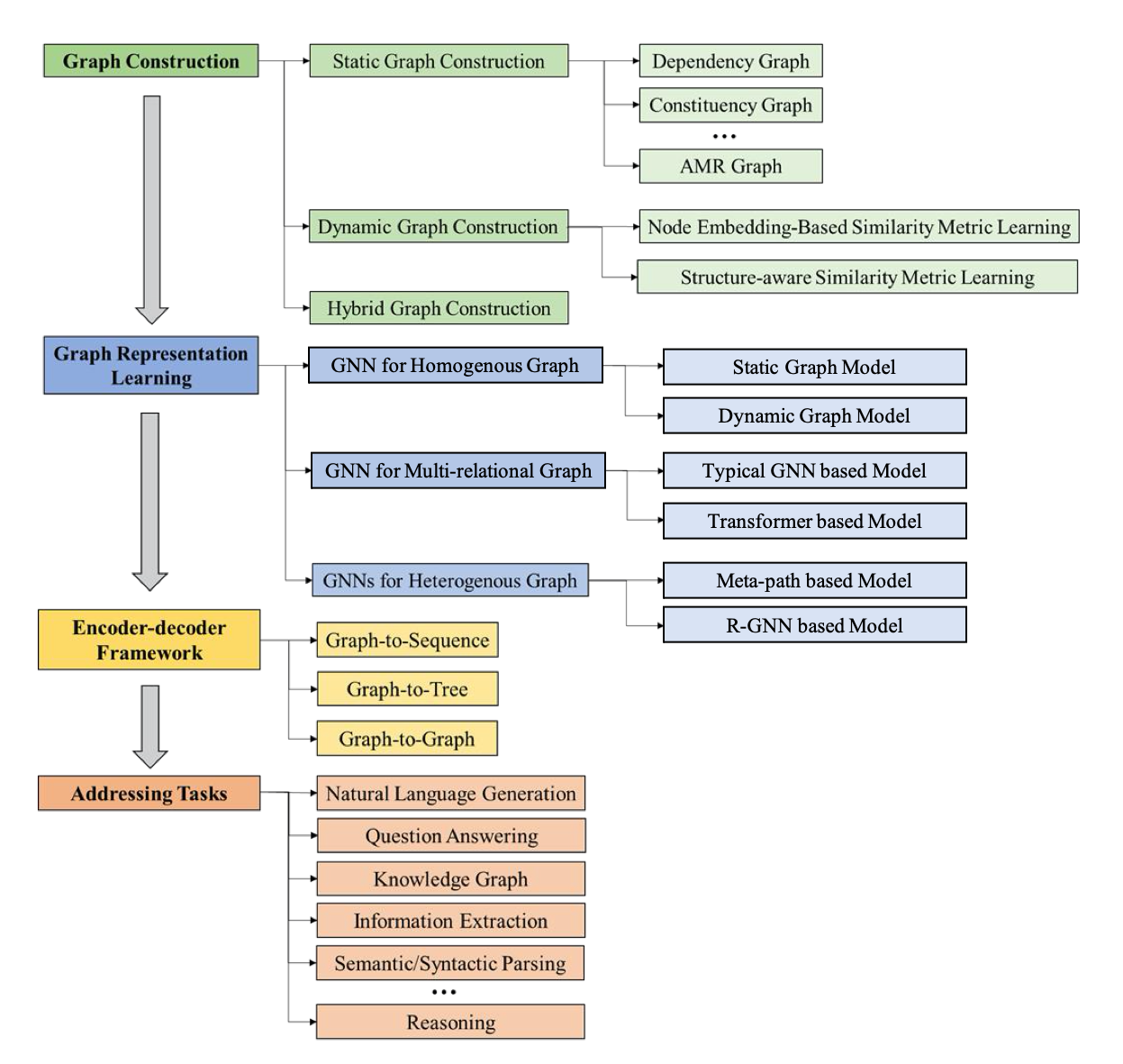}
\caption{The taxonomy, which systematically organizes GNNs for NLP along four axes: graph construction, graph representation learning, encoder-decoder models, and the applications.}
\label{fig:taxonomy}
\end{figure}

This wave of research at the intersection of deep learning on graphs and NLP has influenced a variety of NLP tasks~\citep{GNNBook-ch21-liu}. There has seen a surge of interests in applying and developing different GNNs variants and achieved considerable success in many NLP tasks, ranging from classification tasks like sentence classification \citep{henaff2015deep,huang_syntax-aware_2019}, semantic role labeling \citep{luo-zhao-2020-bipartite,gui-etal-2019-lexicon}, and relation extraction \citep{qu2020few,sahu-etal-2019-inter}, to generation tasks like machine translation \citep{bastings-etal-2017-graph,beck-etal-2018-graph}, question generation \citep{pan2020semantic,sachan2020stronger}, and summarization \citep{fernandes2018structured,yasunaga-etal-2017-graph}. Despite the successes these existing research has achieved, deep learning on graphs for NLP still encounters many challenges, namely,

\begin{itemize}
\item Automatically transforming original text sequence data into highly graph-structured data. Such challenge is profound in NLP since most of the NLP tasks involving using the text sequences as the original inputs. Automatic graph construction from the text sequence to utilize the underlying structural information is a crucial step in utilizing graph neural networks for NLP problems.\vspace{-0.1cm}
\item Properly determining graph representation learning techniques. It is critical to come up with specially-designed GNNs to learn the unique characteristics of different graph-structures data such as undirected, directed, multi-relational and heterogeneous graphs. \vspace{-0.1cm}
\item Effectively modeling complex data. Such challenge is important since many NLP tasks involve learning the mapping between the graph-based inputs and other highly structured output data such as sequences, trees, as well as graph data with multi-types in both nodes and edges. \vspace{-0.1cm}
\end{itemize}

In this survey, we will present for the first time a comprehensive overview of \textit{Graph Neural Networks for Natural Language Processing}. Our survey is timely for both Machine Learning and NLP communities, which covers relevant and interesting topics, including automatic graph construction for NLP, graph representation learning for NLP, various advanced GNNs-based encoder-decoder models (i.e. graph2seq, graph2tree, and graph2graph) for NLP, and the applications of GNNs in various NLP tasks. 
We highlight our main contributions as follows:
\begin{itemize}
    \item We propose a new taxonomy of GNNs for NLP, which systematically organizes existing research of GNNs for NLP along four axes: graph construction, graph representation learning, and graph based encoder-decoder models. \vspace{-0.1cm}
    \item We present the most comprehensive overview of the state-of-the-art GNNs-based approaches for various NLP tasks. We provide detailed descriptions and necessary comparisons on various graph construction approaches based on the domain knowledge and semantic space, graph representation learning approaches for various categories of graph-structures data, GNNs-based encoder-decoder models given different combinations of inputs and output data types. \vspace{-0.1cm}
    \item We introduce a large number of NLP applications that are exploiting the power of GNNs, including how they handle these NLP tasks along three key components (i.e., graph construction, graph representation learning, and embedding initialization), as well as providing corresponding benchmark datasets, evaluation metrics, and open-source codes. \vspace{-0.1cm}
    \item We outline various outstanding challenges for making the full use of GNNs for NLP and provides discussions and suggestions for fruitful and unexplored research directions. 
\end{itemize}

The rest of the survey is structured as follows. 
Section \ref{sec:Graph Based Algorithms for Natural Language Processing} reviews the NLP problems from a graph perspective, and then briefly introduces some representative traditional graph-based methods for solving NLP problems. 
Section \ref{sec:Graph Neural Networks} elaborates basic foundations and methodologies for graph neural networks, which are a class of modern neural networks that directly operate on graph-structured data. We also provide a list of notations used throughout this survey. 
Section \ref{sec:Graph Construction Methods for Natural Language Processing} focuses on introducing two major graph construction approaches, namely static graph construction and dynamic graph construction for constructing graph structured inputs in various NLP tasks. 
Section \ref{sec:Graph Representation Learning for NLP} discusses various graph representation learning techniques that are directly operated on the constructed graphs for various NLP tasks.
Section \ref{sec:GNN Based Encoder-Decoder Models} first introduces the typical Seq2Seq models, and then discusses two typical graph-based encoder-decoder models for NLP tasks (i.e., graph-to-tree and graph-to-graph models).
Section \ref{sec:Applications} discusses 12 typical NLP applications using GNNs bu providing the summary of all the applications with their sub-tasks, evaluation metrics and open-source codes.
Section \ref{sec:General Challenges and Future Directions} discusses various general challenges of GNNs for NLP and pinpoints the future research directions.
Finally, Section \ref{sec:Conclusions} summarizes the paper. The taxonomy, which systematically organizes GNN for NLP approaches along four axes: graph construction, graph representation learning, encoder-decoder models, and the applications is illustrated in Fig.\ref{fig:taxonomy}.



\section{Graph Based Algorithms for NLP}
\label{sec:Graph Based Algorithms for Natural Language Processing}

In this section, we will first review the NLP problems from a graph perspective, and then briefly introduce some representative traditional graph-based methods for solving NLP problems.


\subsection{Natural Language Processing: A Graph Perspective}



The way we represent natural language reflects our particular perspective on it, and has a fundamental influence on the way we process and understand it.
In general, there are three different ways of representing natural language.
The most simplified way is to represent natural language as a bag of tokens. This view of natural language completely ignores the specific positions of tokens appearing in text, and only considers how many times a unique token appears in text. If one randomly shuffles a given text, the meaning of the text does not change at all from this perspective.
The most representative NLP technique which takes this view is topic modeling~\citep{blei2003latent} which aims to model each input text as a mixture of topics where each topic can be further modeled as a mixture of words. 

A more natural way is to represent natural language as a sequence of tokens. This is how human beings normally speak and write natural language. Compared to the above bag perspective, this view of natural language is able to capture richer information of text, e.g., which two tokens are consecutive and how many times a word pair co-occurs in local context.
The most representative NLP techniques which take this view include the linear-chain CRF~\citep{DBLP:conf/icml/LaffertyMP01} which implements sequential dependencies in the predictions, and the word2vec~\citep{DBLP:conf/nips/MikolovSCCD13} which learns word embeddings by predicting the context words of a target word.

The third way is to represent natural language as a graph. Graphs are ubiquitous in NLP. While it is probably most apparent to regard text as sequential data, in the NLP community, there is a long history of representing text as various kinds of graphs. Common graph representations of text or world knowledge include dependency graphs, constituency graphs, AMR graphs, IE graphs, lexical networks, and knowledge graphs. Besides, one can also construct a text graph containing multiple hierarchies of elements such as document, passage, sentence and word. In comparison with the above two perspectives, this view of natural language is able to capture richer relationships among text elements. As we will introduce in next section, many traditional graph-based methods (e.g., random walk, label propagation) have been successfully applied to challenging NLP problems including word-sense disambiguation, name disambiguation, co-reference resolution, sentiment analysis, and text clustering.

\subsection{Graph Based Methods for Natural Language Processing}


In this previous subsection, we have discussed that many NLP problems can be naturally translated into graph-based problems. In this subsection, we will introduce various classical graph-based algorithms that have been successfully applied to NLP applications. Specifically, we will first briefly illustrate some representative graph-based algorithms and their applications in the NLP field. And then we further discuss their connections to GNNs.
For a comprehensive coverage of traditional graph-based algorithms for NLP, we refer the readers to \citep{mihalcea2011graph}.
    

\subsubsection{Random Walk Algorithms}
\paragraph{Approach}
Random walk is a class of graph-based algorithms that produce random paths in a graph. In order to do a random walk, one can start at any node in a graph, and repeatedly choose to visit a random neighboring node at each time based on certain transition probabilities. All the visited nodes in a random walk then form a random path. After a random walk converges, one can obtain a \emph{stationary distribution} over all the nodes in a graph, which can be used to either select the most salient node in a graph with high structural importance by ranking the probability scores or measure the relatedness of two graphs by computing the similarity between two random walk distributions.

\paragraph{Applications}
Random walk algorithms have been applied in various NLP applications including 
measures of semantic similarity of texts~\citep{ramage2009random} and semantic distance on semantic networks~\citep{hughes2007lexical}, word-sense disambiguation~\citep{mihalcea2005unsupervised,tarau2005semantic}, name disambiguation~\citep{minkov2006contextual}, query expansion~\citep{collins2005query}, keyword extraction~\citep{mihalcea2004textrank}, and cross-language information retrieval~\citep{monz2005iterative}.
For example, given a semantic network and a word pair, \citet{hughes2007lexical} computed a word-specific stationary distribution using a random walk algorithm, and measured the distance between two words as the similarity between the random walk distributions on this graph, biased on each input word in a given word pair.
To solve a name disambiguation task on email data, \citet{minkov2006contextual} built a graph of email-specific items (e.g., sender, receiver and subject) from a corpus of emails, and proposed a ``lazy'' topic-sensitive random walk algorithm which introduces a probability that the random walk would stop at a given node. 
Given an email graph and an ambiguous name appearing in an input email, a random walk is performed biased toward the text of the given email, and the name is resolved to the correct reference by choosing the person node that has the highest score in the stationary distribution after convergence.
To solve the keyword extraction task, \citet{mihalcea2004textrank} proposed to perform a random walk on a co-occurrence graph of words, and rank the importance of the words in the text based on their probability scores in the stationary distribution.

\subsubsection{Graph Clustering Algorithms}

\paragraph{Approach}
Common graph clustering algorithms include spectral clustering, random walk clustering and min-cut clustering.
Spectral clustering algorithms make use of the spectrum (eigenvalues) of the Laplacian matrix of the graph to perform dimensionality reduction before conducting clustering using existing algorithms like K-means.
Random walk clustering algorithms operate by conducting a $t$-step random walk on the graph, as a result, each node is represented as a probability vector indicating the $t$-step generation probabilities to all of the other nodes in the graph. Any clustering algorithm can be applied on the generation-link vectors.
Note that for graph clustering purposes, a small value of $t$ is more preferred because we are more interested in capturing the local structural information instead of the global structural information (encoded by the stationary distribution after a random walk converges).
The min-cut algorithms can also be used to partition the graph into clusters.

\paragraph{Applications}
Graph clustering algorithms have been successfully applied to solve the text clustering task. For instance, \citet{erkan2006language} proposed to use the $n$-dim probabilistic distribution derived from a $t$-step random walk on a directed generation graph (containing $n$ document nodes) as the vector representation of each document in a corpus. Then these document representations can be consumed by a graph clustering algorithm to generate document clusters.
Note that the generation graph is constructed by computing the generation probability of each ordered document pair in the corpus following the language-model approach proposed by~\citet{ponte1998language}.



\subsubsection{Graph Matching Algorithms}

\paragraph{Approach}
Graph matching algorithms aim to compute the similarity between two graphs. Among them, Graph Edit Distance is the most commonly used method to measure the dissimilarity of two graphs. It computes the distance as the number of changes (i.e., add, delete, substitute) needed to transform one graph into the other. Then the dissimilarity score can be converted into the similarity score.



\paragraph{Applications}
Graph matching algorithms have applications in the textual entailment task that aims at deciding whether a given sentence can be inferred from text. For example, \citet{haghighi2005robust} assumed that a hypothesis is entailed from the text when the cost of matching the hypothesis graph to the text graph is low, and thus applied a graph matching algorithm to solve the problem.







\subsubsection{Label Propagation Algorithms}

\paragraph{Approach}
Label propagation algorithms (LPAs) is a class of semi-supervised graph-based algorithms that propagate labels from labeled data points to previously unlabeled data points.
Basically, LPAs operate by propagating and aggregating labels iteratively across the graph. At each iteration, each node changes its label based on the labels that its neighboring nodes possess. As a result, the label information diffuses through the graph.

\paragraph{Applications}
LPA have been widely used in the network science literature for discovering community structures in complex networks.
In the literature of NLP, LPA have been successfully applied in word-sense disambiguation~\citep{niu2005word} and sentiment analysis~\citep{goldberg2006seeing}.
These applications usually focus on the semi-supervised learning setting where labeled data is scarce, and leverage the LPA algorithm for propagating labels from limited labeled examples to a large amount of similar unlabeled examples with the assumption that similar examples should have similar labels.

\subsubsection{Limitations and Connections to GNNs}

Although traditional graph-based algorithms have been successfully applied in many NLP tasks, they have several limitations.
First of all, they have limited expressive power. They mostly focus on capturing the structural information of graphs but do not consider the node and edge features which are also very important for many NLP applications.
Secondly, there is not a unified learning framework for traditional graph-based algorithms. Different graph-based algorithms have very different properties and settings, and are only suitable to some specific use cases.

The above limitations of traditional graph-based algorithms call for a unified graph-based learning framework with strong expressive power on modeling both the graph structures and node/edge properties. Recently, GNNs have gained increasing attention as a special class of neural networks which can model arbitrary graph-structured data. 
Most GNN variants can be regarded as a message passing based learning framework. Unlike traditional message passing based algorithms like LPA which operates by propagating labels across a graph, GNNs typically operate by transforming, propagating and aggregating nodes/edge features through several neural layers so as to learn better graph representations. As a general graph-based learning framework, GNNs can be applied to various graph-related tasks such as node classification, link prediction and graph classification.

\section{Graph Neural Networks}
\label{sec:Graph Neural Networks}

In the previous chapter, we have illustrated various conventional graph-based methods for different NLP applications. In this chapter, we will elaborate basic foundations and methodologies for graph neural networks (GNNs) which are a class of modern neural networks which directly operate on graph-structured data~\citep{GNNBook2022}. To facilitate the description of the technologies, we list all the notations used throughout this survey in~\cref{table:notation}, which includes variables and operations in the domain of both the graph neural networks and NLP.



\begin{center}
\begin{longtable}{ p{.80\textwidth} | p{.20\textwidth} } 
\caption{Notation}
\label{table:notation}
\endfirsthead
\endhead







\\
\textbf{Graph Basics} &\\
\\

A graph                               & $\mathcal{G}$\\
Edge set                       & $\mathcal{E}$\\
Vertex (node) set                      & $\mathcal{V}$\\
The number of vertexes (nodes)   &   $n$ \\
The number of edges & $m$ \\
A single vertex(node) $v_i \in \mathcal{V}$   &   $v_i$ \\
A single edge $e_{i,j}$(connecting vertex $v_i$ and vertex $v_j\in \mathcal{E}$ &   $e_{i,j}$ \\
The neighbours of a vertex (node) $v_i$  &  $N(v_i)$ \\
Adjacent matrix of a graph      & $A$\\
Laplacian matrix             & $L$\\
Diagonal degree matrix          & $D$\\
The initial attributes of vertex $v_i \in \mathcal{V}$   &   $\mathbf{x}_{i}$    \\
The initial attributes of edge $e_{i, j} \in \mathcal{E}$ & $ \mathbf{r}_{i, j}$   \\

The embedding of vertex $v_i \in \mathcal{V}$   &   $\mathbf{h}_{i}$    \\
The embedding of edge $e_{i, j} \in \mathcal{E}$ & $ \mathbf{e}_{i, j}$   \\


\\
\textbf{NLP Basics} &\\
\\
Vocabulary  &   $V$ \\
Source language & s \\
Target language & t \\
Corpus of words/aligned sentences used for training   & C\\ 
The $i^{th}$ word in corpus C   &   $w_{i}$ \\
The embedding of word $w_i$ &   $\mathbf{w}_{i}$\\
The embedding vector's dimensionality   & d \\
The number of words &   n   \\

The $i^{th}$ document in source(target) language    & $doc^{s}_{i} (doc^{t}_{i})$ \\
Representation of document $doc^{s}_{i}(doc^{t}_{i})$  & $\mathbf{d}^{s}_{i}(\mathbf{d}^{t}_{i})$\\
The $i^{th}$ paragraph in source(target) language   & $para^{s}_{i} (para^{t}_i)$ \\
Representation of paragraph $para^{s}_{i} (para^{t}_{i})$    &   $\mathbf{p}^{s}_{i} (\mathbf{p} ^ {t} _ {i})$  \\
The $i^{th}$ sentence in source(target) language   & $sent^{s}_{i} (sent^{t}_{i})$ \\
Representation of sentence $sent^{s}_{i}(sent^{t}_i)$ & $\mathbf{s}^{s}_{i}(\mathbf{s}^{t}_i)$  \\
\\

\end{longtable}
\end{center}




\subsection{Foundations}
Graph neural networks are essentially graph representation learning models and can be applied to node-focused tasks and graph-focused tasks. GNNs learn embeddings for each node in the graph and aggregate the node embeddings to produce the graph embeddings. Generally, the learning process of node embeddings utilizes graph structure and input node embeddings, which can be summarized as:
\begin{equation}
\label{eq-filter}
 \mathbf{h}^{(l)}_i = f_\mathbf{filter}(A, \mathbf{H}^{(l-1)})
\end{equation}
where $A \in \mathbb{R}^{n \times n}$ is the adjacency matrix of the graph, $\mathbf{H}^{(l-1)} = \{\mathbf{h}_1^{(l-1)},  \mathbf{h}_2^{(l-1)},$ $ ..., \mathbf{h}_n^{(l-1)}\} \in \mathbb{R}^{n \times d}$ denotes the input node embeddings at the $l-1$-th GNN layer, and $\mathbf{H}^{(l)}$ is the updated node embeddings. $d$ is the dimension of $\mathbf{h}_i^{(l-1)}$. 
We refer to the process depicted in Eq.\eqref{eq-filter} as \textit{graph filtering} and $f_\mathbf{filter}(\cdot, \cdot)$ is named as a graph filter. The specific models then differ only in how $f_\mathbf{filter}(\cdot, \cdot)$ is chosen and parameterized. Graph filtering does not change the structure of graph, but refines the node embeddings. Graph filtering layers are stacked to $L$ layers to generate final node embeddings.

Since graph filtering does not change the graph structure, pooling operations are introduced to aggregate node embeddings to generate graph-level embeddings inspired by CNNs. In GNN models, the \textit{graph pooling} takes a graph and its node embeddings as inputs and then generates a smaller graph with fewer nodes and its corresponding new node embeddings. The graph pooling operation can be summarized as follows:
\begin{equation}
    A', \mathbf{H}' = f_\mathbf{pool}(A, \mathbf{H})
    \label{eq:pool}
\end{equation}
where $f_\mathbf{pool}(\cdot, \cdot)$   $A \in \mathbb{R}^{n \times n}$ and $A' \in \mathbb{R}^{n' \times n'}$ are the adjacency matrices before and after graph pooling. $\mathbf{H} \in \mathbb{R}^{n \times d}$ and $\mathbf{H}' \in \mathbb{R}^{n' \times d'}$ are the node embeddings before and after graph pooling. $n'$ is set to be 1 in most cases to get the embedding for the entire graph.

\subsection{Methodologies}

\subsubsection{Graph Filtering}
There exists a variety of implementations of graph filter $f$ in Eq.\eqref{eq-filter}, which could be roughly categorized into spectral-based graph filters, spatial-based graph filters, attention-based graph filters and recurrent-based graph filters. Conceptually, the spectral-based graph filters are based on spectral graph theory while the spatial-based methods compute a node embedding using its spatially close neighbor nodes on the graph. Some spectral-based graph filters can be converted to spatial-based graph filters. The attention-based graph filters are inspired by the self-attention mechanism \citep{vaswani2017attention} to assign different attention weights to different neighbor nodes. Recurrent-based graph filters introduce gating mechanism, and the model parameters are shared across different GNN layers. Next, we will explain these four types of graph filters in detail by introducing some of their representative GNN models. 

\paragraph{Spectral-based Graph Filters}
Inspired by graph signal processing, \citeauthor{defferrard2016convolutional} proposed a spectral graph theoretical formulation of CNNs, which generalizes CNNs to graphs and provides the same linear computational complexity and constant learning complexity as classical CNNs. A more typical example of spectral-based graph filters is Graph Convolutional Networks (GCN)~\citep{kipf2016semi}. Spectral convolution on graphs is defined as the multiplication of a signal $\mathbf{x}_i \in \mathbb{R}^n$ (a scalar for node $v_i$) with the filter $f_\mathbf{filter} = \text{diag}(\theta)$ parameterized by $\theta \in \mathbb{R}^n$ in the Fourier domain:
\begin{equation}
\label{gcn-0}
    f_\mathbf{filter} \ast \mathbf{x}_i = \mathbf{U} f(\mathbf{\Lambda})\mathbf{U}^T\mathbf{x}_i
\end{equation}
where $\mathbf{U}$ is the matrix of eigenvectors of the normalized graph Laplacian ${L}={I}_n-{D}^{-\frac{1}{2}}{A}{D}^{-\frac{1}{2}}$. ${I}_n$ is the identity matrix, ${D}$ is the degree matrix and $\mathbf{\Lambda}$ is the eigenvalues of ${L}$.

However, the computation of the full eigen-decomposition is prohibitively expensive. To solve this problem,   \citet{defferrard2016convolutional} uses a truncated expansion in terms of Chebyshev polynomials $\mathbf{T}_p(x)$ up to $P^{th}$-order to approximate $\mathbf{g}_\theta(\mathbf{\Lambda})$. Eq. \eqref{gcn-0} can be represented as follows:
\begin{equation}
\label{gcn-1}
    f'_\mathbf{filter} \ast \mathbf{x}_i \approx \sum_{p=0}^P\theta_p' \mathbf{T}_p(\Tilde{\mathbf{L}})\mathbf{x}_i
\end{equation}
where $\Tilde{{L}}=\frac{2}{\lambda_{max}}{L}-{I}_n$. $\lambda_{max}$ is the largest eigenvalue of ${L}$. $\theta_k' \in \mathbb{R}^P$ is a vector of Chebyshev coefficients. The Chebyshev polynomials can be defined recursively: $\mathbf{T}_k(\mathbf{x}_i)=2\mathbf{x}_i\mathbf{T}_{k-1}(\mathbf{x}_i)-\mathbf{T}_{k-2}(\mathbf{x}_i)$, with $\mathbf{T}_0(\mathbf{x}_i)=1$ and $\mathbf{T}_1(\mathbf{x}_i)=\mathbf{x}_i$. Eq.\eqref{gcn-1} is a $K$th-order polynomial in the Laplacian, which shows that every central node depends only on nodes in the $P$-hop range.

Therefore, a neural network model based on graph convolution can stack multiple convolutional layers using Eq. \eqref{gcn-1}. By limiting the layer-wise convolution operation to $P = 1$ and stacking multiple convolutional layers, \citet{kipf2016semi} proposed a multi-layer Graph Convolutional Network (GCN). It further approximates $\lambda_{max} \approx 2$ and Eq. \eqref{gcn-1} is simplified to:
\begin{equation}
   f'_\mathbf{filter} \ast \mathbf{h}_i^{(l)} \approx \theta_0' \mathbf{h}_i^{(l)} + \theta_1' ({L}-{I}_n)\mathbf{h}_i^{(l)} = \theta_0' \mathbf{h}_i^{(l)} - \theta_1' {D}^{-\frac{1}{2}}{A}{D}^{-\frac{1}{2}} \mathbf{h}_i^{(l)}
\end{equation}
with two free parameters $\theta_0'$ and $\theta_1'$. To alleviate the problem of overfitting and minimize the number of operations (such as matrix multiplications), it is beneficial to constrain the number of parameters by setting a single parameter $\theta = \theta_0' = -\theta_1'$:
\begin{equation}
    f_\mathbf{filter} \ast\mathbf{h}_i^{(l)} \approx \theta({I}_n+{D}^{-\frac{1}{2}}{A}{D}^{-\frac{1}{2}})\mathbf{h}_i^{(l)}
\end{equation}
Repeat application of this operator may cause numerical instability and explosion/vanishing gradients, \citet{kipf2016semi} proposed to use a \textit{renormalization trick}: ${I}_n+{D}^{-\frac{1}{2}}{A}{D}^{-\frac{1}{2}} \rightarrow {\Tilde{D}}^{-\frac{1}{2}}{\Tilde{A}}{\Tilde{D}}^{-\frac{1}{2}}$, with ${\Tilde{A}}={A}+{I}_n$ and ${\Tilde{D}}_{ii} = \sum_j {\Tilde{A}}_{ij}$. Finally, the definition can be generalized with a signal $\mathbf{H} \in \mathbb{R}^{n \times d}$ with $d$ input channels (i.e. a $d$-dimensional feature vector for each node) and $F$ filters or feature maps as follows:
\begin{equation}
    \mathbf{H}^{(l)}=\sigma( {\Tilde{D}}^{-\frac{1}{2}}{\Tilde{A}}{\Tilde{D}}^{-\frac{1}{2}} \mathbf{H}^{(l-1)} \mathbf{W}^{(l-1)})
\end{equation}
Here, $\mathbf{W}^{(l-1)}$ is a layer-specific trainable weight matrix and $\sigma(\cdot)$ denotes an activation function. $\mathbf{H}^{(l)} \in \mathbb{R}^{n \times d}$ is the activated node embeddings at $(l-1)$-th layer.

\paragraph{Spatial-based Graph Filters}
Analogous to the convolutional operation of a conventional
CNN, spatial-based graph filters operate the graph convolutions based on a node’s spatial relations. The spatial-based graph filters derive the updated representation for the target node via convolving its representation with its neighbors’ representations. On the other hand, spatial-based graph filters hold the idea of information propagation, namely, message passing. The spatial-based graph convolutional operation essentially propagates node information as messages along the edges. Here we introduce two typical GNNs based on spatial-based graph filters are Message Passing Neural Network (MPNN)~\citep{gilmer2017neural} and GraphSage~\citep{hamilton2017inductive}.

MPNN~\citep{gilmer2017neural} proposes a general framework of spatial-based graph filters $f_\mathbf{filter}$ which is a composite function consisting of $f_U$ and $f_M$. It treats graph convolutions as a message passing process in which
information can be passed from one node to another along
the edges directly. MPNN runs $K$-step message passing iterations
to let information propagate further to K-hop neighboring nodes. The message passing
function, namely the spatial-based graph filter, on the target node $v_i$ is defined as
\begin{equation}
    \mathbf{h}^{(l)}_{i}=f_\mathbf{filter}(A, \mathbf{H}^{(l-1)})=f_U(\mathbf{h}^{(l-1)}_i,\sum_{v_j\in N(v_i)} f_M(\mathbf{h}_i^{(l-1)},\mathbf{h}_j^{(l-1)},\mathbf{e}_{i,j})),
\end{equation}
where $\mathbf{h}^{(0)}_{i}=\mathbf{x}_i$, $f_U(\cdot)$ and $f_M(\cdot)$ are the update and message aggregate functions with learnable parameters, respectively. After deriving the hidden representations of each node, $\mathbf{h}^{(L)}_i$ ($L$ is the number of graph convolution layers) can be passed to an output layer to perform
node-level prediction tasks or to a readout function to perform graph-level prediction tasks. MPNN is very general to include many existing GNNs by applying different functions of $f_U(\cdot)$ and $f_M(\cdot)$.

Considering that the number of neighbors of a node can vary from one to a thousand or even more, it is inefficient to take the full size of a node’s neighborhood in a giant graph with thousands of millions of nodes. GraphSage~\citep{hamilton2017inductive} adopts sampling to obtain a fixed number of neighbors for each node as
\begin{equation}
    f_\mathbf{filter}(A, \mathbf{H}^{(l-1)})=\sigma(\mathbf{W}^{(l)}\cdot f_M(\mathbf{h}^{(l-1)}_i,\{\mathbf{h}^{(l-1)}_j, \forall v_j\in N(v_i)\})),
\end{equation}
where $N(v_i)$ is a random sample of the neighboring nodes of node $v_i$. The aggregation function can be any functions that are invariant to the permutations of node orderings such as mean, sum or max operations.


\paragraph{Attention-based Graph Filters}
The original versions of GNNs take edge connections of the input graph as fixed, and do not dynamically adjust the connectivity information during the graph learning process.
Motivated by the above observation, and inspired by the successful applications of multi-head attention mechanism in the Transformer model~\citep{vaswani2017attention, velivckovic2017graph} proposed the Graph Attention Network (GAT) by introducing the multi-head attention mechanism to the GNN architecture which is able to dynamically learn the weights (i.e., attention scores) on the edges when performing message passing. 
More specifically, when aggregating embeddings from neighboring nodes for each target node in the graph, the semantic similarity between the target node and each neighboring node will be considered by the multi-head attention mechanism, and important neighboring nodes will be assigned higher attention scores when performing the neighborhood aggregation. For the $l$-th layer, GAT thus uses the following formulation of the attention mechanism,
\begin{align}
\alpha_{ij} = \frac{\text{exp}(\text{LeakyReLU}({\vec{u}^{(l)T}} [\vec{W}^{(l)} \mathbf{h}_i^{(l-1)} || \vec{W}^{(l)} \mathbf{h}_j^{(l-1)}]))}{\sum_{v_k \in N(v_i)} \text{exp}(\text{LeakyReLU}({\vec{u}^{(l)T}} [\vec{W}^{(l)} \mathbf{h}_i^{(l-1)} || \vec{W}^{(l)} \mathbf{h}_k^{(l-1)}]))}
\end{align}
where $\vec{u}^{(l)}$ and $\vec{W}^{(l)}$ are the weight vector and weight matrix at $l$-th layer, respectively, and $||$ is the vector concatenation operation. Note that $N(v_i)$ is the 1-hop neighborhood of $v_i$ including itself.
After obtaining the attention scores $\alpha_{ij}$ for each pair of nodes $v_i$ and $v_j$, the updated node embeddings can be computed as a linear combination of the input node features followed by some nonlinearity $\sigma$, formulated as,
\begin{align}
\mathbf{h}_i^{(l)} =f_\mathbf{filter}(A, \mathbf{H}^{(l-1)})=\sigma(\sum_{v_j \in N(v_i)}\alpha_{ij}\vec{W}^{(l)}\mathbf{h}^{(l-1)}_j)
\end{align}

In order to stabilize the learning process of the above self-attention, inspired by \citet{vaswani2017attention}, multiple independent self-attention mechanisms are employed and their outputs are concatenated to produce the following node embedding:
\begin{align}
f_\mathbf{filter}(A, \mathbf{H}^{(l-1)})= ||_{k=1}^{K} \sigma(\sum_{v_j \in N(v_i)}\alpha_{ij}^k\vec{W}^{(l)}_k\mathbf{h}^{(l-1)}_j),
\end{align}
while the final GAT layer (i.e., the $L$-th layer for a GNNs with $L$ layers) employs averaging instead of concatenation to combine multi-head attention outputs.
\begin{align}
f_\mathbf{filter}(A, \mathbf{H}^{(L-1)})= \sigma(\frac{1}{K} \sum_{k=1}^K \sum_{v_j \in N(v_i)}\alpha_{ij}^k\vec{W}^{(L)}_k\mathbf{h}_j^{(L-1)})
\end{align}

\paragraph{Recurrent-based Graph Filters}
A typical example of recurrent-based graph filters is the Gated Graph Neural Networks (GGNN)-filter. The biggest modification from typical GNNs to GGNNs is the use of Gated Recurrent Units (GRU) \citep{DBLP:conf/emnlp/ChoMGBBSB14}. Analogous to RNN, GGNN unfolds the recurrence in a fixed T time steps and uses back propagation through time to calculate the gradients. The GGNN-filter also takes the edge type and edge direction into consideration. To this end, $e_{i,j}$ denotes the directed edge from node $v_i$ to node $v_j$ and the edge type of $e_{i,j}$ is $t_{i,j}$. The propagation process of recurrent-based filter  $f_\mathbf{filter}$ in GGNN can be summarized as follows:
\begin{align}
    \mathbf{h}_i^{(0)} &= [\mathbf{x}_i^T, \mathbf{0}]^T 
    \label{ggnn-0} \\
    \mathbf{a}_i^{(l)} &= A_{i:}^T[\mathbf{h}_1^{(l-1)}...\mathbf{h}_n^{(l-1)}]^T
    \label{eq:ggnn-aggregation}
    \\
    \mathbf{h}_i^{(l)} &= \text{GRU}(\mathbf{a}_i^{(l)}, \mathbf{h}_i^{(l-1)})
\end{align}
where $A \in \mathbb{R}^{{dn} \times 2dn}$ is a matrix determining how nodes in the graph communicating with each other. $n$ is the number of nodes in the graph. $A_{i:} \in \mathbb{R}^{d \times 2d}$ are the two columns of blocks in $A$ corresponding to node $v_i$. In Eq. \eqref{ggnn-0}, the initial node feature $\mathbf{x}_i$ are padded with extra zeros to make the input size equal to the hidden size. Eq. \eqref{eq:ggnn-aggregation} computes $\mathbf{a}_i^{(l)} \in \mathbb{R}^{2d}$ by aggregating information from different nodes via incoming and outgoing edges with parameters dependent on the edge type and direction. The following step uses a GRU unit to update the hidden state of  node $v$ by incorporating $\mathbf{a}_i^{(l)}$ and the previous timestep hidden state $\mathbf{h}_i^{(l-1)}$.


\subsubsection{Graph Pooling}
Graph pooling layers are proposed to generate graph-level representations for graph-focused downstream tasks, such as graph classification and prediction based on the node embedding learned from the graph filtering. This is because the learned node embeddings are sufficient for node-focused tasks, however, for graph-focused tasks, a representation of the entire graph is required. To this end, we need to summarize the node embeddings information and the graph structure information. The graph pooling layers can be classified into two categories: flat graph pooling and hierarchical graph pooling. The flat graph pooling generates the graph-level representation directly from the node embeddings in a single step. In contrast, the hierarchical graph pooling contains several graph pooling layers and each of the pooling layer follows a stack of graph filters. In this section, we briefly introduce several representative flat pooling layers and hierarchical pooling layers.

\paragraph{Flat Graph Pooling} Ideally, an aggregator function would be invariant to permutations of its input while maintaining a large expressive capacity. The graph pooling operation $f_\mathbf{pool}$ is commonly implemented as Max-pooling and Average-pooling. Another popular choices are the variants of the Max-pooling and Average pooling operations by following a fully-connected layer (FC) transformation. The resulting max pooling and FCmax can be expressed as:
\begin{equation}
    \mathbf{r}_i = \text{max}(\mathbf{H}_{:,i}) \ \ \textit{or } \ \ \mathbf{r}_i = \text{max}(\mathbf{W} \mathbf{H}_{:,i})
\end{equation}
where $i$ denotes the $i$-th channel of the node embedding and $\mathbf{H}_{:,i} \in \mathbf{R}^{n \times 1}$ is a vector. $\mathbf{W}$ is a matrix that denotes to the trainable parameters of the FCmax pooling layer. $\mathbf{r}_i$ is a scalar and the final graph embedding $\mathbf{R} = [r_1, r_2,...,r_n]^T$. 
Finally, a powerful but less common pooling operation is the BiLSTM aggregation
function which is not permutation invariant on the set of
node embeddings. However, it has been often demonstrated to have better expressive power than other flat pooling operations \citep{hamilton2017inductive,HGNN:zhang2019heterogeneour}.

\paragraph{Hierarchical Graph Pooling}  Hierarchical graph pooling coarsens the graph step by step to learn the graph-level embeddings. Hierarchical pooling layers can be divided into two categories according to the ways to coarsen the graph. One type of hierarchical pooling layer coarsens the graph by sub-sampling the most important nodes as the nodes of the coarsened graph \citep{gao2019learning}. Another type of hierarchical pooling layer combines nodes in the input graph to form supernodes, which serve as the nodes of the coarsened graph \citep{ying2018hierarchical, ma2019graph}. After sub-sampling nodes or generating supernodes, the hierarchical graph pooling $f_\mathbf{pool}$ can be summarized as: (1) generating graph structure for the coarsened graph; (2) generating node features for the coarsened graph. The graph structure for the coarsened graph is generated from the input graph:
\begin{equation}
    {A}' = COARSEN({A})
\end{equation}
where ${A} \in \mathbb{R}^{n \times n}$ is the adjacent matrix of the input graph, and ${A}' \in \mathbb{R}^{n' \times n'}$ is the adjacent matrix of the coarsened graph. $f(.)$ is the graph sub-sampling or supernodes generating function. 

\section{Graph Construction Methods for NLP}
\label{sec:Graph Construction Methods for Natural Language Processing}

In the previous section, we have discussed the basic foundations and methods of GNNs once given a graph input. Unfortunately, for most of the NLP tasks, the typical inputs are sequence of text rather than graphs. Therefore, how to construct a graph input from sequences of text becomes a demanding step in order to leverage the power of GNNs. 
In this chapter, we will focus on introducing two major graph construction approaches, namely static graph construction and dynamic graph construction for constructing graph structured inputs in various NLP tasks.



\begin{table}[tb]
\caption{Two major graph construction approaches: static  and dynamic graph constructions}
\resizebox{\textwidth}{!}{%
\begin{tabular}{|c|c|c|l|l|}
\hline
\textbf{Approaches} & \multicolumn{3}{c|}{\textbf{Techniques}} & \textbf{References} \\ \hline
\multirow{18}{*}{Static Graph} & \multicolumn{3}{c|}{\multirow{8}{*}{Dependency Graph}}  & \citet{zhang_aspect-based_2019,guo-etal-2019-attention,zhang-qian-2020-convolution,fei2020cross} \\
& \multicolumn{3}{c|}{} & \citet{bastings-etal-2017-graph,nguyen2018graph,ji-etal-2019-graph,liu-etal-2018-jointly} \\
& \multicolumn{3}{c|}{} & \citet{xu-etal-2018-exploiting,zhang-etal-2018-graph,song-etal-2018-n, li-etal-2017-context} \\
& \multicolumn{3}{c|}{} & \citet{do-rehbein-2020-neural,yan-etal-2019-event,marcheggiani-etal-2018-exploiting,zhou-etal-2020-amr} \\
& \multicolumn{3}{c|}{} & \citet{vashishth-etal-2018-reside,xia2020semantic,jin2020semsum,huang_syntax-aware_2019} \\ 
& \multicolumn{3}{c|}{} & \citet{sahu-etal-2019-inter, cui-etal-2020-edge,Xu2020DocumentGF, zhang2020syntax} \\ 
& \multicolumn{3}{c|}{} & \citet{liu2019learning, li-etal-2020-graph-tree, wang_relational_2020, tang-etal-2020-dependency} \\ 
& \multicolumn{3}{c|}{} & \citet{qian2019graphie, pouran_ben_veyseh_improving_2020, wang-etal-2020-answer} \\ \cline{2-5}
& \multicolumn{3}{c|}{Constituency Graph} & \citet{li-etal-2020-graph-tree,marcheggiani-titov-2020-graph,xu-etal-2018-exploiting}  \\ \cline{2-5} 
& \multicolumn{3}{c|}{\multirow{5}{*}{AMR Graph}}  & \citet{liao2018abstract,wang2020amr,ijcai2020-542,ribeiro2019enhancing} \\
& \multicolumn{3}{c|}{} & \citet{jin-gildea-2020-generalized,jin2020semsum, cai2020graph} \\
& \multicolumn{3}{c|}{} & \citet{bai-etal-2020-online, beck-etal-2018-graph, yao2020heterogeneous} \\
& \multicolumn{3}{c|}{} & \citet{zhang-etal-2020-lightweight,zhao-etal-2020-line,zhu-etal-2019-modeling} \\
& \multicolumn{3}{c|}{} & \citet{song-etal-2020-structural,song2018graph, song2019semantic, damonte-cohen-2019-structural} \\ \cline{2-5} 
& \multicolumn{3}{c|}{Information Extraction}  & \citet{wu2020extracting,vashishth-etal-2018-reside} \\ 
& \multicolumn{3}{c|}{Graph}  & \citet{huang-etal-2020-knowledge,gupta-etal-2019-care} \\ \cline{2-5} 
 & \multicolumn{3}{c|}{Discourse Graph} &   \citet{song-etal-2018-n,li2020leveraging,yasunaga-etal-2017-graph,xu2020discourse} \\ \cline{2-5} 
& \multicolumn{3}{c|}{\multirow{10}{*}{Knowledge Graph}} & \citet{ye2019vectorized,yang2019aligning,gupta-etal-2019-care,xu2020coordinated} \\
& \multicolumn{3}{c|}{} & \citet{sun2020knowledge,xu-etal-2019-cross-lingual,Wang_Kapanipathi_Musa_Yu_Talamadupula_Abdelaziz_Chang_Fokoue_Makni_Mattei_Witbrock_2019} \\
& \multicolumn{3}{c|}{} & \citet{Kapanipathi2020InfusingKI, zhang-etal-2019-long, zhang2020relational, sun-etal-2018-open} \\
& \multicolumn{3}{c|}{} & \citet{malaviya2020commonsense,huang-etal-2020-knowledge, schlichtkrull2018modeling, sun-etal-2019-pullnet} \\
& \multicolumn{3}{c|}{} & \citet{bansal2019a2n, saxena-etal-2020-improving, koncel-kedziorski-etal-2019-text} \\
& \multicolumn{3}{c|}{} & \citet{teru2020inductive,lin-etal-2019-kagnet,ghosal-etal-2020-kingdom,feng2020scalable} \\
& \multicolumn{3}{c|}{} & \citet{wu2019jointly, wu2019relation,wu2020extracting,wu-etal-2020-knowledge} \\
& \multicolumn{3}{c|}{} & \citet{wang-etal-2019-incorporating, wang2019logic,wang-etal-2020-knowledge-graph, wang2019robust} \\
& \multicolumn{3}{c|}{} & \citet{zhao-etal-2020-knowledge,shang2019end,jin-etal-2019-fine,nathani-etal-2019-learning} \\
& \multicolumn{3}{c|}{} & \citet{sorokin2018modeling,cao2019multi,han-etal-2020-open, xie2020reinceptione} \\ \cline{2-5}
& \multicolumn{3}{c|}{\multirow{2}{*}{Coreference Graph}} & \citet{sahu-etal-2019-inter,qian2019graphie,Xu2020DocumentGF,xu2020discourse} \\
& \multicolumn{3}{c|}{} & \citet{de2018question,luan2019general} \\ \cline{2-5} 
& \multicolumn{3}{c|}{Topic Graph} & \citet{linmei-etal-2019-heterogeneous,li2020leveraging} \\ \cline{2-5} 
& \multicolumn{3}{c|}{\multirow{3}{*}{Similarity Graph Construction}}  & \citet{Xia_Huang_Liu_Shi_2019,yao2019graph,yasunaga-etal-2017-graph} \\ 
& \multicolumn{3}{c|}{}  & \citet{linmei-etal-2019-heterogeneous, zhou_neural_2020,wang-etal-2020-heterogeneous} \\
& \multicolumn{3}{c|}{}  & \citet{DBLP:conf/acl/LiuNWGHLX19,hu2020multi,jia-etal-2020-neural,li2020leveraging} \\ \cline{2-5} 
& \multicolumn{3}{c|}{\multirow{3}{*}{Co-occurrence Graph}}  & \citet{christopoulou-etal-2019-connecting,zhang-qian-2020-convolution,hu2020multi} \\
& \multicolumn{3}{c|}{}  & \citet{zhang2020every,yao2019graph, de2018question} \\
& \multicolumn{3}{c|}{}  & \citet{edouard-etal-2017-graph,zhu_graphbtm_2018,DBLP:conf/acl/LiuNWGHLX19} \\ \cline{2-5} 
& \multicolumn{3}{c|}{\multirow{8}{*}{App-driven Graph}} & \citet{ ding-etal-2019-neural,yin-etal-2020-novel,luo-zhao-2020-bipartite,DBLP:conf/acl/DingZCYT19}  \\
& \multicolumn{3}{c|}{} & \citet{sui2019leverage,tang2020multi, DBLP:conf/emnlp/RanLLZL19, DBLP:conf/ijcai/HuCL0MY19} \\
& \multicolumn{3}{c|}{} & \citet{gui-etal-2019-lexicon,li-goldwasser-2019-encoding,xiao-etal-2019-lattice,xu-etal-2018-sql} \\
& \multicolumn{3}{c|}{} & \citet{qu2020few,bogin-etal-2019-global,huo-etal-2019-graph,Shao_Gong_Qi_Cao_Ji_Lin_2020} \\
& \multicolumn{3}{c|}{} & \citet{fernandes2018structured,DBLP:conf/aaai/LiuYZWL20,DBLP:conf/emnlp/HuangMLZW19,linmei-etal-2019-heterogeneous} \\
& \multicolumn{3}{c|}{} & \citet{bogin-etal-2019-representing, leclair2020improved, qiu-etal-2019-dynamically,DBLP:conf/acl/ZhengWLDCJZL20} \\ 
& \multicolumn{3}{c|}{} & \citet{ ferreira-freitas-2020-premise, zheng2020srlgrn, fang2020hierarchical} \\
& \multicolumn{3}{c|}{} & \citet{allamanis2018learning,christopoulou-etal-2019-connecting, thayaparan-etal-2019-identifying} \\ \hline
\multirow{6}{*}{Dynamic graph} & {Graph Similarity} & \multicolumn{2}{c|}{Node Embedding Based } &  \citet{chen2020graphflow, chen2020reinforcement, chen2020iterative, chen2020inducing} \\ \cline{3-5} 
& Metric Learning  & \multicolumn{2}{c|}{Structure-aware} &  \citet{liu2019contextualized, liu2021retrieval} \\ \cline{2-5} 
& \multicolumn{3}{c|}{Graph Sparsification Techniques}  & \citet{chen2020graphflow, chen2020reinforcement, chen2020iterative} \\ \cline{2-5} 
& \multicolumn{3}{c|}{ Combining Intrinsic and}  & \multirow{2}{*}{ \citet{chen2020iterative, liu2021retrieval} }\\
& \multicolumn{3}{c|}{ Implicit Graph Structures}  & \\\hline
\end{tabular}
}
\label{tab:graph-construction}
\end{table}

\subsection{Static Graph Construction}
\label{subsec: Static-Graph-Construction}

The static graph construction approach aims to construct the graph structures during preprocessing typically by leveraging existing relation parsing tools (e.g., dependency parsing) or manually defined rules. Conceptually, a static graph incorporates different domain/external knowledge hidden in the original text sequences, which augments the raw text with rich structured information. 

In this subsection, we summarize various static graph construction methods in the GNN for NLP literature and group them into totally eleven categories.
 We assume that the input is a document $doc = \{ para_1, para_2, ..., para_n\}$, which consists of $n$ paragraph denoted as $para$. Similarly, a paragraph consists of $m$ sentences denoted as $para_i = \{sent_1, sent_2, ..., sent_m\}$. Each sentence then consists of $l$ words denoted as $sent_i = \{w_1, w_2, ..., w_l\}$.

\subsubsection{Static Graph Construction Approaches}

\begin{figure}[ht!]
\centering
\vspace{-2mm}
\includegraphics[width=12.0cm]{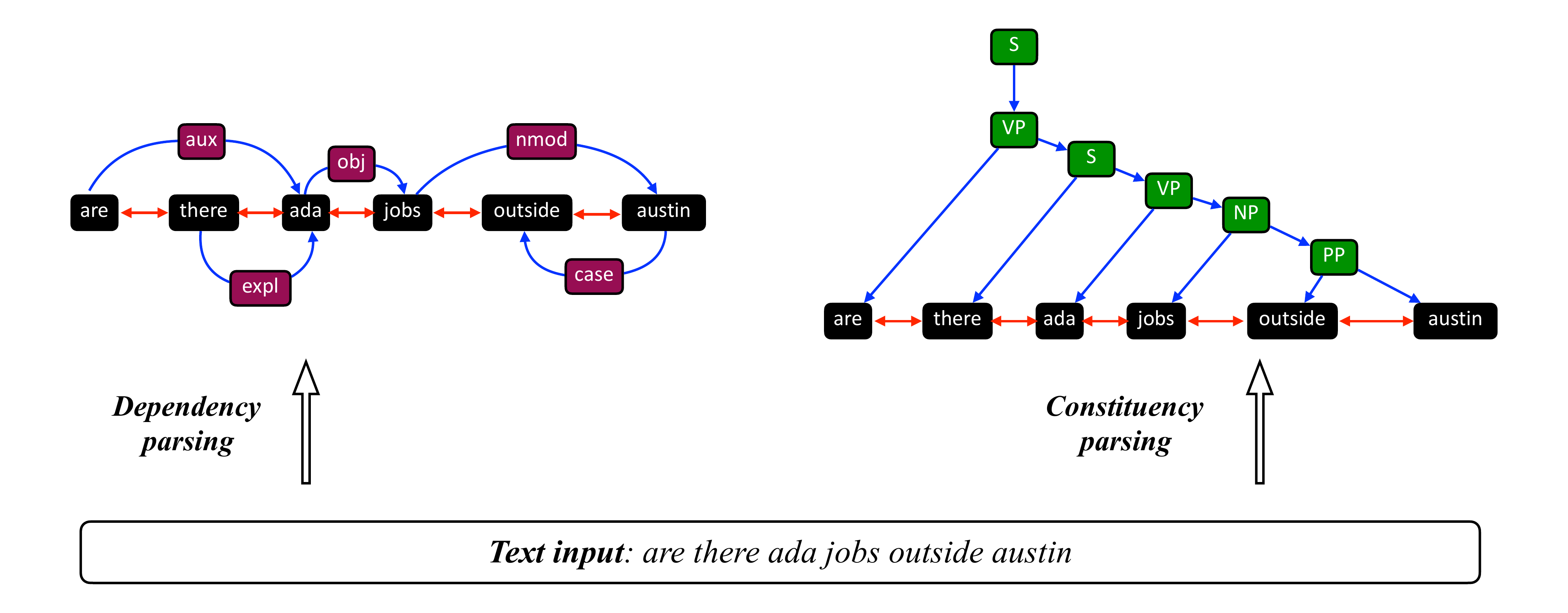}
\vspace{-4mm}
\caption{An example is shown for the dependency graph (left) and the constituency graph (right), respectively. The text input is from \textbf{JOBS640}~\citep{luke2005ze} dataset.}
\label{fig:dependency-constitency-graph-sample}
\vspace{-0mm}
\end{figure}

\paragraph{Dependency Graph Construction}
\label{subsubsec:dep-graph}
The dependency graph is widely used to capture the dependency relations between different objects in the given sentences. Formally, given a paragraph, one can obtain the dependency parsing tree (e.g., syntactic dependency tree or semantic dependency parsing tree) by using various NLP parsing tools (e.g., Stanford CoreNLP~\citep{lee2011stanford}). Then one may extract the dependency relations from the dependency parsing tree and convert them into a dependency graph~\citep{xu2018graph2seq,song-etal-2018-graph}.
Moreover, since the given paragraph has sequential information while the graph nodes are unordered, one may introduce the sequential links to reserve such vital information in the graph structure~\citep{sahu-etal-2019-inter,qian2019graphie,xu-etal-2018-exploiting,li-etal-2017-context}. Next, we will discuss a representative dependency graph construction method given the inputs $para$ and its extracted parsing tree, including three key steps: 1) constructing dependency relation, 2) constructing sequential relation, and 3) final graph conversion. An example for the dependency graph is shown in Fig. \ref{fig:dependency-constitency-graph-sample}.


\noindent\emph{Step 1: Dependency Relations.} Given the sentences in a specific paragraph, one first obtains the dependency parsing tree for each sentence. We denote dependency relations in the dependency tree as $(w_i, rel_{i, j}, w_j)$, where $w_i$, $w_j$ are the word nodes linked by an edge type $rel_{i, j}$. Conceptually, an edge denotes a dependency relation "$w_i$ depends on $w_j$ with relation $rel_{i, j}$". We define the dependency relation set as $\mathcal{R}_{dep}$.


\noindent\emph{Step 2: Sequential Relations.} \label{sec:sequential_relation} The sequential relation encodes the adjacent relation of the elements in the original paragraph. Specifically, for dependency graph constructing, we define the sequential relation set $\mathcal{R}_{seq} \subseteq \mathcal{V} \times \mathcal{V}$, where $\mathcal{V}$ is the basic element (i.e., word) set. For each sequential relation $(w_i, w_{i+1}) \in \mathcal{R}_{seq}$, it means $w_i$ is adjacent to $w_{i+1}$ in the given paragraph.


\noindent\emph{Step 3: Dependency Graph.} The dependency graph $\mathcal{G}(\mathcal{V}, \mathcal{E})$ consists of the word nodes and two relations discussed above. Given the paragraph $para$, dependency relation set $\mathcal{R}_{dep}$, and the sequential relation set $\mathcal{R}_{seq}$, firstly, for each relation $(w_i, rel_{i, j}, w_j) \in \mathcal{R}_{dep}$, one adds the nodes $v_i$ (for the word $w_i$) and $v_j$ (for the word $w_j$) and  a directed edge from node $v_i$ to node $v_j$ with edge type $rel_{i, j}$. Secondly, for each relation $(w_i, w_j) \in \mathcal{R}_{seq}$, one adds two nodes $v_i$ (for the word $w_i$) and $v_j$ (for the word $w_j$) and an undirected edge between nodes $v_i$ and $v_j$ with specific sequential type.


\paragraph{Constituency Graph Construction}
\label{subsubsec: Constituency Graph Construction}
The constituency graph is another widely used static graph that is able to capture phrase-based syntactic relations in one or more sentences. Unlike dependency parsing, which only focuses on one-to-one correspondences between single words (i.e., word level), constituency parsing models the assembly of one or several corresponded words (i.e., phrase level). Thus it provides a new insight about the grammatical structure of a sentence. In this following subsection, we will discuss the typical approach for constructing a  constituency graph~\citep{li-etal-2020-graph-tree,marcheggiani-titov-2020-graph,xu-etal-2018-exploiting}. We first explain the basic concepts of the constituency relations and then illustrate the constituency graph construction procedure. An example for the Constituency graph is shown in Fig. \ref{fig:dependency-constitency-graph-sample}.

\noindent\emph{Step 1: Constituency Relations.}
In linguistics, constituency relation means the relation following the phrase structure grammars instead of the dependency relation and dependency grammars. Generally, the constituency relation derives from the subject(noun phrase NP)-predicate(verb phrase VP) relation. In this part, we only discuss the constituency relation deriving from the constituency parsing tree. Unlike the dependency parsing tree, in which all nodes have the same type, the constituency parsing tree distinguishes between the terminal and non-terminal nodes. Non-terminal categories of the constituency grammar label the parsing tree's interior nodes (e.g., S for sentence, and NP for noun phrase). In contrast, the leaf nodes are labeled by terminal categories (words in sentences). The nodes set can be denoted as: 1) non-terminal nodes set $\mathcal{V}_{nt}$ (e.g. S and NP) and 2) terminal nodes set $\mathcal{V}_{words}$. The constituency relation set are associated with the tree's edges, which can be denoted as $\mathcal{R}_{cons} \subseteq \mathcal{V}_{nt} \times (\mathcal{V}_{nt} + \mathcal{V}_{words})$.

\noindent\emph{Step 2: Constituency Graph.}
A constituency graph $\mathcal{G}(\mathcal{V}, \mathcal{E})$ consists of both the non-terminal nodes $\mathcal{V}_{nt}$ and the terminal nodes $\mathcal{V}_{words}$, and the constituency edges as well as the sequential edges. Similar to the dependency graph, given a paragraph $para$ and the constituency relation set $\mathcal{R}_{cons}$, for each constituency relation $(w_i, rel_{i, j}, w_j) \in \mathcal{R}_{cons}$, one adds the nodes $v_i$ (for the word $w_i$) and $v_j$ (for the word $w_j$) and a directed edge from node $v_i$ to node $v_j$. And then for each word nodes pair $(v_i, v_j)$ for the words which are adjacent in the original text, one adds an undirected edge between them with the specific sequential type. These sequential edges are used to reserve the sequential information~\citep{li-etal-2020-graph-tree,xu-etal-2018-exploiting}.



\begin{figure}[ht!]
\centering
\includegraphics[width=10.0cm]{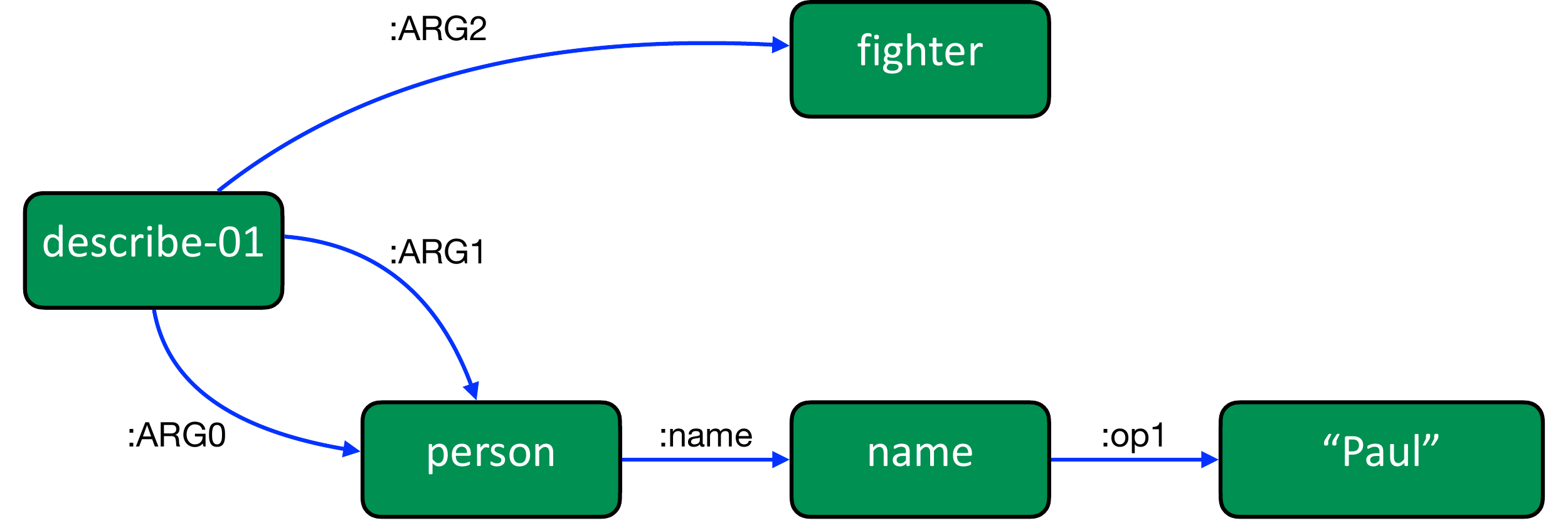}
\caption{An example of AMR graph, the original sentence is "Pual's description of himself: a fighter".}
\label{fig:amr-graph-sample}
\vspace{-2mm}
\end{figure}

\paragraph{AMR Graph Construction}
\label{subsubsec: AMR Graph Construction}
The AMR graphs are rooted, labeled, directed, acyclic graphs, which are widely used to represent the high-level semantic relations between abstract concepts of the unstructured and concrete natural text.
Different from the syntactic idiosyncrasies, the AMR is the high-level semantic abstraction. More concretely, the different sentences that are semantically similar may share the same AMR parsing results, e.g., "Paul described himself as a fighter" and "Paul's description of himself: a fighter", as shown in Fig.~\ref{fig:amr-graph-sample}. Despise the fact that the AMR is biased toward English, it is a powerful auxiliary representation for linguistic analysis
~\citep{song2018graph,damonte-cohen-2019-structural,wang2020amr}. 
Similar to the previously introduced dependency and constituency trees, an AMR graph is derived from an AMR parsing tree. Next, we focus on introducing the general procedure of constructing the AMR graph based on the AMR parsing tree. We will discuss the basic concept of AMR relation and then show how to convert the relations into an AMR graph.

\noindent\emph{Step 1: AMR Relations.}
Conceptually, there are two types of nodes in the AMR parsing tree: 1) the name (e.g. "Paul") is the specific value of the node instance and 2) the concepts are either English words (e.g. "boy"), PropBank framesets~\citep{kingsbury2002treebank} (e.g. "want-01"), or special keywords. The name nodes are the unique identities, while the concept nodes are shared by different instances. The edges that connect nodes are called relations (e.g. :ARG0 and :name). One may extract these AMR relations from the node pairs with edges, which is denoted as $(n_i, r_{i, j}, n_j) \in \mathcal{R}_{amr}$.

\noindent\emph{Step 2: AMR Graph.}
The AMR graph $\mathcal{G}(\mathcal{V}, \mathcal{E})$, which is rooted, labeled, directed, acyclic graph (DAG), consists of the AMR nodes and AMR relations discussed above. Similar to the dependency and constituency graphs, given the sentence $sent$ and the AMR relation set $\mathcal{R}_{amr}$, for each relation $(n_i, r_{i, j}, n_j) \in \mathcal{R}_{amr}$, one adds the nodes $v_i$ (for the AMR node $n_i$) and $v_j$ (for the AMR node $n_j$) and add a directed edge from node $v_i$ to node $v_j$ with edge type $r_{i, j}$

\begin{figure}[ht!]
\centering
\vspace{-2mm}
\includegraphics[width=12.0cm]{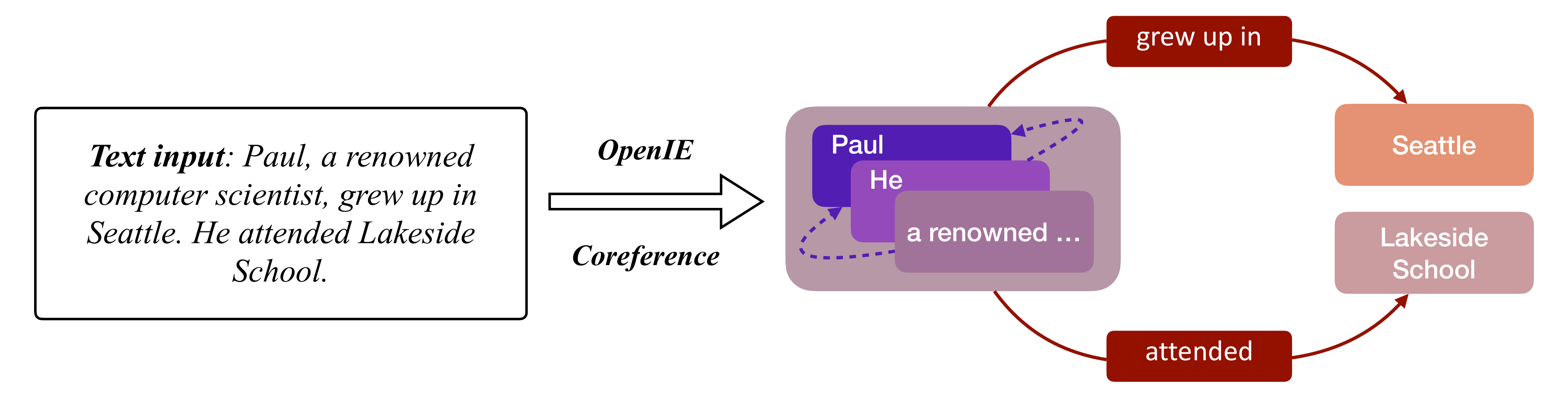}
\vspace{-4mm}
\caption{An example for IE graph construction which contains both the Co-reference process and the Open Information Extraction process.}
\label{fig:IE-graph-construction-sample}
\vspace{-2mm}
\end{figure}

\paragraph{Information Extraction Graph Construction} 
\label{subsubsec: IE Graph Construction}
The information extraction graph (IE Graph) aims to extract the structural information to represent the high-level information among natural sentences, e.g., text-based documents.
These extracted relations that capture relations across distant sentences have been demonstrated helpful in many NLP tasks~\citep{wu2020extracting,vashishth-etal-2018-reside,gupta-etal-2019-care}. In what follows, we will discuss the technical details on how to construct an IE graph for a given paragraph $para$~\citep{huang-etal-2020-knowledge,vashishth-etal-2018-reside}. We divide this process into two three basic steps: 1) coreference resolution, 2) constructing IE relations, and 3) graph construction.



\noindent\emph{Step 1: Coreference Resolution.}
\label{sec: coreference-resolution}
Coreference resolution is the basic procedure for information extraction task which aims to find expressions that refer to the same entities in the text sequence~\citep{huang-etal-2020-knowledge}. As shown in Figure \ref{fig:IE-graph-construction-sample},
the name "Pual", the noun-term "He" and "a renowned computer scientist" may refer to the same object (person). Many NLP tools such as OpenIE~\citep{angeli-etal-2015-leveraging} provide coreference resolution function to achieve this goal. We denotes the coreference cluster $C$ as a set of phrases referring to the same object. Given a paragraph, one can obtain the coreference sets $\mathcal{C} = \{C_1, C_2, ..., C_n\}$ extracting from unstructured data. 

\noindent\emph{Step 2: IE Relations.}
To construct an IE graph, the first step is to extract the triples from the paragraphs, which could be completed by leveraging some well-known information extraction systems (i.e. OpenIE~\citep{angeli-etal-2015-leveraging}). We call each triple (subject, predicate, object) as a relation, which is denoted $(n_i, r_{i, j}, n_j) \in \mathcal{R}_{ie}$. It is worth noting if two triples differ only by one argument, and the other arguments overlap, one only keep the longer triple.

\noindent\emph{Step 3: IE Graph Construction.} 
The IE graph $\mathcal{G}(\mathcal{V}, \mathcal{E})$ consists of the IE nodes and IE relations discussed above. Given the paragraph $para$ and the IE relation set $\mathcal{R}_{ie}$, for each relation $(n_i, r_{i, j}, n_j) \in \mathcal{R}_{ie}$, one adds the nodes $v_i$ (for the subject $n_i$) and $v_j$ (for the object $n_j$) and add a directed edge from node $v_i$ to node $v_j$ with the corresponding predicate types~\citep{huang-etal-2020-knowledge}. And then, for each coreference cluster $C_i \in \mathcal{C}$, one may collapse all coreferential phrases in $C_i$ into one node. This could help greatly reduce the number of nodes and eliminate the ambiguity by keeping only one node.


\paragraph{Discourse Graph Construction}
\label{subsubsec: Discourse Graph Construction}
Many NLP tasks suffer from long dependency challenge when the candidate document is too long. The discourse graph, which describes how two sentences are logically connected to one another, are proved effective to tackle such challenge~\citep{christensen2013towards}. In the following subsection, we will briefly discuss the discourse relations between given sentences and then introduce the general procedure to construct the discourse graphs~\citep{song-etal-2018-n,li2020leveraging,yasunaga-etal-2017-graph,xu2020discourse}.



\noindent\emph{Step 1: Discourse Relation.}
The discourse relations derive from the discourse analysis, which aims to identify sentence-wise ordering constraints over a set of sentences. Given two sentences $sent_i$ and $sent_j$, one can define the discourse relation as $(sent_i, sent_j)$, which represents the discourse relation "sentence $sent_j$ can be placed after sentence $sent_i$." The discourse analysis has been explored for years, and many theories have been developed for modeling discourse relations such as the Rhetorical Structure Theory (RST)~\citep{mann1987rhetorical} and G-Flow~\citep{christensen2013towards}. In many NLP tasks, given a document $doc$, one firstly segments $doc$ into sentences set $\mathcal{V} = \{sent_1, sent_2, ..., sent_m\}$. Then one applies discourse analysis to get the pairwise discourse relation set denoted as $\mathcal{R}_{sep} \subseteq \mathcal{V} \times \mathcal{V}$.

\noindent\emph{Step 2: Discourse Graph.} 
The discourse graph $\mathcal{G}(\mathcal{V}, \mathcal{E})$ consists of the sentences nodes and discourse relations discussed above. Given the document $doc$ and the discourse relation set $\mathcal{R}_{dis}$, for each relation $(sent_i, sent_j) \in \mathcal{R}_{dis}$, one adds the nodes $v_i$ (for the sentence $sent_i$) and $v_j$ (for the sentence $sent_j$) and add a directed edge from node $v_i$ to node $v_j$.


\begin{figure}[ht!]
\centering
\vspace{-2mm}
\includegraphics[width=12.0cm]{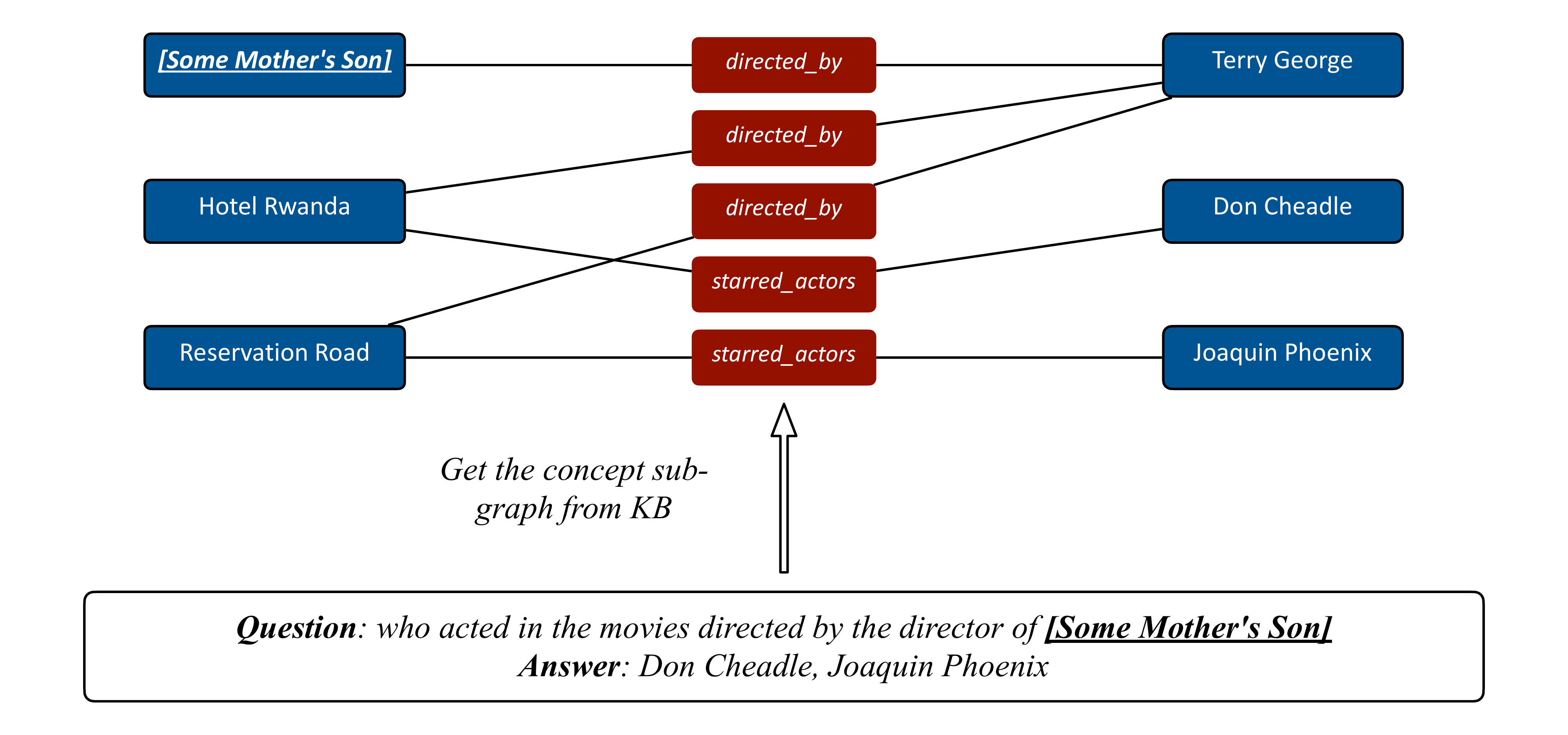}
\vspace{-4mm}
\caption{An example for knowledge graph construction, where the knowledge base (KB) used and the generated concept graph are both from the dataset \textit{MetaQA}~\citep{zhang2018variational}.}
\label{fig:KG-graph-construction-sample}
\vspace{-2mm}
\end{figure}

\paragraph{Knowledge Graph Construction}
\label{subsubsec: Knowledge Graph Construction}
Knowledge Graph (KG) that captures entities and relations can greatly facilitate learning and reasoning in many NLP applications. In general, the KGs can be divided into two main categories depending on their graph construction approaches. Many applications treat the KG as the compact and interpretable intermediate representation of the unstructured data (e.g., the document)~\citep{wu2020extracting,koncel-kedziorski-etal-2019-text,huang-etal-2020-knowledge}. Conceptually, it is almost similar to the IE graph, which we have discussed previously. 
On the other hand, many other works~\citep{wu-etal-2020-knowledge,ye2019vectorized,bansal2019a2n,yang2019aligning} incorporate the existing knowledge bases such as YAGO~\citep{suchanek2008yago}) and ConceptNet~\citep{speer2017conceptnet} to further enhance the performance of downstream tasks~\citep{zhao-etal-2020-knowledge}. 
In what follows, we will briefly discuss the second category of KG from the view of the graph construction.

The KG can be denoted as $\mathcal{G}(\mathcal{V}, \mathcal{E})$, which is usually constructed by elements in knowledge base. Formally, one defines the triple $(e_1, rel, e_2)$ as the basic elements in the knowledge base, in which $e_1$ is the source entity, $e_2$ is the target entity, and $rel$ is the relation type. Then one adds two nodes $v_1$ (for the source element $e_1$) and $v_2$ (for the target element $e_2$) in the KG and add a directed edge from node $v_1$ to node $v_2$ with edge type $rel$. An example of such KG is shown in Fig. \ref{fig:KG-graph-construction-sample}.

It is worth noting that the KG plays different roles in various applications. In some applications (e.g. knowledge graph completion and knowledge base question answering), KG is always treated as part of the inputs. In this scenario~\citep{ye2019vectorized,zhang2020relational,li2019semi,wu-etal-2020-temp}, 
researchers typically use the whole KG $\mathcal{G}$ as the learning object. But for some other applications (e.g. natural language translation), the KG can be treated as the data augmentation method. In this case, the whole KG such as ConceptNet~\citep{speer2017conceptnet} is usually too large and noisy for some domain-specific applications~\citep{Kapanipathi2020InfusingKI,lin-etal-2019-kagnet}, and thus it is not suitable to use the whole graph as inputs. In contrast, as shown in Figure \ref{fig:KG-graph-construction-sample}, 
one instead usually constructs subgraphs from the given query (it is often the text-based inputs like the queries in the reading comprehension task) ~\citep{xu2019dynamically,teru2020inductive,Kapanipathi2020InfusingKI}.

The construction methods could may vary dramatically in the literature. Here, we only present one representative method for illustration purpose~\citep{teru2020inductive}. The first thing for constructing KG is to fetch the term instances in the given query. Then, they could link the term instances to the concepts in the KG by some matching algorithms such as max-substring matching. The concepts are regarded as the initial nodes in the extracted subgraph. Next step is to fetch the 1-hop neighbors of the initial nodes in the KG. Additionally, one may calculate the relevance of the neighbors with the initial nodes by applying some graph node relevance model such as the Personalized PageRank (PPR) algorithm~\citep{page1999pagerank}. Then based on the results, one may further prune out the edges with relevance score that is below the confidence threshold and remove the isolated neighbors. The remaining final subgraph is then used to feed any graph representation learning module later.

\paragraph{Coreference Graph Construction}
\label{subsubsec: Coreference Graph Construction}
In linguistics, coreference (or co-reference) occurs when two or more terms in a given paragraph refer to the same object. Many works have demonstrated that such phenomenon is helpful for better understanding the complex structure and logic of the corpus and resolve the ambiguities~\citep{Xu2020DocumentGF,de2018question,sahu-etal-2019-inter}.
To effectively leverage the coreference information, the coreference graph is constructed to explicitly model the implicit coreference relations. Given a set of phrases, a coreference graph can link the nodes (phrases) which refer to the same entity in the text corpus. In the following subsection, we focus on the coreference graph construction for a paragraph $para$ consisting of $m$ sentences. We will briefly discuss the coreference relation and then discuss the approaches for building the coreference graph in various NLP tasks~\citep{de2018question,sahu-etal-2019-inter,qian2019graphie,Xu2020DocumentGF,xu2020discourse,luan2019general}. It is worth noting that although it is similar to the IE graph's first step, the coreference graph will explicitly model the coreference relationship by graph instead of collapse into one node.

\noindent\emph{Step 1: Coreference Relation.}
The coreference relations can be obtained easily by the coreference resolution system, as discussed in IE graph construction. Similarly, we can obtain the coreference clusters $\mathcal{C}$ given a specific paragraph. All phrases in a cluster $C_i \in \mathcal{C}$ refer to the same object.


\noindent\emph{Step 2: Coreference Graph.}
The coreference graph are built on the coreference relation set $\mathcal{R}_{coref}$. It can be generally divided into two main category depending on the node type: 1) phrases (or mentions)~\citep{koncel-kedziorski-etal-2019-text, luan2019general,de2018question}, 2) words~\citep{sahu-etal-2019-inter}. For the first class, the coreference graph $\mathcal{G}$ consists of all mentions in relation set $\mathcal{R}_{coref}$. For each phrase pair $p_i, p_j$ in cluster $C_k \in \mathcal{C}$, one may add an undirected edge between node $v_i$ (for the phrase $p_i$) and node $v_j$ (for the phrase $p_j$). For the second case, the coreference graph $\mathcal{G}$ consists of words. One minor difference is that one only links the first word of each phrase for each associated phrases.




\begin{figure}[ht!]
\centering
\vspace{0mm}
\includegraphics[width=10.0cm]{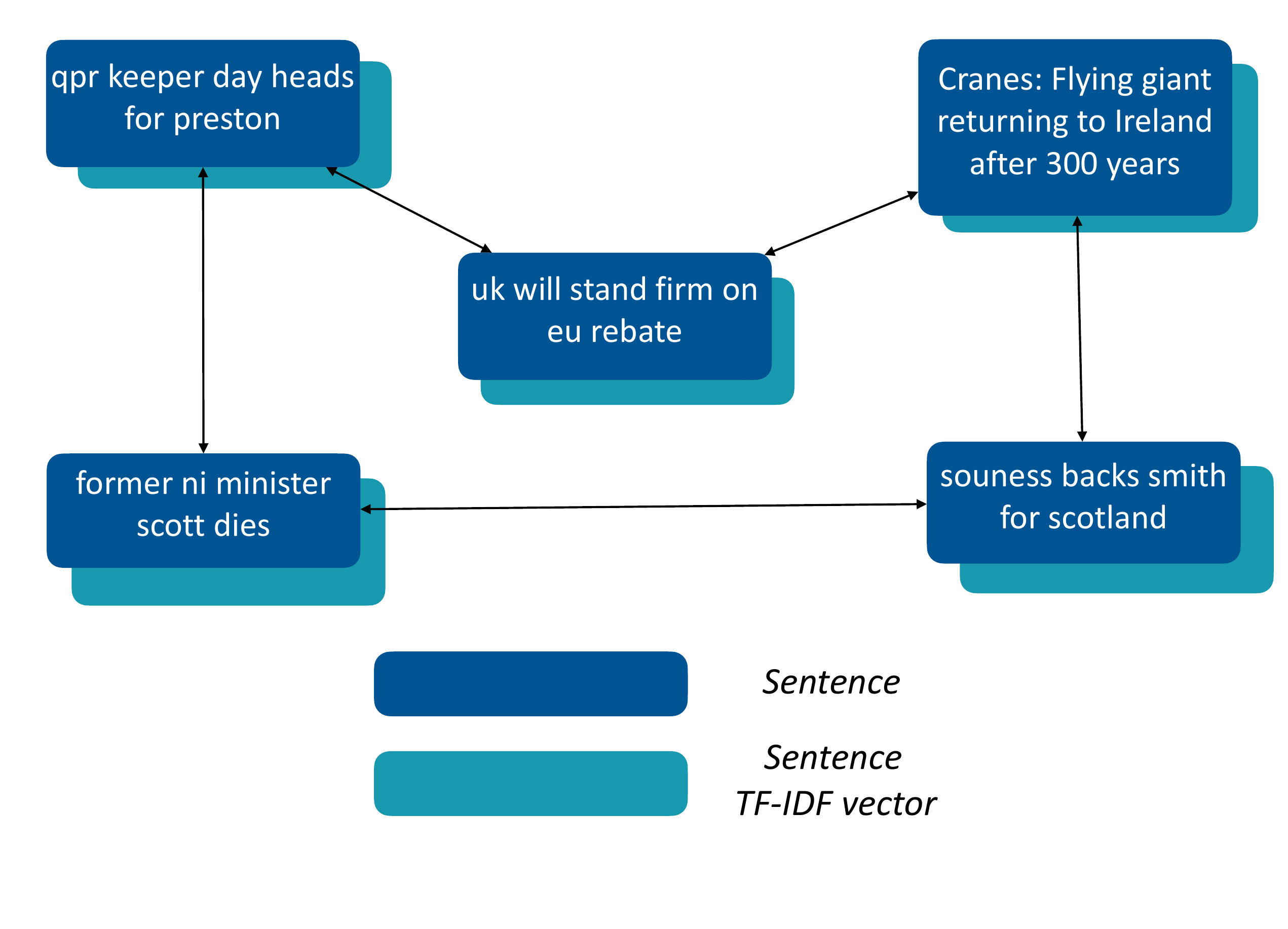}
\vspace{-4mm}
\caption{An example for similarity graph construction. We use sentences as nodes and initialize their features with \textit{TF-IDF} vectors.}
\label{fig:sim-graph-construction-sample}
\vspace{-2mm}
\end{figure}

\paragraph{Similarity Graph Construction}
\label{subsubsec: Similarity Graph Construction}
The similarity graphs aim to quantify the similarity between nodes, which are widely used in many NLP tasks~\citep{DBLP:conf/acl/LiuNWGHLX19,linmei-etal-2019-heterogeneous,yasunaga-etal-2017-graph}.
Since the similarity graph is typically application-oriented, we focus on the basic procedure of constructing the similarity graph for various types of elements such as entities, sentences and documents, and neglect the application specific details. It is worth noting that the similarity graph construction is conducted during preprocessing and is not jointly trained with the remaining learning system in an end-to-end manner.
One example of similarity graph is shown in Fig. \ref{fig:sim-graph-construction-sample}. 

\noindent\emph{Step 1: Similarity Graph.}
Given a corpus $C$, in a similarity graph $\mathcal{G}(\mathcal{V}, \mathcal{E})$, the graph nodes can be defined in different granularity levels such as entities, sentences and documents. We denote the basic node set as $\mathcal{V}$ regardless of specific node types. One can calculate the node features by various mechanisms such as TF-IDF for sentences (or documents)~\citep{DBLP:conf/acl/LiuNWGHLX19,yasunaga-etal-2017-graph} and embeddings for entities~\citep{linmei-etal-2019-heterogeneous}. Then, similarity scores between node pairs can be computed by various metrics such as cosine similarity~\citep{DBLP:conf/acl/LiuNWGHLX19,linmei-etal-2019-heterogeneous,yasunaga-etal-2017-graph}, and used to indicate edge weights of the node pairs.

\noindent\emph{Step 2: Sparse mechanism.}
The initial similarity graph is typically dense even some edge weights are very small or even negative. These values can be treated as noise, which plays little roles in the similarity graph. Thus various sparse techniques are proposed to further improve the quality of graph by sparsifying a graph. One widely used sparse method is k-NN~\citep{DBLP:conf/acl/LiuNWGHLX19}. Specifically, for node $v_i$ and its' neighbor set $N(v_i)$, one only reserves edges by keeping $k$ largest edge weights and dropping the remaining edges. The other widely used method is $\epsilon-$sparse~\citep{linmei-etal-2019-heterogeneous,yasunaga-etal-2017-graph}. In particular, one will remove the edges whose weights are smaller than the certain threshold $\epsilon$.

\begin{figure}[ht!]
\centering
\vspace{-2mm}
\includegraphics[width=12.0cm]{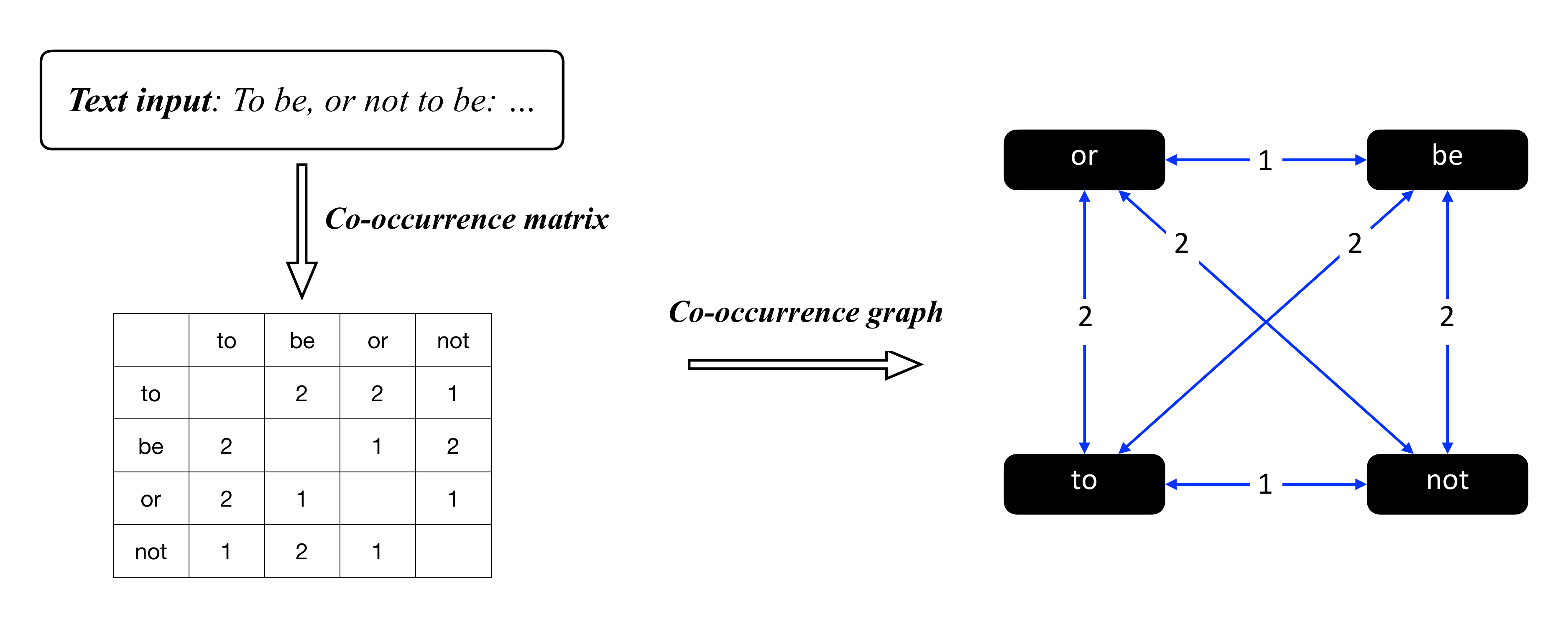}
\vspace{-6mm}
\caption{An example for co-occurrence graph construction where edge weights stand for the co-occurrence frequency between words. We set the window size as 3.}
\label{fig:co-occurrence-graph-construction-sample}
\vspace{-4mm}
\end{figure}

\paragraph{Co-occurrence Graph Construction}
\label{subsubsec: Co-occurrence Graph Construction}


The co-occurrence graph aims to capture the co-occurrence relation between words in the text, which is widely used in many NLP tasks
~\citep{christopoulou-etal-2019-connecting,zhang-qian-2020-convolution,zhang2020every}.
The co-occurrence relation, which describes the frequency of two words that co-occur within a fix-sized context window, is an important feature capturing the semantic relationship among words in the corpus. In what follows, we will first present the approaches of obtaining the co-occurrence relations and then discuss the basic procedure of building a co-occurrence graph for a corpus $C$. An example of co-occurrence graph can be seen in Fig. \ref{fig:co-occurrence-graph-construction-sample}.



\noindent\emph{Step 1: Co-occurrence Relation.}
The co-occurrence relation is defined by the co-occurrence matrix of the given corpus $C$. For a specific paragraph $para$ consists of $m$ sentences, the co-occurrence matrix describes how words occur together. One may denote the co-occurrence the matrix as $\mathbf{M} \in \mathbb{R}^{|V|\times |V|}$, where $|V|$ is the vocabulary size of $C$. $\mathbf{M}_{w_i, w_j}$ describes how many times word $w_i$, $w_j$ occur together within a fix-size sliding windows in the corpus $C$. After obtaining the co-occurrence matrix, there are two typical methods to calculate the weights between words: 1) co-occurrence frequency~\citep{zhang2020every,christopoulou-etal-2019-connecting,zhang-qian-2020-convolution,edouard-etal-2017-graph,zhu_graphbtm_2018} and 2) point-wise mutual information (PMI)~\citep{yao2019graph,hu2020multi,hu2019hierarchical}.

\noindent\emph{Step 2: Co-occurrence Graph.}
The co-occurrence graph $\mathcal{G}(\mathcal{V}, \mathcal{E})$ consists of the words nodes and co-occurrence relations discussed above. Given the corpus $C$ and the co-occurrence relation set $\mathcal{R}_{co}$, for each relation $(w_i, w_j) \in \mathcal{R}_{co}$, one adds the nodes $v_i$ (for the word $w_i$) and $v_j$ (for the word $w_j$) and add an undirected edge from node $v_i$ to node $v_j$ initialized with the aforementioned calculated edge weights.


\begin{figure}[ht!]
\centering
\vspace{-2mm}
\includegraphics[width=12.0cm]{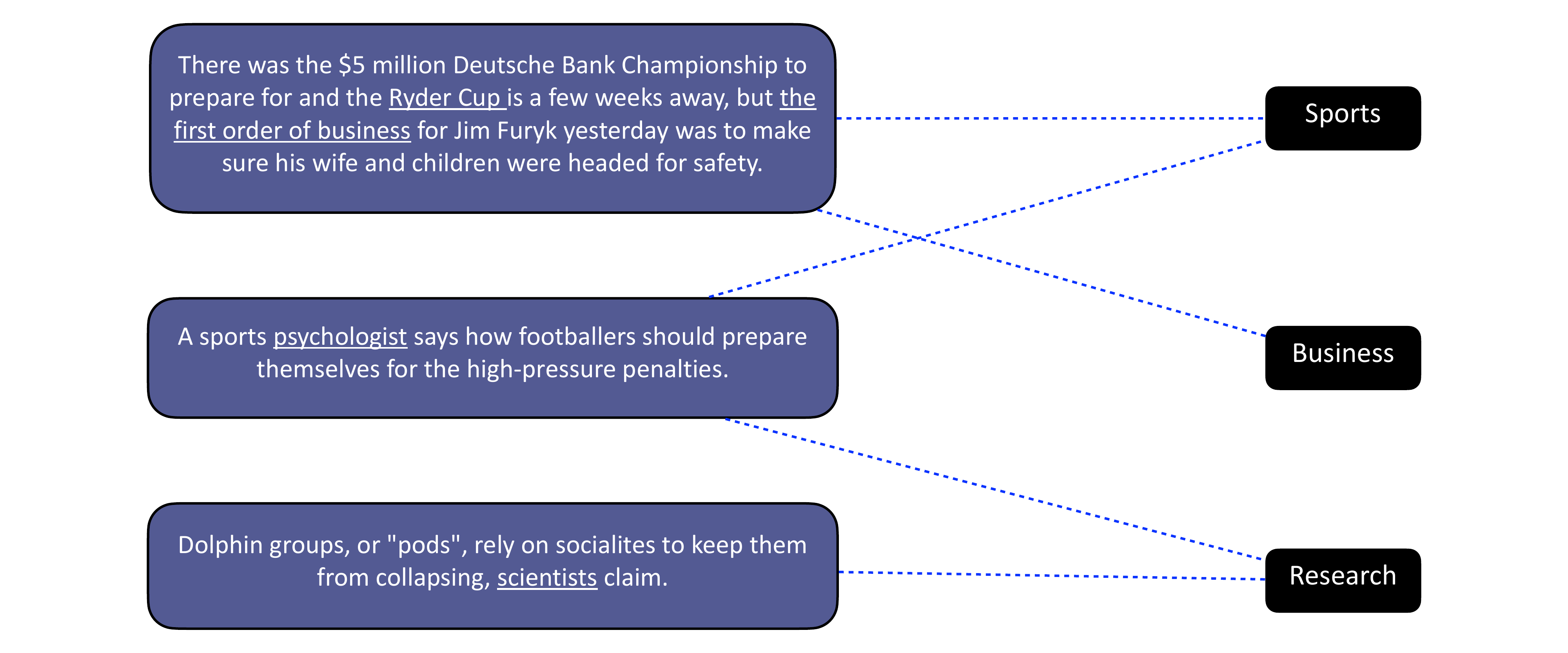}
\vspace{-4mm}
\caption{An example for topic graph construction, where the dash line stands for the topic modeling process by leveraging the LDA algorithm on dataset \textit{AG news}~\citep{zhang2015character}.}
\label{fig:topic-graph-construction-sample}
\vspace{-4mm}
\end{figure}

\paragraph{Topic Graph Construction}
\label{subsubsec: Topic Graph Construction}
The topic graph is built on several documents, which aims to model the high-level semantic relations among different topics~\citep{linmei-etal-2019-heterogeneous,li2020leveraging}. In particular, given a set of documents $\mathcal{D} = \{doc_1, doc_2, ..., doc_m\}$, one first learns the latent topics denoted as $\mathcal{T}$ using some topic modeling algorithms such as LDA~\citep{blei2003latent}. Then one could construct the topic graph $\mathcal{G}(\mathcal{V}, \mathcal{E})$ with $\mathcal{V} = \mathcal{D} \cup \mathcal{T}$. The undirected edge between the node $v_i$ (for a document) and the node $v_j$ (for a topic) is built only if the document has that topic. An example of topic graph is shown in Fig. \ref{fig:topic-graph-construction-sample}.


\begin{figure}[ht!]
\centering
\includegraphics[width=12.0cm]{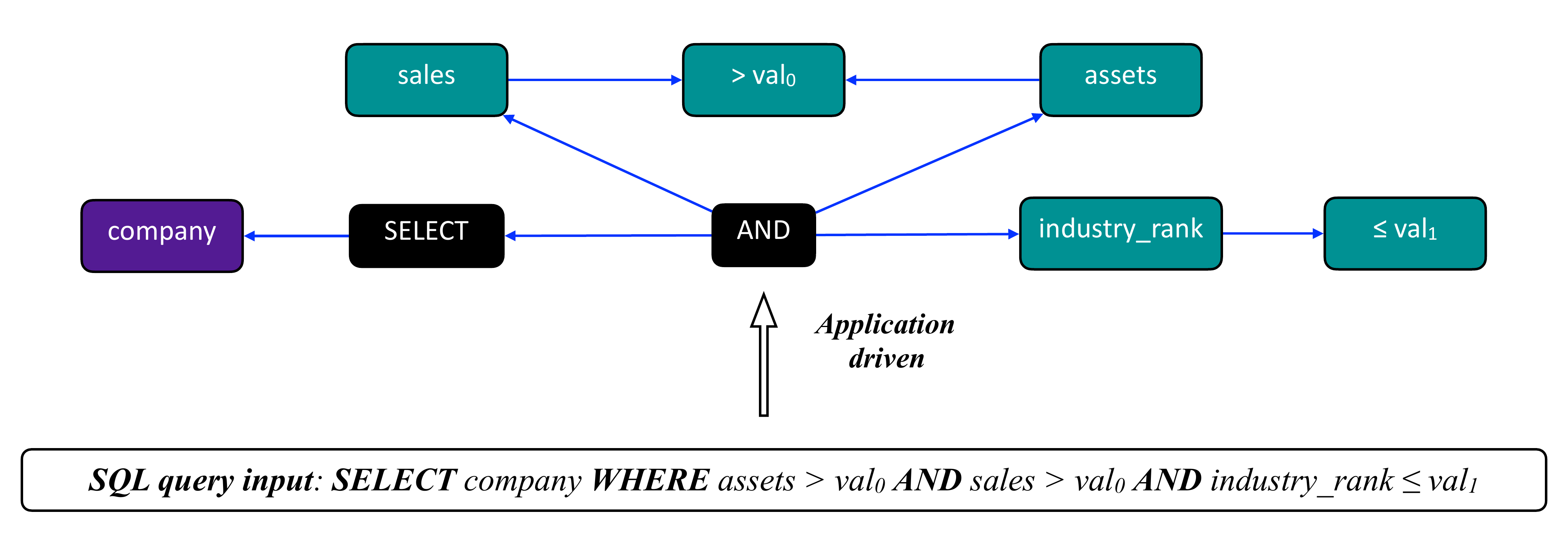}
\vspace{-2mm}
\caption{An example for application-driven graph construction, which is specially designed for SQL query input.}
\label{fig:application-driven-graph-construction-sample}
\vspace{-2mm}
\end{figure}

\paragraph{App-driven Graph Construction}
\label{subsubsec: App-driven Graph Construction}
The app-driven graphs refer to the graph specially designed for specific NLP tasks
~\citep{gui-etal-2019-lexicon,ding-etal-2019-neural,yin-etal-2020-novel,luo-zhao-2020-bipartite}, 
which cannot be trivially covered by the previously discussed static graph types. In some NLP tasks, it is common to represent unstructured data by structured formation with application-specific approaches. For example, the SQL language can be naturally represented by the SQL parsing tree. Thus it can be converted to the SQL graph~\citep{xu-etal-2018-sql,bogin-etal-2019-representing,huo-etal-2019-graph,bogin-etal-2019-global}. Since these graphs are too specialized based on the domain knowledge, there are no unified pattern to summarize how to build an app-driven graph. An example of such application-driven graph like SQL graph is illustrated in Fig. \ref{fig:application-driven-graph-construction-sample}. In Sec. \ref{sec:Applications}, we will further discuss how these graph construction methods are used in various popular NLP tasks.

\subsubsection{Hybrid Graph Construction and Discussion}
Most previous static graph construction methods only consider one specific relation between nodes. Although the obtained graphs capture the structure information well to some extent, they are also limited in exploiting different types of graph relations. To address this limitation, there is an increasing interest in building a hybrid graph by combing several graphs together in order to enrich the semantic information in graph~\citep{jia-etal-2020-neural,sahu-etal-2019-inter,xu-etal-2018-exploiting,zeng-etal-2020-double,Xu2020DocumentGF,xu2020discourse,christopoulou-etal-2019-connecting,yao2019graph}. The method of constructing a hybrid graph is mostly application specific, and thus we only present some representative approach for such a graph construction.


To capture multiple relations, a common strategy is to construct a heterogeneous graph, which contains multiple types of nodes and edges. Without losing generality, we assume that one may create a hybrid graph $\mathcal{G}_{hybrid}$ with two different graph sources $\mathcal{G}_{a}(\mathcal{V}_{a},\mathcal{E}_{a})$ and  $\mathcal{G}_{b}(\mathcal{V}_{b}, \mathcal{E}_{b})$. Graphs $a, b$ are two different graph types such as \textit{dependency graph} and \textit{constituency graph}. Given these textual inputs, if $\mathcal{G}_{a}$ and $\mathcal{G}_{b}$ share the same node sets (i.e., $\mathcal{V}_{a}$ = $\mathcal{V}_{b}$), we merge the edge sets by annotating relation-specific edge types~\citep{xu-etal-2018-exploiting,sahu-etal-2019-inter,zeng-etal-2020-double,Xu2020DocumentGF,xu2020discourse}. Otherwise, we merge the $\mathcal{V}_{a}$ and $\mathcal{V}_{b}$ to get the hybrid node set, denoted as $\mathcal{V} = \mathcal{V}_{a} \cup \mathcal{V}_{b}$~\citep{jia-etal-2020-neural,christopoulou-etal-2019-connecting}. Then we generate $\mathcal{E}_{a}$ and $\mathcal{E}_{b}$ to $\mathcal{E}$ by mapping the source and target nodes from $\mathcal{V}_{a}$ and $\mathcal{V}_{b}$ to $\mathcal{V}$.


\subsection{Dynamic Graph Construction}
\label{subsec:Dynamic Graph Construction}


Although static graph construction has the advantage of encoding prior knowledge of the data into the graph structure, it has several limitations.
First of all, extensive human efforts and domain expertise are needed in order to construct a reasonably performant graph topology.
Secondly, the manually constructed graph structure might be error-prone (e.g., noisy or incomplete). Thirdly, since the graph construction stage and graph representation learning stage are disjoint, the errors introduced in the graph construction stage cannot be corrected and might be accumulated to later stages, which can result in degraded performance.
Lastly, the graph construction process is often informed solely by the insights of the NLP practitioners, and might be sub-optimal for the downstream prediction task.

\begin{figure}[thb!]
\centering
\vspace{-2mm}
\includegraphics[keepaspectratio=true,scale=0.15]{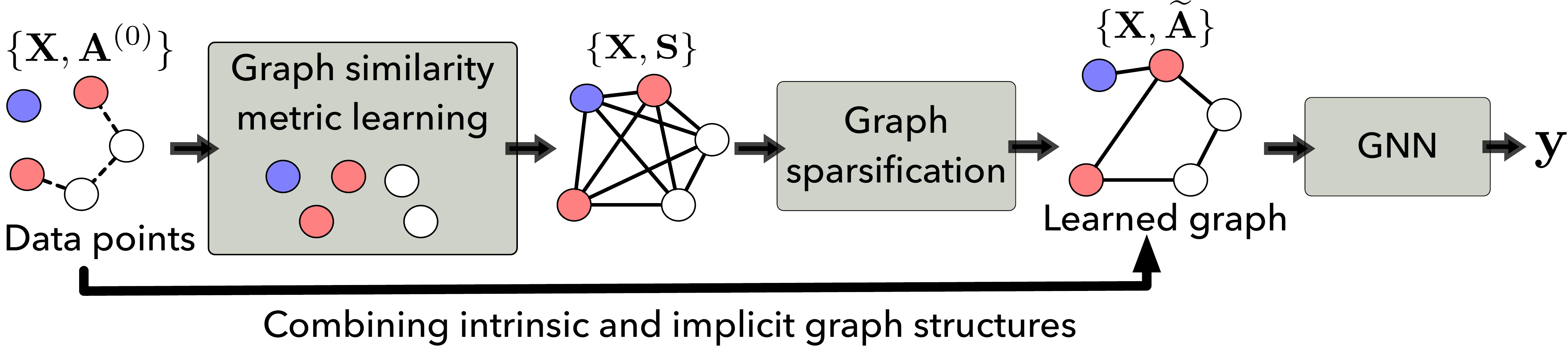}
\vspace{-4mm}
\caption{Overall illustration of dynamic graph construction approaches. Dashed lines (in data points on left) indicate the optional intrinsic graph topology.}
\label{fig:dynamic_graph_overall}
\vspace{-0mm}
\end{figure}

In order to tackle the above challenges, recent attempts on GNN for NLP~\citep{chen2020iterative,chen2020reinforcement,liu2021retrieval,liu2019contextualized} have explored dynamic graph construction without resorting to human efforts or domain expertise.
Most dynamic graph construction approaches aim to dynamically learn the graph structure (i.e., a weighted adjacency matrix) on the fly, and the graph construction module can be jointly optimized with the subsequent graph representation learning modules toward the downstream task in an end-to-end manner.
One good example of dynamic graph construction is when constructing a graph capturing the semantic relationships among all the words in a text passage in the task of conversational machine reading comprehension~\citep{reddy2019coqa}, instead of building a fixed static graph based on domain expertise, one can jointly train a graph structure learning module together with the graph embedding learning module in order to learn an optimal graph structure considering not only the semantic meanings of the words but also the conversation history and current question.

As shown in ~\cref{fig:dynamic_graph_overall}, these dynamic graph construction approaches typically consist of a graph similarity metric learning component for learning an adjacency matrix by considering pair-wise node similarity in the embedding space, and a graph sparsification component for extracting a sparse graph from the learned fully-connected graph. 
It is reported to be beneficial to combine the intrinsic graph structures and learned implicit graph structures for better learning performance~\citep{li2018adaptive,chen2020iterative,liu2021retrieval}.
Moreover, in order to effectively conduct the joint graph structure and representation learning, various learning paradigms have been proposed.
In what follows, we will discuss all these effective techniques for dynamic graph construction.
More broadly speaking, graph structure learning for GNNs itself is a trending research problem in the machine learning field, and has been actively studied beyond the NLP community~\citep{li2018adaptive,norcliffe2018learning,velickovic2020pointer,kalofolias2019large,franceschi2019learning}. However, in this survey, we will focus on its recent advances in the NLP field.
We hereafter use dynamic graph construction and graph structure learning interchangeably.

\subsubsection{Graph Similarity Metric Learning Techniques} 
\label{subsubsec: Graph Similarity Metric Learning Techniques}

Based on the assumption that node attributes contain useful information for learning the implicit graph structure, recent work has explored to cast the graph structure learning problem as the problem of similarity metric learning defined upon the node embedding space.
The learned similarity metric function can be later applied to an unseen set of node embeddings to infer a graph structure, thus enabling inductive graph structure learning.
For data deployed in non-Euclidean domains like graphs, the Euclidean distance is not necessarily the best metric for measuring node similarity.
Various similarity metric functions have been proposed for graph structure learning for GNNs.
According to the types of information sources utilized, we group these metric functions into two categories: \emph{Node Embedding Based Similarity Metric Learning} and \emph{Structure-aware Similarity Metric Learning}.

\paragraph{Node Embedding Based Similarity Metric Learning}

Node embedding based similarity metric functions are designed to learn a weighted adjacency matrix by computing the pair-wise node similarity in the embedding space. Common metric functions include attention-based metric functions and cosine-based metric functions.

\noindent\textbf{Attention-based Similarity Metric Functions.}
Most of the similarity metric functions proposed so far are based on the attention mechanism~\citep{bahdanau2015neural,vaswani2017attention}. 
In order to increase the learning capacity of dot product based attention mechanism, \citet{chen2020graphflow} proposed a modified dot product by introducing learnable parameters, formulated as follows:
\begin{equation}
\begin{aligned}
S_{i,j} = (\vec{v}_i \odot \vec{u})^T \vec{v}_j
\end{aligned}
\end{equation}
where $\vec{u}$ is a non-negative weight vector learning to highlight different dimensions of the node embeddings, and $\odot$ denotes element-wise multiplication.

Similarly, \citet{chen2020reinforcement} designed a more expressive version of dot product by introducing a learnable weight matrix, formulated as follows:
\begin{equation}
\begin{aligned}
S_{i,j} = \mathrm{ReLU}(\vec{W} \vec{v}_i)^T \mathrm{ReLU}(\vec{W} \vec{v}_j)
\end{aligned}
\end{equation}
where $\vec{W}$ is a $d \times d$ weight matrix, and $\mathrm{ReLU}(x) = \max(0, x)$ is a rectified linear unit (ReLU)~\citep{nair2010rectified} used to enforce the sparsity of the similarity matrix.

\noindent\textbf{Cosine-based Similarity Metric Functions.}
\citet{chen2020iterative} extended the vanilla cosine similarity to a multi-head weighted cosine similarity to capture pair-wise node similarity from multiple perspectives, formulated as follows:
\begin{equation}
\begin{aligned}
S_{i,j}^p &= \mathrm{cos}(\vec{w}_p \odot \vec{v}_i, \vec{w}_p \odot \vec{v}_j)\\
S_{i,j} &= \frac{1}{m}\sum_{p=1}^{m}{S_{ij}^p}
\end{aligned}
\end{equation}
where $\vec{w}_p$ is a weight vector associated to the $p$-th perspective, and has the same dimension as the node embeddings.
Intuitively, $S_{i,j}^p$ computes the pair-wise cosine similarity for the $p$-th perspective where each perspective considers one part of the semantics captured in the embeddings.
Besides increasing the learning capacity, employing multi-head learners is able to stabilize the learning process, which has also been observed in~\citep{vaswani2017attention,velivckovic2017graph}.

\paragraph{Structure-aware Similarity Metric Learning}
\label{sec:graph-construction-structure-aware}
Inspired by structure-aware transformers~\citep{zhu2019modeling,cai2020graph}, recent approaches employ structure-aware similarity metric functions that additionally consider the existing edge information of the intrinsic graph beyond the node information. For instance,
\citet{liu2019contextualized} proposed a structure-aware attention mechanism for learning pair-wise node similarity, formulated as follows:
\begin{equation}
\begin{aligned}
     S_{i, j}^l = \mathrm{softmax}(\vec{u}^T \mathrm{tanh}(\vec{W}[\vec{h}^l_i, \vec{h}^l_j, \vec{v}_i, \vec{v}_j, \vec{e}_{i,j}]))
\end{aligned}
\end{equation}
where $\vec{v}_i$ represents the embedding of node $i$, $\vec{e}_{i,j}$ represents the embedding of the edge connecting node $i$ and $j$, $\vec{h}^l_i$ is the embedding of node $i$ in the $l$-th GNN layer, and $\vec{u}$ and $\vec{W}$ are trainable weight vector and weight matrix, respectively.

Similarly, \citet{liu2021retrieval} introduced a structure-aware global attention mechanism, formulated as follows,
\begin{equation}
\begin{aligned}
S_{i, j} = \frac{\mathrm{ReLU}(\vec{W}^Q \vec{v}_i )^T (\mathrm{ReLU}( \vec{W}^K \vec{v}_i)+ \mathrm{ReLU}(\vec{W}^R \vec{e}_{i, j}))}{\sqrt{d}}
\end{aligned}
\end{equation}
where $\vec{e}_{i, j}$ is the embedding of the edge connecting node $i$ and $j$, and $\vec{W}^Q$, $\vec{W}^K$, and $\vec{W}^R$ are linear transformations that map the node and edge embeddings to the latent embeddding space.

\subsubsection{Graph Sparsification Techniques}

Most graphs in real-world scenarios are sparse graphs.
Similarity metric functions consider relations between any pair of nodes and returns a fully-connected graph, which is not only computationally expensive but also might introduce noise such as unimportant edges.
Therefore, it can be beneficial to explicitly enforce sparsity to the learned graph structure. 
Besides applying the $\mathrm{ReLU}$ function in the similarity metric functions~\citep{chen2020reinforcement,liu2021retrieval}, various graph sparsification techniques have been adopted to enhance the sparsity of the learned graph structure.

\citet{chen2020reinforcement,chen2020graphflow} applied a kNN-style sparsification operation to obtain a sparse adjacency matrix from the node similarity matrix computed by the similarity metric learning function, formulated as follows:
\begin{equation}
\begin{aligned}
\vec{A}_{i,:} = \mathrm{topk}(\vec{S}_{i,:})
\end{aligned}
\end{equation}
where for each node, only the $K$ nearest neighbors (including itself) and the associated similarity scores are kept, and the remaining similarity scores are masked off.

\citet{chen2020iterative} enforced a sparse adjacency matrix by considering only the $\varepsilon$-neighborhood for each node, formulated as follows:
\begin{equation}
\begin{aligned}
A_{i,j} = 
\left\{
        \begin{array}{ll}
             S_{i,j} & \quad  S_{i,j} > \varepsilon  \\
              0 & \quad \text{otherwise}
        \end{array}
    \right.
\end{aligned}
\end{equation}
where those elements in $S$ which are smaller than a non-negative threshold $\varepsilon$ are all masked off (i.e., set to zero).

Besides explicitly enforcing the sparsity of the learned graph by applying certain form of threshold, sparsity has also been enforced implicitly in a learning-based manner. 
\citet{chen2020iterative} introduced the following regularization term to encourage sparse graphs.
\begin{equation}
\begin{aligned}
\frac{1}{n^2} ||A||_F^2
\end{aligned}
\end{equation}
where $||\cdot||_F$ denotes the Frobenius norm of a matrix.

\subsubsection{Combining Intrinsic Graph Structures and Implicit Graph Structures} 

Recent studies~\citep{li2018adaptive,chen2020iterative,liu2021retrieval} have shown that it could hurt the downstream task performance if the intrinsic graph structure is totally discard while doing dynamic graph construction. This is probably because the intrinsic graph typically still carries rich and useful information regarding the optimal graph structure for the downstream task.
They therefore proposed to combine the learned implicit graph structure with the intrinsic graph structure based on the assumption that the learned implicit graph is potentially a ``shift'' (e.g., substructures) from the intrinsic graph structure which is supplementary to the intrinsic graph structure.
The other potential benefit is incorporating the intrinsic graph structure might help accelerate the training process and increase the training stability. Since there is no prior knowledge on the similarity metric and the trainable parameters are randomly initialized, it may usually take long time to converge.

Different ways for combining intrinsic and implicit graph structures have been explored.
For instance, \citet{li2018adaptive,chen2020iterative} proposed to compute a linear combination of the normalized graph Laplacian of the intrinsic graph structure $L^{(0)}$ and the normalized adjacency matrix of the implicit graph structure $\mathrm{f}(A)$, formulated as follows:
\begin{equation}
\begin{aligned}
\widetilde{A} = \lambda L^{(0)} + (1 - \lambda) \mathrm{f}(A)
\end{aligned}
\end{equation}
where $\mathrm{f}: \mathbb{R}^{n \times n} \to \mathbb{R}^{n \times n}$ can be arbitrary normalization operations such as graph Laplacian operation~\citep{li2018adaptive} and row-normalization operation~\citep{chen2020iterative}.
Instead of explicitly fusing the two graph adjacency matrices, \citet{liu2021retrieval} proposed a hybrid message passing mechanism for GNNs which fuses the two aggregated node vectors computed from the intrinsic graph and the learned implicit graph, respectively, and then feed the fused vector to a GRU to update node embeddings.

\subsubsection{Learning Paradigms}

Most existing dynamic graph construction approaches for GNNs consist of two key learning components: graph structure learning (i.e., similarity metric learning) and graph representation learning (i.e., GNN module), and the ultimate goal is to learn the optimized graph structures and representations with respect to certain downstream prediction task. 
How to optimize the two separate learning components toward the same ultimate goal becomes an important question.
Here we highlight three representative learning paradigms.
The most straightforward strategy~\citep{chen2020reinforcement,chen2020graphflow,liu2021retrieval} is to jointly optimize the whole learning system in an end-to-end manner toward the downstream (semi-)supervised prediction task.
Another common strategy~\citep{yang2018glomo,liu2019contextualized,huang2020location} is to adaptively learn the input graph structure to each stacked GNN layer to reflect the changes of the intermediate graph representations.
This is similar to how transformer models learn different weighted fully-connected graphs in each layer.
Unlike the above two paradigms, \citet{chen2020iterative} proposed an iterative graph learning framework by learning a better graph structure based on better graph representations, and in the meantime, learning better graph representations based on a better graph structure in an iterative manner.
As a result, this iterative learning paradigm repeatedly refines the graph structure and the graph representations toward the optimal downstream performance.

\section{Graph Representation Learning for NLP}
\label{sec:Graph Representation Learning for NLP}



In the previous section, we have presented various graph construction methods, including static graph construction and dynamic graph construction. 
In this section, we will discuss various graph representation learning techniques that are directly operated on the constructed graphs for various NLP tasks. 
The goal of graph representation learning is to find a way to incorporate information of graph structures and attributes into a low-dimension embeddings via a machine learning model~\citep{hamilton2017representation}. To mathematically formalize this problem, we give the mathematical notations for arbitrary graphs as $\mathcal{G}(\mathcal{V}, \mathcal{E}, \mathcal{T}, \mathcal{R})$, where $\mathcal{V}$ is the node set, $\mathcal{E}$ is the edge set, $\mathcal{T} = \{T_1, T_2, ..., T_{p}\}$ is the collection of node types, and $\mathcal{R} = \{R_1, ..., R_q\}$ is the collection of edge types. $|\cdot|$ is the number of elements. $\tau(\cdot) \in \mathcal{T}$ is the node type indicator function (e.g., $\tau(v_i) \in \mathcal{T}$ is the type of node $v_i$), and $\phi(\cdot) \in \mathcal{R}$ is the edge type indicator function (e.g., $\phi(e_{i, j}) \in \mathcal{R}$ is the type of edge $e_{i, j}$), respectively.

Generally speaking, the constructed graphs from the raw text data are either homogeneous or heterogeneous graphs. Thus, in \cref{sec:graph_representation_learning_homogeneous}, we will discuss various graph representation learning methods for homogeneous graphs, including scenarios for the original homogeneous graph and some converting from heterogeneous graphs. In \cref{sec:graph_representation_learning_heterogeneous}, we will discuss the GNN-based methods for multi-relational graphs, and in \cref{subsec:graph-representation-learning-heterogeneousgnn}, we will discuss the GNNs for dealing with the heterogeneous graphs. 



\subsection{GNNs for Homogeneous Graphs}
\label{sec:graph_representation_learning_homogeneous}
By definition, a graph $\mathcal{G}(\mathcal{V}, \mathcal{E}, \mathcal{T}, \mathcal{R}), s.t. |\mathcal{T}| = 1, |\mathcal{R}| = 1$ is called homogeneous graph. Most graph neural networks such as GCN, GAT, and GraphSage are designed for homogeneous graphs, which, however, can not fit well in many NLP tasks. For example, given a natural language text, the constructed dependency graph is arbitrary graph that contains multiple relations, which cannot be exploited by traditional GNN methods directly. Thus, in this subsection, we will first discuss the various strategies for converting arbitrary graphs to homogeneous graphs, including static graphs and dynamic graphs. Then, we will discuss the graph neural networks considering bidirectional encoding.


\subsubsection{Static Graph: Treating Edge Information as Connectivity} 
GNNs for dealing with the static graphs normally consists of two stages, namely, converting edge information and node representation learning, as described in the following.

\paragraph{Converting Edge Information to Adjacent Matrix} 

Basically, the edges are viewed as the connection information between nodes. In this case, it is normal to discard the edge type information and retain the connections to convert the heterogeneous graphs~\citep{yang2019aligning,zhang_aspect-based_2019,yao2019graph,wu2019jointly,li2019semi} to homogeneous graphs. After obtaining such a graph, we can represent the topology of the graph as a unified adjacency matrix $A$. Specifically, for an edge $e_{i, j} \in \mathcal{E}$ which connect node $v_i$ and $v_j$, $A_{i,j}$ denotes to the edge weight for weighted static graph, or $A_{i,j}=1$ for unweighted connections and $A_{i,j}=0$ otherwise. 
The static graphs can also be divided into directed and undirected graphs. For the undirected case, the adjacency matrix is symmetric matrix, which means $A_{i, j} = A_{j,i}$. And for the other case, it is always not symmetric. The $A_{i, j}$ is strictly defined by the edge from node $v_i$ to node $v_j$. It is worth noting that the directed graphs can be transformed to undirected graphs~\citep{yasunaga-etal-2017-graph} by averaging the edge weights in both directions. The edge weights are rescaled, whose maximum edge weight is 1 before feeding to the GNN.



\paragraph{Node Representation Learning} 
Next, given initial node embedding $\mathbf{X}$ and adjacency matrix $A$, the node representation is extracted base on the classical GNNs techniques. 
For undirected graphs, most works~\citep{liu2019learning,wang2018cross,yao2019graph,DBLP:conf/acl/ZhangYCWWW20} mainly adopt graph representation learning algorithms such as GCN, GGNN, GAT, GraphSage, etc. and stack them to explore deep semantic correlations in the graph. When it comes to the directed graphs, few GNN methods such as GGNN, GAT still work~\citep{chen2020graphflow,DBLP:conf/coling/WangXLZS20,qiu-etal-2019-dynamically,yan-etal-2019-event,ji-etal-2019-graph,sui2019leverage}. While for the other GNNs that can not be directly applied into directed graphs, the simple strategy is to ignore the directions (i.e., converting the directed graphs to undirected graphs)~\citep{yasunaga-etal-2017-graph,wang2018cross,liu2019learning}. However, such methods allow the message to propagate in both directions without constraints. To solve this problem, many efforts have been made to adapt the GNN to directed graphs. For GCN, some spatial-based GCN algorithms are designed for directed graphs such as DCNN~\citep{NIPS2016_390e9825}. GraphSage can be easily extended to directed graphs by modifying the aggregation function via specifying the edge directions and aggregating them separately)~\citep{xu2018graph2seq}.



\subsubsection{Dynamic Graph}
Dynamic graphs that aim to learn the graph structure together with the downstream task jointly are widely adopted by graph representation learning~\citep{chen2020reinforcement,chen2020graphflow,chen2020iterative,hashimoto-tsuruoka-2017-neural,guo-etal-2019-attention}.
Early works mainly adopt the recurrent network by treating the graph node embeddings as RNN's state encodings~\citep{hashimoto-tsuruoka-2017-neural}, which can be regarded as the rudiment of GGNN. Then the classic GNNs such as GCN~\citep{guo-etal-2019-attention}, GAT~\citep{cui2020enhancing}, GGNN~\citep{chen2020reinforcement,chen2020graphflow} are utilized to learn the graph embedding effectively.
Recent researchers adopt attention-based or metric learning-based mechanisms to learn the implicit graph structure (i.e., the graph adjacency matrix $A$) from unstructured texts. The learning process of graph structure is jointly with the downstream tasks via an end-to-end fashion~\citep{shaw-etal-2018-self,chen2020graphflow,luan2019general,chen2020reinforcement}. 

\subsubsection{Graph Neural Networks: Bidirectional Graph Embeddings}
\label{sec:bidirectional-graph-embeddings}
In the previous sub-sections, we present the typical techniques for constructing and learning node embeddings from the static homogeneous graphs. In this subsection, we provide a detailed discuss on how to handle the edge directions. In reality, many graphs are directed acyclic graphs~(DAG)~\citep{cai2020graph}, which information is propagated along the specific edge direction. 
However, some researchers allow the information propagate equally in both directions~\citep{yasunaga-etal-2017-graph,wang2018cross,liu2019learning} and others discard the information containing in outgoing edges~\citep{yan-etal-2019-event,DBLP:conf/coling/WangXLZS20,qiu-etal-2019-dynamically,ji-etal-2019-graph,sui2019leverage}, both of which will lose some important structure information for the final representation learning.

To deal with this, bidirectional graph neural network (bidirectional GNN) is proposed to learn the node representation from both incoming and outgoing edges in a interleaved fashion. 
To introduce different variants of bidirectional GNN, we first give some unified mathematical notations. For a specific node $v_i \in \mathcal{V}$ in the graph $\mathcal{G}(\mathcal{V}, \mathcal{E})$ and its neighbor nodes $N(v_i)$ (i.e. any node $v_j$ satisfy $e_{i, j} \in \mathcal{E}$ or $e_{j, i} \in \mathcal{E}$), we define the incoming (backward) nodes set as $N_{\dashv}(v_i)$ satisfying $e_{j, i} \in \mathcal{E}, v_{j} \in N_{\dashv}(v_i)$ and outgoing (forward) nodes set as $N_{\vdash}(v_i)$ holding $e_{i, j} \in \mathcal{E}, v_{j} \in N_{\vdash}(v_i)$.

\citet{xu2018graph2seq} firstly extend the GraphSage to a bi-directional version by calculating the graph embedding separately for each direction and combine them at last. At each computation hop, for each node in the graph, they aggregate the incoming nodes and outgoing nodes separately to get backward and forward immediate-aggregated representation as follows:
\begin{equation}
    \begin{split}
    \mathbf{h}^{(k)}_{i, \dashv} &= \sigma(\mathbf{W}^{(k)}\cdot f_k^{\dashv}(\mathbf{h}^{(k-1)}_{i, \dashv},\{\mathbf{h}^{(k-1)}_{j, \dashv}, \forall v_j\in N_{\dashv}(v_i)\})), \\
    \mathbf{h}^{(k)}_{i, \vdash} &= \sigma(\mathbf{W}^{(k)}\cdot f_k^{\vdash}(\mathbf{h}^{(k-1)}_{i, \vdash},\{\mathbf{h}^{(k-1)}_{j, \vdash}, \forall v_j\in N_{\vdash}(v_i)\})),
    \end{split}
\end{equation}
where $k \in \{1, 2, ..., K\}$ denotes the layer number, and $\mathbf{h}^{(k)}_{i, \dashv}, \mathbf{h}^{(k)}_{i, \vdash}$ denote the backward and forward aggregated results respectively. At the final step, the final forward and backward representation is concatenated to calculate the final bi-directional representation.



Although works effectively, the bidirectional GraphSage learns both directions separately. To this end, \citet{chen2020reinforcement} proposes the bidirectional GGNN to address this issue. Technically, at each iteration, 
after obtaining aggregated vector representations $\mathbf{h}^{(k)}_{i, \vdash}, \mathbf{h}^{(k)}_{i, \dashv}$, they opt to fuse them into one vector as follows:
\begin{equation}
    \mathbf{h}^{(k)}_{N(v_i)} = \text{Fuse}(\mathbf{h}^{(k)}_{i, \vdash}, \mathbf{h}^{(k)}_{i, \dashv}),
\end{equation}
where the function $\text{Fuse}(\cdot, \cdot)$ is the gated sum of two information sources:
\begin{equation}
    \text{Fuse}(\mathbf{a}, \mathbf{b}) = \mathbf{z} \odot \mathbf{a} + (1 - \mathbf{z}) \odot \mathbf{b}, \mathbf{z} = \sigma (\mathbf{W}_{z} [\mathbf{a},\mathbf{b},\mathbf{a}\odot\mathbf{b},\mathbf{a}-\mathbf{b}] + \mathbf{b}_z)
\end{equation}
where $\mathbf{a} \in \mathbb{R}^d, \mathbf{b} \in \mathbb{R}^d$ are inputs, $\mathbf{W}_z \in \mathbb{R}^{d \times 4d}, \mathbf{b}_z \in \mathbb{R}^{d}$ are learnable weights and $\sigma(\cdot)$ is the sigmoid function. Then, a Gated Recurrent Unit (GRU) is used to update the node embeddings by
incorporating the aggregation information as follows:
\begin{equation}
    \mathbf{h}^{(k)}_{i} = \text{GRU}(\mathbf{h}^{(k-1)}_{i}, \mathbf{h}^{(k)}_{N(v_i)}).
\end{equation}

Unlike previous methods, which are specially designed for the specific GNN methods, \citet{ribeiro-etal-2019-enhancing} further proposes a general bidirectional GNN framework, which can be easily applied to most existing GNNs. Technically, they first encode the graph in two directions:
\begin{equation}
\begin{split}
    \mathbf{h}^{(k)}_{i, \dashv} &= \text{GNN}(\mathbf{h}^{(k-1)}_i, \{\mathbf{h}^{(k-1)}_j: \forall v_j \in N_{\dashv}(v_i)  \}), \\
    \mathbf{h}^{(k)}_{i, \vdash} &= \text{GNN}(\mathbf{h}^{(k-1)}_i, \{\mathbf{h}^{(k-1)}_j: \forall v_j \in N_{\vdash}(v_i)  \}),
\end{split}
\end{equation}
where $\text{GNN}(\cdot)$ can denote to any variant of GNNs. Similar to strategy in the bidirectional RNNs~\citep{650093}, they learn the forward and backward directions separately and concatenate them together with the original node feature as follows:
\begin{equation}
    \mathbf{h}^{(k)}_i = [\mathbf{h}^{(k)}_{i, \dashv},\mathbf{h}^{(k)}_{i, \vdash},\mathbf{h}^{(k-1)}_{i}].
\end{equation}
They stack several layers to achieve better performance. At last, the $\mathbf{h}^{(K)}_i$ is employed in a sequence input of a Bi-LSTM in depth-first order to the final node representation.

\subsection{Graph Neural Networks for Multi-relational Graphs}
\label{sec:graph_representation_learning_heterogeneous}

In practice, many graphs have various edge types, such as knowledge graph, AMR graph, etc., which can be formalized as multi-relational graphs. Formally, for a graph $\mathcal{G}$, s.t. $|\mathcal{T}| = 1$ and $|\mathcal{R}| > 1$ is defined as multi-relational graphs. In this section, we introduce different techniques for representing and learning the multi-relational graphs. Specifically, in Section. \ref{subsec:graph-representation-learning-asnode}, we will discuss the formalization for the multi-relational graphs from an original heterogeneous graph. 
In Section. \ref{subsec:graph-representation-learning-rgnn} and \ref{subsec:graph-representation-learning-advanced-rgnn}, we will discuss the basic graph representation learning methods and transformers for relational heterogeneous, respectively (we denote it as multi-relational graph neural network for simplification). 



\subsubsection{Multi-relational Graph Formalization}
\label{subsec:graph-representation-learning-asnode}
Since heterogeneous graphs are commonly observed in NLP domian, such as knowledge graph, AMR graph, etc, most of the researchers~\citep{guo-etal-2019-densely,ribeiro-etal-2019-enhancing,beck-etal-2018-graph,damonte-cohen-2019-structural,koncel-kedziorski-etal-2019-text} propose to convert it to a multi-relational graph, which can be learned by relational GNN in Section. \ref{subsec:graph-representation-learning-rgnn} and Section. \ref{subsec:graph-representation-learning-advanced-rgnn}. 

As defined before, the multi-relational graph is denoted as $\mathcal{G}(\mathcal{V}, \mathcal{E}, \mathcal{T}, \mathcal{R})$, s.t. $|\mathcal{T}| = 1$ and $|\mathcal{R}| >=  1$. 
To get the multi-relational graph, technically, they ignore node types (i.e., project the nodes to the unified embedding space regardless of original nodes or relational nodes). As for edges, they assign the initial edges with the type "default". For each edge $e_{i, j}$, they add a reverse edge $e_{j, i}$ with type "reverse". Besides, for each node $v_i$, they add the self-loops with the type "self". Thus the converted graph is the multi-relational graph with $|E| = 3$ and $|V|=1$.


%




\subsubsection{Multi-relational Graph Neural Networks}
\label{subsec:graph-representation-learning-rgnn}
The multi-relational GNN is the extension of classic GNN for multi-relational graphs, which has the same node type but different edge types. They are originally introduced to encode relation-specific graphs such as knowledge graphs~\citep{schlichtkrull2018modeling,malaviya2020commonsense} and parsing graphs~\citep{beck-etal-2018-graph,song2019semantic}, which have complicated relationships between nodes with the same type. Generally, most multi-relational GNNs employ type-specific parameters to model the relations individually. In this subsection, we will introduce the classic relational GCN (R-GCN)~\citep{schlichtkrull2018modeling}, relational GGNN (R-GGNN)~\citep{beck-etal-2018-graph} and relational GAT (R-GAT)~\citep{wang_relational_2020,wang-etal-2020-knowledge-graph}.


\paragraph{R-GCN}
The R-GCN~\citep{schlichtkrull2018modeling} is explicitly developed to handle highly multi-relational graphs, especially knowledge bases. The R-GCN is a natural extension of the message-passing GCN framework~\citep{gilmer2017neural} which operates on local graph neighborhoods. They group the incoming nodes according to the label types and then apply messaging passing separately. 
Thus, the aggregation of node $v_i$'s immediate neighbor nodes is defined as



\begin{equation}
    \mathbf{h}^{(k)}_i = \sigma(\sum_{r \in \mathcal{E}} \sum_{v_j \in N_{r}(v_i)} {\frac{1}{c_{i, r}} {\mathbf{W}^{(k)}_{r} \mathbf{h}^{(k-1)}_{j}} } + \mathbf{W}^{(k)}_{0}\mathbf{h}^{(k-1)}_{i}),
\end{equation}
where $\mathbf{W}^{(k)}_{r} \in \mathbb{R}^{d \times d}, \mathbf{W}^{(k)}_{0} \in \mathbb{R}^{d \times d}$ are trainable parameters, $N_{r}(v_i)$ is the neighborhoods of node $v_i$ with relation $r \in \mathcal{E}$, $c_{i, r}$ is the problem-specific normalization scalar such as $|N_{r}(v_i)|$, and $\sigma(\cdot)$ is the ReLU activation function. Intuitively, such a step projects neighbor nodes with different relations by relation-specific transformation to unified feature space and then accumulates them through a normalized sum. The self-connection with a special relation type is added to ensure the node itself feature can be held.


However, modeling the multi-relations using separate parameters for each relation can lead to severe over-parameterization, especially on the rare relations. Thus two regularization methods: $\textit{basis}$ and $\textit{basis-diagonal-}$decomposition are proposed to address this issue. Firstly, for \textit{basis} decomposition, each relation weight $\mathbf{W}_{r}^{(k)}$ is defined as follows:
\begin{equation}
    \mathbf{W}_{r}^{(k)} = \sum_{b=1}^{B}a_{rb}^{(k)}\mathbf{V}_{b}^{(k)},
    \label{eq:basis}
\end{equation}
where $\mathbf{V}_{b}^{(k)} \in \mathbb{R}^{d \times d}$ is the basis and $a_{rb}^{(k)}$ is the associated coefficients. This strategy actually regards the relation matrices as the linear combination of shared basis, which can be seen as a form of weight sharing between different relations.

In the \textit{basis-diagonal decomposition}, each $\mathbf{W}_r^{(k)}$ is defined through the direct sum over a set of low-dimensional matrices as
\begin{equation}
    \mathbf{W}^{(k)}_{r} = \bigoplus_{b=1}^{B}\mathbf{Q}_{br}^{(k)},
\end{equation}
where $\mathbf{Q}_{br}^{(k)} \in \mathbb{R}^{d/B \times d/B}$ is the low-dimensional matrix. Thereby, the $\mathbf{W}_{r}^{(k)}$ is represented by a set of sub matrices as $\textit{diag}(\mathbf{Q}_{1r}^{(k)}, \mathbf{Q}_{2r}^{(k)}, ..., \mathbf{Q}_{Br}^{(k)})$. This strategy can be seen as a matrix sparsity constraint. It holds the hypothesis that the latent features can be represented by sets of variables that are more tightly coupled within groups than across groups.

There are also some other GCN-based multi-relational graph neural networks for different purposes. For example, Directed-GCN~\citep{marcheggiani2017encoding} is developed to exploit the syntactic graph, which has massive and unbalanced relations. The basic idea of incorporating edge-type information is similar to the R-GCN~\citep{schlichtkrull2018modeling}, but they solve the over-parameterization issue by sharing projection matrix weights for all edges with the same directions but only keeping the relation-specific biases. The other example is weighted-GCN~\citep{shang2019end}, which adopt relation-specific transformation to learn relational information. The weighted-GCN learns the weight score for each relation type end-to-end and inject it into the GCN framework. In this way, the weighted-GCN is capable of controlling how much information each type contributes to the aggregation procedure. As a combination model, the Comp-GCN~\citep{vashishth2019composition} generalizes several of the existing multi-relational GCN methods (i.e., R-GCN~\citep{schlichtkrull2018modeling}, Weighted-GCN~\citep{shang2019end} and Directed-GCN~\citep{marcheggiani2017encoding}) and jointly learn the nodes and relations representation.

\paragraph{R-GGNN}
The relational GGNN~\citep{beck-etal-2018-graph} is originally developed for the graph-to-sequence problem. It is capable of capturing long-distance relations. Similarly to R-GCN, R-GGNN uses relation-specific weights to capture relation-specific correlations between nodes better. Thus, the propagation process of R-GGNN can be summarized as
\begin{equation}
\begin{split}
    \mathbf{r}_i^{(k)} &= \sigma(\sum_{v_j \in N(v_i)} \frac{1}{c_{v_i, r}} \mathbf{W}^{r}_{\phi(e_{i, j})} \mathbf{h}_{j}^{(k-1)} + \mathbf{b}^{r}_{\phi(e_{i, j})}), \\
    \mathbf{z}_i^{(k)} &= \sigma(\sum_{v_j \in N(v_i)} \frac{1}{c_{v_i, z}} \mathbf{W}^{z}_{\phi(e_{i, j})} \mathbf{h}_{j}^{(k-1)} + \mathbf{b}^{z}_{\phi(e_{i, j})}), \\
    \tilde{h}_i^{(k)} &= \rho(\sum_{v_j \in N(v_i)} \frac{1}{c_{v_i}} \mathbf{W}_{\phi(e_{i, j})}(\mathbf{r}_j^{(k)} \odot \mathbf{h}_i^{(k-1)}) + \mathbf{b}_{\phi(e_{i, j})}),\\
    \mathbf{h}^{(k)}_i &= (1 - \mathbf{z}^{(k)}_{i}) \odot \mathbf{h}^{(k-1)}_{i} + \mathbf{z}^{(k)}_{i} \odot \tilde{h}^{(k)}_i,
\end{split}
\end{equation}
where $\mathbf{W}^{r/z/\cdot}_{\phi(e_{i, j})} \in \mathbb{R}^{d \times d}, \mathbf{b}^{r/z/\cdot}_{\phi(e_{i, j})}$ are trainable relation-specific parameters, $\sigma(\cdot)$ is the sigmoid function, $c_{v_i, r/z/\cdot} = |N(v_i)|$, and $\rho(\cdot)$ is the non-linear activation function such as tanh and ReLU.

\paragraph{R-GAT}

\citet{wang_relational_2020,wang-etal-2020-knowledge-graph} propose to extend the classic GAT to fit the multi-relational graphs. In this section, we will discuss two R-GAT variants. Intuitively, neighbor nodes with different relations should have different influences.


\citet{wang_relational_2020} propose to extend the homogeneous GAT with additional relational heads. Technically, they propose the relational node representation as
\begin{equation}
    \mathbf{h}^{(k), m}_{i, rel} = \sum_{v_j \in N(v_i)} \beta_{ij}^{(k), m} \mathbf{W}^{(k), m}\mathbf{h}^{(k-1)}_{j}
\end{equation}
where $m \in [1, M]$ is the $m-$th head and $\beta_{ij}^{(k), m}$ is the corresponding attention score for relation head $m$, which is calculated as
\begin{equation}
\begin{split}
    s_{ij}^{(k), m} &= f(\mathbf{e}_{i, j}),\\
    \beta_{ij}^{(k), m} &= \text{softmax}_{j}(s_{ij}^{(k), m}),
\end{split}
\end{equation}
where $s_{ij}^{(k), m}$ is the similarity between node $v_i$ and $v_j$, and $f(\cdot): \mathbb{R}^{d^{k}} \rightarrow \mathbb{R}$ is the multi-layer transformation (MLP) with non-linearity. The relational representation of node $v_i$ is the concatenation with linear transformation of $M$ heads' results as:
\begin{equation}
    \mathbf{h}^{(k)}_{i, rel} = g(||_{m=1}^{M}\mathbf{h}^{(k), m}_{i, rel}),
\end{equation}
where $||$ denotes the vector concatenation operation and $g(\cdot): \mathbb{R}^{m \times d^{k}} \rightarrow \mathbb{R}^{d^{k}}$ is the liner projection. Thus, the final node representation is the combination of $\mathbf{h}_{i, rel}^{(k)}$ and $\mathbf{h}^{(k)}_{i, att}$ as follows:
\begin{equation}
    \mathbf{h}^{(k)}_{i} = \sigma(\mathbf{W}^{(k)}(\mathbf{h}^{(k)}_{i, rel}||\mathbf{h}^{(k)}_{i, att}) + \mathbf{b}^{(k)}),
\end{equation}
where $\mathbf{W}^{(k)} \in \mathbb{R}^{d \times d}, \mathbf{b}^{(k)} \in \mathbb{R}^{d}$ are trainable parameters, $\sigma(\cdot)$ is the ReLU activation function.

Unlike the work by \citet{wang_relational_2020}, which learn and fuse the relation-specific node embedding regarding each type of edges,  \citet{wang-etal-2020-knowledge-graph} develops a relation-aware attention mechanism to calculate the attention score $\alpha_{ij}^{(k), m}$ as
\begin{equation}
    \begin{split}
        \alpha_{ij}^{(k), m} &= \text{softmax}_j(s_{ij}^{(k), m}), \\
        s_{ij}^{(k), m} &= \sigma(f^{(k), m}([\mathbf{W}^{(k), m}\mathbf{h}^{(k-1)}_{i};\mathbf{W}^{(k), m}\mathbf{h}^{(k-1)}_{j};\mathbf{e}_{i,j}^{(k-1)}])),
    \end{split}
\end{equation}
 where $\mathbf{W}^{(k), m} \in \mathbb{R}^{d \times d}$ is the learnable matrix, and $f(\cdot)^{(k), m}: \mathbb{R}^{3 \times d} \rightarrow \mathbb{R}$ is the single linear projection layer. 
 They learn a global representation for each relation type $r = \phi(e_{i, j}) \in \mathcal{R}$. Technically, for all edges with type $r \in \mathcal{R}$, two node sets $S_{r}$ and $T_{r}$. $S_{r}$ are collected regarding the source and target node set of relation $r$, respectively. Thus the edge type embedding $\mathbf{t}_{r}$ can be calculated by:

 \begin{equation}
     \mathbf{e}_{r} = \lvert \frac{\sum_{o \in S_r}\mathbf{W}\mathbf{h}_o}{|S_r|} - \frac{\sum_{o \in T_r}\mathbf{W}\mathbf{h}_o}{|T_r|} \rvert.
 \end{equation}
Thus the edge representation is the absolute difference between mean vectors of source and target nodes connected by edges whose type are $r$.

\paragraph{Gating Mechanism}
The multi-relational graph neural networks also face the  over-smoothing problem when stacking several layers to exploit implicit correlations between distant neighbors (i.e., not directly connected)~\citep{tu-etal-2019-multi}. To solve this, the gating mechanism, which combines the nodes' input features and aggregated features by gates, is introduced to the multi-relational graph neural network~\citep{tang2020multi,de2018question,tu-etal-2019-multi,DBLP:conf/naacl/CaoFT19}. Intuitively, the gating mechanism can be regarded as a trade-off between the original signals and the learned information. It regulates how much of the update message that are propagated to the next step, thus preventing the model from thoroughly overwriting the past information.
Here we introduce the gating mechanism by taking the classic R-GCN~\citep{schlichtkrull2018modeling} as an example, which actually can be extended to arbitrary variants. 

We denote the representation before activation $\sigma(\cdot)$ as
\begin{equation}
    \mathbf{u}^{(k)}_{i} = f^{(k)}(\mathbf{h}^{(k-1)}_{i}),
\end{equation}
where $f$ denotes to the aggregation function. Ultimately, the final representation of node $i$'s representation is a gated combination of the previous embedding $\mathbf{h}^k_{i}$ and GNN output representation $\sigma(\mathbf{u}^k_{i})$ as:
\begin{equation}
    \mathbf{h}^{(k)}_{i} = \sigma(\mathbf{u}^{(k)}_{i}) \odot \mathbf{g}^{(k)}_{i} + \mathbf{h}^{(k-1)}_{i} \odot (1 - \mathbf{g}^{(k-1)}_{i})
\end{equation}
where $\mathbf{g}^k_{i}$ is the gating vectors, and $\sigma(\cdot)$ is often the $\textit{tanh}(\cdot)$ function. The gating vectors are calculated by both the inputs and outputs as follows:
\begin{equation}
    g^{(k)}_{i} = \sigma (f^{(k)}([\mathbf{u}^{(k)}_{i},\mathbf{h}^{(k-1)}_{i}]))
\end{equation}
where $\sigma$ is the sigmoid activation function, and $f^{(k)}(\cdot): \mathbb{R}^{2d} \rightarrow \mathbb{R}$ is the linear transformation. We repeat calculating $g^{(k)}_{i}$ for $d$ times to get the gating vector $\mathbf{g}^{(k)}_{i}$.

\subsubsection{Graph Transformer}
\label{subsec:graph-representation-learning-advanced-rgnn}

The transformer architecture~\citep{vaswani2017attention} has achieved great success in NLP fields. Roughly speaking, the transformer's self-attention mechanism is a special procedure of fully connected implicit graph learning, as we discussed in sec. \ref{sec:graph-construction-structure-aware}, thus bridging the concept of GNN and transformer. However, the traditional transformer fails to leverage structure information. Inspired by GAT~\citep{velivckovic2017graph}, which combines the message passing with attention mechanism, much literature incorporates structured information to the transformer (we name it as graph transformer) by developing structure-aware self-attention mechanism~\citep{yao2020heterogeneous,levi1942finite,xiao-etal-2019-lattice,zhu-etal-2019-modeling,cai2020graph,wang2020amr}. In this section, we will discuss the techniques of structure-aware self-attention for the multi-relational graph.


As a preliminary knowledge, here we give a brief review of self-attention. To make it clear, we omit the multi-head mechanism and only present the self-attention function. Formally, we denote the input of self-attention as $\mathbf{Q} = \{\mathbf{q}_1, \mathbf{q}_2, ..., \mathbf{q}_m\} \in \mathbb{R}^{m \times d^{q}}, \mathbf{K} = \{\mathbf{k}_1, \mathbf{k}_2, ..., \mathbf{k}_n\} \in \mathbb{R}^{n \times d^{k}}, \mathbf{V} = \{\mathbf{v}_1, \mathbf{v}_2, ..., \mathbf{v}_n\} \in \mathbb{R}^{n \times d^{v}}$. Then the output representation $\mathbf{z}_i$ is calculated as
\begin{align}
    \mathbf{z}_i &= \text{Attention}(\mathbf{q}_{i}, \mathbf{K}, \mathbf{V}) = \sum_{j=1}^{n} \alpha_{i, j} \mathbf{W}^{v} \mathbf{v}_{j} \label{eq:transformer-aggregation}\\
    \alpha_{i, j} &= \text{softmax}_{j}(u_{i, j}) \\
    u_{i, j} &= \frac{ (\mathbf{W}^q \mathbf{q}_{i})^T (\mathbf{W}^k \mathbf{k}_{j}) }{\sqrt{d}} \label{eq:self-attn}
\end{align}
where $\mathbf{W}^{q} \in \mathbb{R}^{d \times d^{q}}, \mathbf{W}^{k} \in \mathbb{R}^{d \times d^{k}}, \mathbf{W}^{v} \in \mathbb{R}^{d \times d^{v}}$ are trainable parameters, and $d$ is the model dimension. Note that for graph transformer, the query, key and value all refer to the nodes' embedding:, namely, $\mathbf{q}_i = \mathbf{k}_i = \mathbf{v}_i = \mathbf{h}_i$. Thus, we will only use $\mathbf{h}_i$ to represent query, key and value considering simplification in the following contents.

There are various graph transformers for relation graphs that incorporate the structure knowledge, which can be categorized into two classes according to the self-attention function. One class is the R-GAT-based methods which adopt relational GAT-like feature aggregation. Another class reserves the fully connected graph while incorporating the structure-aware relation information to the self-attention function.


\paragraph{R-GAT Based Graph Transformer.}
The GAT-based graph transformer~\citep{yao2020heterogeneous} adopts the GAT-like feature aggregation, which leverages the graph connectivity inductive bias. Technically, they first aggregate neighbors with type-specific aggregation step and then fuse them through feed-forward layer as follows:
\begin{equation}
\begin{split}
     \mathbf{z}_i^{r,(k)} &=  \sum_{v_j \in N_{r}(v_i)} \alpha_{i, j}^k \mathbf{W}^{v, (k)} \mathbf{h}_{j}^{(k-1)}, r \in \mathcal{E}\\
     \mathbf{h}^{(k)}_i &= \text{FFN}^{(k)}(\mathbf{W}^{O, (k)} [\mathbf{z}_i^{R_1, (k)},...,\mathbf{z}_i^{R_q, (k)}]),
\end{split}
\end{equation}
where $\text{FFN}^{(k)}(\cdot)$ denotes the feed-forward layer in transformer~\citep{vaswani2017attention}, and $\alpha_{i, j}$ denotes the dot-product score in eq. \ref{eq:self-attn}. 

To incorporate the bi-directional information, \citet{wang2020amr} learns forward and backward aggregation representation in graph transformer. Specifically, given the backward and forward features (i.e., $\mathbf{h}_{i, \vdash}$ and $\mathbf{h}_{i, \dashv}$) for node $v_i$, the backward aggregated feature $\mathbf{z}_{i, \dashv}^{(k)}$ for node $v_i$ is formulated by:
\begin{equation}
    \begin{split}
        \mathbf{z}_{i, \dashv}^{(k)} &= \sum_{v_j \in N_{\dashv}(v_i)} \alpha_{i, j} \mathbf{W}^{v, (k)} \mathbf{a}^{(k)}_{i, j}, \\
        \mathbf{a}^{(k)}_{i, j} &= f^{(k)}([\mathbf{h}_{i, \vdash},\mathbf{e}_{i, j};\mathbf{h}_{j, \dashv}]),
    \end{split}
\end{equation}
where $f^{(k)}(\cdot): \mathbb{R}^{3 \times d \rightarrow \mathbb{R}^{d}}$ is the linear transformation, and $\alpha_{i, j}$ is the softmax score of incoming neighbors' dot-production results $u_{i, j}$ which can be formulated by:
\begin{equation}
\begin{split}
    u_{i, j} = \frac{ (\mathbf{W}^{q, (k)} \mathbf{h}_{i, \dashv})^T (\mathbf{W}^{k, (k)} \mathbf{a}^{(k)}_{i, j}) }{\sqrt{d}}.
\end{split}
\end{equation}
Then they adopt the gating mechanism to fuse bidirectional aggregated features to get the packaged node representation:
\begin{equation}
\begin{split}
    \mathbf{g}^{(k)}_{i} &= \sigma(f'^{k}([\mathbf{z}^{(k)}_{i, \vdash};\mathbf{z}^{(k)}_{i, \dashv}])), \\
    \mathbf{p}^{(k)}_{i} &= \mathbf{g}^{(k)}_i \odot \mathbf{z}^{(k)}_{i, \vdash} + (1 - \mathbf{g}^{(k)}_{i}) \odot \mathbf{z}^{(k)}_{i, \dashv}
\end{split}
\end{equation}
where $f'(\cdot): \mathbb{R}^{2 \times d} \rightarrow \mathbb{R}^{d}$, and $\sigma(\cdot)$ is the sigmoid activation function. They calculate the the forward and backward node representation based on the packaged representation, respectively:
\begin{equation}
\begin{split}
    [\mathbf{o}^{(k)}_{i, \vdash}, \mathbf{o}^{(k)}_{i, \dashv}] &= \text{FFN}^{(k)}(\mathbf{p}^{(k)}_{i}), \\
    \mathbf{h}^{(k)}_{i, \vdash} &= \text{LayerNorm}^{(k)}(\mathbf{o}^{(k)}_{i, \vdash} + \mathbf{h}^{(k-1)}_{i, \vdash}), \\
    \mathbf{h}^{(k)}_{i, \dashv} &= \text{LayerNorm}^{(k)}(\mathbf{o}^{(k)}_{i, \dashv} + \mathbf{h}^{(k-1)}_{i, \dashv}), 
\end{split}
\end{equation}
where $\text{FFN}(\cdot): \mathbb{R}^{d} \rightarrow \mathbb{R}^{2 \times d}$ is the feed-forward function, and $\text{LayerNorm}(\cdot)$ is the layer normalization~\citep{ba2016layer}. The final node representation is the concatenation of the last layer $K$'s bidirectional representations:
\begin{equation}
    \mathbf{h}^{(K)}_{i} = f''^{K}([\mathbf{h}^{(K)}_{i, \vdash}, \mathbf{h}^{(K)}_{i, \dashv}]),
\end{equation}
where $f''^{(K)}(\cdot): \mathbb{R}^{2 \times d} \rightarrow \mathbb{R}^{d}$ is the linear transformation.

\paragraph{Structure-aware Self-attention Based Graph Transformer.} Unlike the R-GAT-based graph transformer, which purely relies on the given graph structure as connectivity, the structure-aware self-attention-based graph transformer reserves the original self-attention architecture, allowing non-neighbor nodes' communication. We will firstly discuss the structure-aware self-attention mechanism and then present its unique edge embedding representation.

\citet{shaw-etal-2018-self} firstly attempts to model the relative relations between words (nodes) in the neural machine translation task. Technically, they consider the relation embedding when calculating node-wise similarity in eq. \ref{eq:self-attn} as follows:
\begin{equation}
    u_{i, j}^{(k)} = \frac{ (\mathbf{W}^{q, (k)} \mathbf{h}^{(k-1)}_{i})^T (\mathbf{W}^{k, (k)} \mathbf{h}^{(k-1)}_{j}) + (\mathbf{W}^{q, (k)} \mathbf{h}^{(k-1)}_{i})^T \mathbf{e}_{i, j} }{\sqrt{d}}.
\end{equation}

Motivated by \citet{shaw-etal-2018-self}, \citet{xiao-etal-2019-lattice,zhu-etal-2019-modeling} propose to extend the conventional self-attention architecture to explicitly encode the relational embedding between nodes pairs in the latent space as
\begin{equation}
    \begin{split}
        u_{i, j}^{(k)} &= \frac{ (\mathbf{W}^{q, (k)} \mathbf{h}^{(k-1)}_{i})^T (\mathbf{W}^{k, (k)} \mathbf{h}^{(k-1)}_{j} + \mathbf{W}^{r, (k)}\mathbf{e}_{i, j}) }{\sqrt{d}}, \\
         \mathbf{h}_i^{(k)} &= \sum_{j=1}^{n} \alpha_{i, j}^k (\mathbf{W}^{v, (k)} \mathbf{h}_{j}^{(k-1)} + \mathbf{W}^{f, (k)}\mathbf{e}_{i, j}).
    \end{split}
\end{equation}

To adopt the bidirectional relations, \citet{cai2020graph} extends the traditional self-attention as follows:
\begin{equation}
    u_{i, j}^{(k)} = \frac{[\mathbf{W}^{q, (k)}(\mathbf{h}^{(k-1)}_{i} + \mathbf{e}_{i, j})]^T [\mathbf{W}^{k, (k)} (\mathbf{h}^{(k-1)}_{j} + \mathbf{e}_{j, i})]}{\sqrt{d}}.
\end{equation}

Given the learnt attention for each relation, edge embedding representation is the next critical step for incorporating the structure-information.
\citet{shaw-etal-2018-self} simply learns the relative position encoding w.r.t the nodes' absolute positions. Technically, they employ $2K+1$ latent labels ($[-K, K]$) and project $j - i$ to one specific label embedding for node pair $(v_i, v_j)$ to fetch the edge embedding $\mathbf{e}_{i, j}$. \citet{xiao-etal-2019-lattice} adopts the similar idea as \citet{shaw-etal-2018-self}. They define a relative position embedding table and fetch the edge embedding by looking up from it.

\citet{zhu-etal-2019-modeling,cai2020graph} learn the edge representation $\mathbf{e}_{i, j}$ by the path from node $v_i$ to node $v_j$. For \citet{zhu-etal-2019-modeling}, the natural way is to view the path as a string, which is added to the vocabulary to vectorize it. Other ways are further proposed to learn from labels' embedding along the path, such as 1) taking average, 2) taking sum, 3) encoding them using self-attention, and 4) encoding them using CNN filters.  \citet{cai2020graph} propose the shortest path based relation encoder. Concretely, they firstly fetch the labels' embedding sequence along the path. Then they employ the bi-directional GRUs for sequence encoding. The last hidden states of the forward and backward GRU networks are finally concatenated to represent the relation embedding $\mathbf{e}_{i, j}$.

\subsection{Graph Neural Networks for Heterogeneous Graph}
\label{subsec:graph-representation-learning-heterogeneousgnn}

In practice, many graphs have various node and edge types, such as knowledge graph, AMR graph, etc., which are called heterogeneous graphs. Formally, for a graph $\mathcal{G}$, s.t. $|\mathcal{T}| > 1$ or $|\mathcal{R}| > 1$, it is called heterogeneous graph. Beside transforming the heterogeneous to relation graphs, as introduced in the previous subsection, sometimes it is required to fully leverage the type information for both nodes and edges~\citep{hu2020heterogeneous,fan2019metapath,feng2020scalable,wang-etal-2020-heterogeneous,linmei-etal-2019-heterogeneous,zhang2019heterogeneous}. Thus, in Section. \ref{subsec:levi}, we first introduce a pre-processing technique for heterogeneous graph. Then, in Section. \ref{subsec:meta-heteronebous gnn} and \ref{subsec:e-gnn heteronenous gnn}, we will introduce two typical graph representation learning methods specially for heterogeneous graphs.


\subsubsection{Levi Graph Transformation}
\label{subsec:levi}

Since most existing GNN methods are only designed for homogeneous conditions and there is a massive computation burden when dealing with lots of edge types~\citep{beck-etal-2018-graph} (e.g. an AMR graph may contain more than 100 edge types), it is typical to effectively to treat the edges as nodes in the heterogeneous graphs~\citep{beck-etal-2018-graph,xu-etal-2018-exploiting,sun-etal-2019-joint,guo-etal-2019-densely}.

\begin{figure}[ht!]
\centering
\vspace{-2mm}
\includegraphics[width=12.0cm]{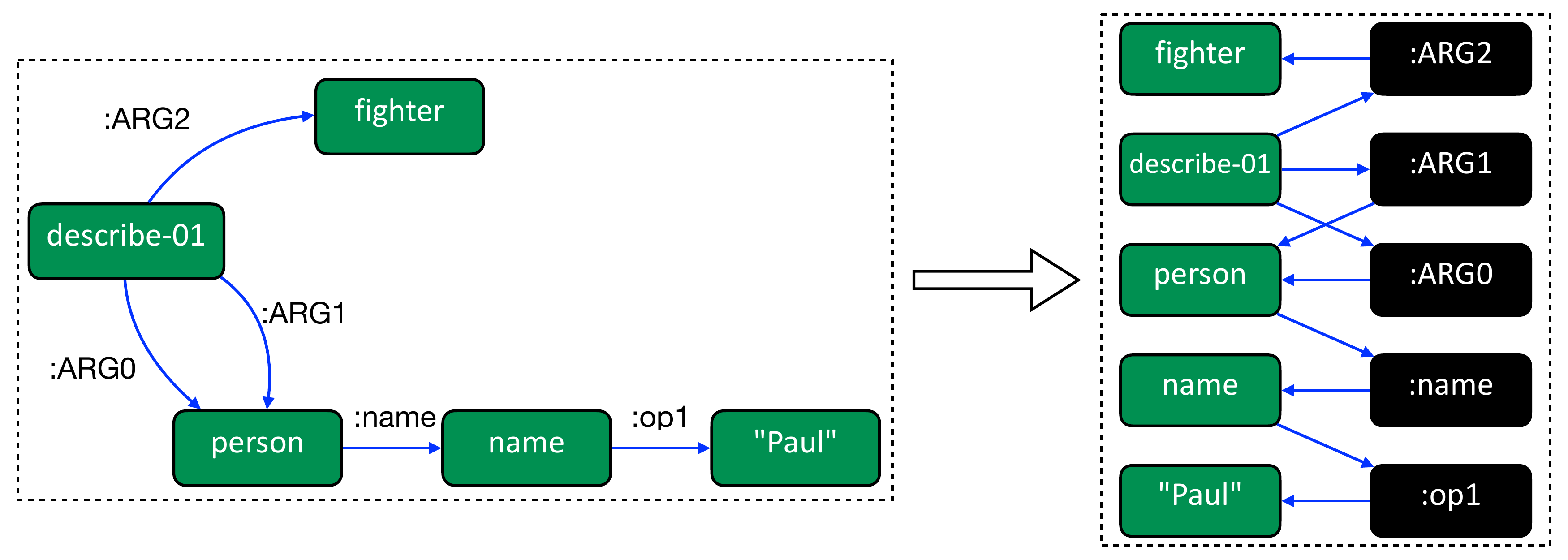}
\vspace{-4mm}
\caption{An example for transforming AMR graph to Levi-graph.}
\label{fig:levi-graph-example}
\vspace{-0mm}
\end{figure}

One of the important graph transformation techniques is \textit{Levi Graph Transformation}.
Technically, for each edge $e_{i,j}$ with edge label $\phi(e_{i,j})$, we will create a new node $v_{e_{i,j}}$. Thus the new graph is denoted as $ \mathcal{G'}(\mathcal{V'}, \mathcal{E'}, \mathcal{T'}, \mathcal{R'})$, where the node set is $\mathcal{V'} = \mathcal{V} \cup \{v_{e_{i,j}}\}$, the node label set is $\mathcal{T'} = \mathcal{T} \cup \{\phi(e)_{i, j}\}$. We cut off the direct edge between node $v_i, v_j$ and the add two direct edges between: 1) $v_i, v_{e_{i,j}}$, and 2) $v_{e_{i,j}}, v_j$. After converting all edges in $\mathcal{E}$, the new graph $\mathcal{G'}$ will be a bipartite graph, s.t. $|\mathcal{R'}| = 1$. An example of transforming AMR graph to desired levi-graph is illustrated in Fig. \ref{fig:levi-graph-example}. The obtained graph is a simplified heterogeneous levi graph that has a single edge type but unrestricted node types, which can then be learnt by heterogeneous GNNs described in Section~\ref{subsec:graph-representation-learning-heterogeneousgnn}.

\subsubsection{Meta-path based Heterogeneous GNN}
\label{subsec:meta-heteronebous gnn}
Meta-path, a composite relation connecting two objects, is a widely used structure to capture the semantics. Take movie data IMDB for example, there are three types of nodes, including movie, actor, and director. The meta-path $Movie \rightarrow Actor \rightarrow Movie$, which covers two movie sets and one actor, describes the co-actor relations. Thus different relations between nodes in the heterogeneous graph can be easily revealed by meta-paths.

\begin{figure}[ht!]
\centering
\vspace{-2mm}
\includegraphics[width=12.0cm]{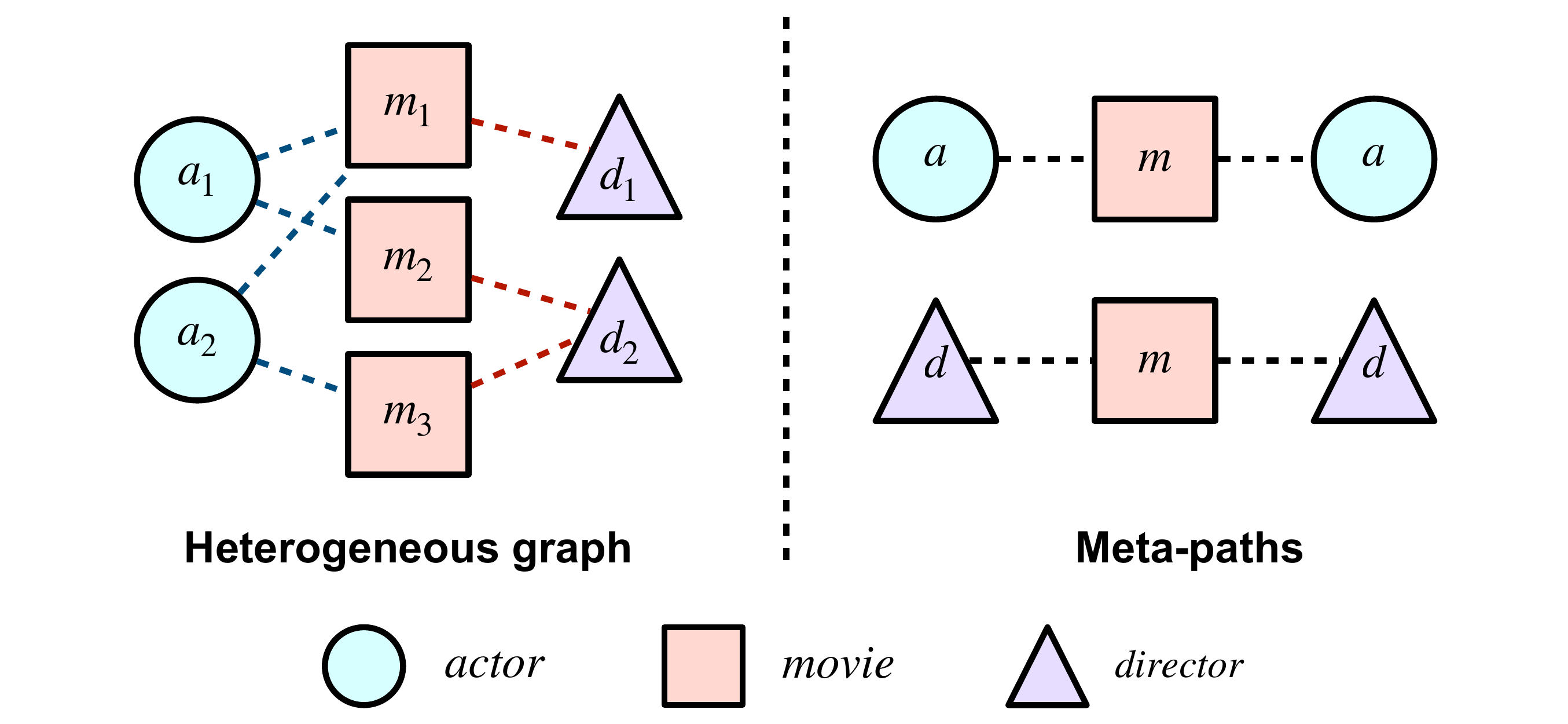}
\vspace{-4mm}
\caption{An example of meta-path based heterogeneous graph.}
\label{fig:metapath-graph-example}
\vspace{-0mm}
\end{figure}

First, we provide the meta-level (i.e., schema-level) description of a heterogeneous graph for better understanding. We follow the setting of heterogeneous information network (HIN)~\citep{sun2011pathsim} and give the concept of Network Schema. The network schema is a meta template for the heterogeneous graph $\mathcal{G}(\mathcal{V}, \mathcal{E})$ with the node type mapping: $\mathcal{V} \rightarrow \mathcal{T}$ and edge type mapping: $\mathcal{E} \rightarrow \mathcal{R}$. We denote it as $\mathcal{M}_{\mathcal{G}}(\mathcal{T}, \mathcal{R})$. A meta path is a path on the network schema denoted as $\Phi = T_1 \overset{R_1}{\rightarrow} T_2 \overset{R_2}{\rightarrow} ... \overset{R_l}{\rightarrow} T_{l+1}$, where $T_i \in \mathcal{T}$ is the schema's node and $R_i \in \mathcal{R}$ is the corresponding relation node. What's more, we denote the meta-path set as $\{\Phi_1, \Phi_2, ..., \Phi_p\}$. For each node $T_i$ on the meta-path $\Phi_j$, we denote it as $T_{i}^{\Phi_j}$. Then we combine the network schema with the concrete heterogeneous graph. For each node $v_i$ in the heterogeneous graph and one meta-path $\Phi_j$, we define the meta-path-based neighbors as $N_{\Phi_j}(v_i)$, which contains all nodes including itself linked by meta-path $\Phi_j$. An example of meta-path based heterogeneous graph is shown in Fig. \ref{fig:metapath-graph-example}. Conceptually, the neighbor set can have multi-hop nodes depending on the length of the meta-path.

Most meta-path-based GNN methods adopt the attention-based aggregation strategy~\citep{wang-etal-2020-heterogeneous,fan2019metapath}. Technically, they can be generally divided into two stages. Firstly, they aggregate the neighborhoods along each meta-paths, which can be named as ``node-level aggregation". After this, the nodes receive neighbors' information via different meta-path. Next, they apply meta-path level attention to learn the semantic impact of different meta-path. In this way, they can learn the optimal combination of neighbors connected by multiple meta-paths. In the following, we will discuss two typical heterogeneous GNN models~\citep{wang-etal-2020-heterogeneous,fan2019metapath}.


\paragraph{HAN}~\citep{wang2019heterogeneous}
Due to the heterogeneity of nodes in graphs, different nodes have different feature spaces, which brings a big challenge for GNN to handle various initial node embedding. To tackle this issue, the type-specific transformation is adopted to project various nodes to a unified feature space as follows:
\begin{equation}
    \mathbf{h}'_{i} = \mathbf{W}_{\tau(v_i)} \mathbf{h}_{i}. \label{eq:feature-prj}
\end{equation}
We overwrite the notation $\mathbf{h}_{i}$ to denote the transformed node embedding in HAN's discussion.

\begin{itemize}
    \item \textbf{Node-level Aggregation.} For the node $v_i$ and its' neighbor node set $N_{\Phi_k}(v_i)$ on the meta-path $\Phi_k$, the aggregated feature of node $v_i$ can be represented by:
    \begin{equation}
    \begin{split}
         \mathbf{z}_{i, \Phi_k} &= \sigma(\sum_{v_j \in N_{\Phi_k}(v_i)} \alpha^{\Phi_k}_{i, j} \mathbf{h}_{j})\\
         \alpha^{\Phi_k}_{i, j} &= \text{softmax}_j(u^{\Phi_{k}}_{i, j}) \\
         u^{\Phi_k}_{i, j} &= \text{Attention}(\mathbf{h}_{i}, \mathbf{h}_j; \Phi_k) = \sigma(\mathbf{W}[\mathbf{h}_{i},\mathbf{h}_{j}]),
    \end{split}
    \end{equation}
    where $\mathbf{W} \in \mathbb{R}^{1 \times 2d}$ is the trainable parameter. To make the training process more stable, they further extend the attention by the multi-head mechanism. The final representation of $\mathbf{z}_{i, \Phi_j}$ is the concatenation of $L$ heads' results. Given p meta-paths, we can obtain the aggregated embedding via the previous nodel-level aggregation step as $\{\mathbf{Z}_{\Phi_1}, ..., \mathbf{Z}_{\Phi_p}\}$, where $\mathbf{Z}_{\Phi_j}$ is the collection of all nodes' representation for meta-path $\Phi_j$.

    \item \textbf{Meta-path Level Aggregation.} Generally, different meta-path conveys different semantic information. To this end, they aim to learn the importance of each meta-path as:
    \begin{equation}
        (\beta_{\Phi_1}, ..., \beta_{\Phi_p}) = \text{Meta\_Attn}(\mathbf{Z}_{\Phi_1}, ..., \mathbf{Z}_{\Phi_p}),
    \end{equation}
    where $\beta_{\Phi_j}$ denotes the learned importance score for meta-path $\Phi_j$, and $\text{Meta\_Attn}()$ is the attention-based scorer. Technically, for each meta-path, they first learn the semantic-level importance for each node. Then they average them to get the meta-path level importance. It can be formulated by:
    \begin{equation}
        \begin{split}
            o_{\Phi_k} &= \frac{1}{|\mathcal{V}|} \sum_{v_i \in \mathcal{V}} \mathbf{q}^T f(\mathbf{z}_{i, \Phi_k}), \\
            \beta_{\Phi_k} &= \text{softmax}_{k} (o_{\Phi_k}),
        \end{split}
    \end{equation}
    where $f(\cdot): \mathbb{R}^{d} \rightarrow \mathbb{R}^d$ is the MLP with $\text{tanh}$ non-linearity.
    
     Finally, we can obtain the final node representation as:
    \begin{equation}
        \mathbf{z}_{i} = \sum_{k=1}^{p} \beta_{\Phi_k} \mathbf{z}_{i, \Phi_k}.
    \end{equation}
    
    
\end{itemize}

\paragraph{MEIRec}
The MEIRec~\citep{fan2019metapath} is a heterogeneous GNN-based recommend system. In the specific recommendation system condition, they restrict the heterogeneous graph's type amount and meta-path's length and propose a special heterogeneous graph neural network, particularly to fully utilize rich structure information. 
Considering that the type-specific transformation requires huge parameters when the amount of nodes is large, they propose an efficient unified embedding learning method. For each node, they fetch the terms in the vocabulary and then average them to get the vector representation. 

\begin{itemize}
    \item \textbf{Node-level Aggregation.} Unlike HAN~\citep{hu2020heterogeneous}, which collect all nodes along the meta-path as neighbors, they treat different hop of neighbors differently. Given the meta-path $\Phi_j$, they define the neighbors of node $v_i$ as $N_{\Phi_j}(v_i)^{o}, o \in [1, 2, ..., O]$ where $o$ denotes the hop. They learn the representation recursively. Take $(o)-$hop neighbors for example, for each nodes in $N_{\Phi_j}(v_i)^{o}$, they first collect the immediate-neighbors from $(o+1)$-hop neighbors and learn the representation to obtain $(o)$-hop representation. Then they repeat this procedure to obtain $(o-1)$-hop nodes' representation. Finally, $v_i$'s representation for meta-path $\Phi_j$ is generated. Formally, for $v_k \in N_{\Phi_j}(v_i)^{o}$, they define its' immediate neighbor set as $N_{\Phi_j}(v_k) \in N_{\Phi_j}(v_i)^{o+1}$. Node $v_j$'s representation $\mathbf{z}_{k, \Phi_j}$ is formulated as:
    \begin{equation}
        \mathbf{z}_{k, \Phi_j} = g(\{\mathbf{z}_{l, \Phi_j}, v_l \in N_{\Phi_j}(v_k) \}),
    \end{equation}
    where $g(\cdot)$ is the aggregation function. In MEIRec, it can be the average function, LSTM function, or the CNN function depend on the nodes' type. Besides, the last hop ($(O)$-hop)'s nodes are represented by initial node embedding.
    
    \item \textbf{Meta-path Level Aggregation.} Given p meta-path with the starting nodes $v_i$, we can obtain p aggregated embedding by the previous step. Then we adopt a similar procedure as node-level aggregation as follows:
    \begin{equation}
        \mathbf{z}_{i} = g(\{\mathbf{z}_{i, \Phi_j}, j \in [1, ..., p]\}),
    \end{equation}
    where $g(\cdot)$ is the aggregation function, as we discussed before.
\end{itemize}



\subsubsection{R-GNN based Heterogeneous GNN}
\label{subsec:e-gnn heteronenous gnn}
Although the meta-path is an effective tool to organize the heterogeneous graph, it requires additional domain expert knowledge. To this end, most researchers adopt a similar idea from R-GNN by using type-specific aggregation. For clarity, we name these methods as R-GNN based heterogeneous GNN and introduce several typical variants of this category in the following.



\paragraph{HGAT}
HGAT~\citep{linmei-etal-2019-heterogeneous} is proposed for encoding heterogeneous graph which contains various node types but single edge types. In other words, the edge only represents connectivity. Intuitively, for a specific node, different types of neighbors may have different relevance. To fully exploit diverse structure information, HGAT firstly focuses on global types' relevance learning (type-level learning) and then learns the representation for specific nodes (node-level learning).

\begin{itemize}
    \item \textbf{Type-level learning}. Technically, for a specific node $v_i$ and its' neighbors $N(v_i)$, HGAT get the type-specific neighbor representation as:
\begin{equation}
    \mathbf{z}_{t}^{(k)} = \sum_{v_j \in N_{t}(v_i)} \mathbf{h}_{j}^{(k-1)}, t \in \mathcal{T}.
\end{equation}
Note that we overwrite $N_{t}(v_i)$ which denotes the neighbors with node type $t$. Then they calculate the relevance of each type by attention mechanism:
\begin{equation}
    \begin{split}
        s_t &= \sigma(\mathbf{q}^T[\mathbf{h}_i^{(k-1)}, \mathbf{z}_{t}^{(k)}]), \\
        \alpha_{t} &= \frac{exp(s_t)}{\sum_{t' \in \mathcal{T}} exp(s_{t'}) },
    \end{split}
\end{equation}
where $\mathbf{q}$ is the trainable vector.

\item \textbf{Node-level learning}
Secondly, they apply R-GCN~\citep{schlichtkrull2018modeling} like aggregation procedure for a different type of nodes. Formally, for the node $v_i$ and the type relevance scores $\{\alpha_t\}, t \in \mathcal{T}$, HGAT calculate each neighbors' attention score as follows:
\begin{equation}
\begin{split}
    b_{i, j} &= \sigma(\mathbf{q}_{1}^T \alpha_{\tau(v_j)} [\mathbf{h}_i^{(k-1)}, \mathbf{h}_j^{(k-1)}]), \\
    \beta_{i, j} &= \frac{exp(b_{i, j})}{\sum_{v_m \in N(v_i)} exp(b_{i, m}) },
\end{split}
\end{equation}
where $\mathbf{q}_{1}$ is the trainable vector. Finally, HGAT applies layer-wise heterogeneous GCN to learn the node representation, which is formulated as:
\begin{equation}
    \mathbf{h}^{(k)}_{i} = \sigma(\sum_{t \in \mathcal{T}} \sum_{v_j \in N_{t}(v_i)} \mathbf{W}^{(k)}_{t} \mathbf{h}^{(k-1)}_{j} ).
\end{equation}
\end{itemize}


\paragraph{MHGRN} 
MHGRN~\citep{feng2020scalable} is an extension of R-GCN, which can leverage multi-hop information directly on heterogeneous graphs. Generally, it borrows the idea of relation path (e.g., meta-path) to model the relation of two not k-hop connected nodes and extend the existing R-GNN to the path-based heterogeneous graph representation learning paradigm. The $K$-hop relation path between node $v_i, v_j$ is denoted as:
\begin{equation}
    \Phi_k = \{(v_i, e_{i, 1}, ..., e_{k-1, j}, v_j) | (v_i, e_{i, 1}, v_1), ..., (v_{k-1}, e_{k-1, j}) \in \mathcal{E}\}.
\end{equation}
Note that the heterogeneous graph may contain more than one k-hop relation path.  

\begin{itemize}


\item \textbf{k-hop feature aggregation}. First, to make the GNN aware of the node type, they project the nodes' initial feature to the unified embedding space by type-specific linear transformation (the same as eq. \ref{eq:feature-prj}). Considering simplification, we overwrite nodes' feature notation $\mathbf{h}$ to represent the unified features. Then given node $v_i$, they aim to aggregate $k$-hop ($k \in [1, K]$) neighbors' feature as follows:
\begin{equation}
\begin{split}
    \mathbf{z}^{\Phi_k}_{i} = \sum_{(v_j, e_{j, 1}, ..., e_{k-1, i}, v_i) \in \Phi_k} \frac{\alpha(v_j, e_{j, 1}, ..., e_{k-1, i}, v_i)}{\sum_{(v_j,..., v_i) \in \Phi_k}\alpha(v_j, ..., v_i)} \prod_{l=1}^{l=K}\prod_{o=1}^{o=K}\mathbf{W}^{l}_{r_o}\mathbf{h}_j, \\
    (1 \leq k \leq K)
\end{split}
\end{equation}
where $\mathbf{W}_{r_o}^{l}, 1 \leq l \leq K, 1 \leq o \leq K$ is the learnable matrix, $r_o$ denotes $o-$th hop's edge, $\alpha(j, v_1, ..., v_k, v_i)$ is the attention score among all k-hop paths between node $v_j$ and $v_i$. We use $\mathbf{z}_{i}$ to denote the learned embedding for node $v_i$. 

\item \textbf{Fusing different relation paths}. Next, they fuse relation paths with different length via attention mechanism:
\begin{equation}
    \mathbf{z}'_i = \sum_{k=1}^{K} \text{Attention}(\mathbf{q}, \mathbf{z}_i^{\Phi_k}) \mathbf{z}_i^{\Phi_k},
\end{equation}
where $\mathbf{q}$ is the task-specific vector (in MHGRN, it is the text-based query vector), $\text{Attention}(): \mathbb{R}^d \rightarrow \mathbb{R}$ is the normalized attention score. Note that we omit the details of $\alpha(j, v_1, ..., v_k, v_i)$ and $\text{Attention}()$ since they are task-specific functions. At last, the final representation of node $v_i$ is the shortcut connection between $\mathbf{z}_i$ and original feature $\mathbf{h}_i$ as follows:
\begin{equation}
    \mathbf{z}_i = \sigma(\mathbf{W}_1 \mathbf{z}'_i + \mathbf{W}_2 \mathbf{h}_i),
\end{equation}
where $\mathbf{W}_1, \mathbf{W}_2$ is the learnable weights.

\end{itemize}

\paragraph{HGT}
The HGT~\citep{hu2020heterogeneous} is the graph transformer for heterogeneous graphs, which build meta-relations among nodes on top of the network schema, as we discuss in the meta-path-based heterogeneous GNN paragraph. Unlike most previous works that assume the graph structure is static (i.e., time-independent), they propose a relative temporal encoding strategy to learn the time dependency.

The meta-relation is a triple based on the network schema. For each edge $e_{i, j}$ which links node $v_i$ and $v_j$, the meta-relation is defined as $\Phi_{v_i, e_{i, j} v_j} = <\tau(v_i), \phi(e_{i, j}), \tau(v_j)>$. To further represent the time-dependent relations, they add the timestamps to the start nodes when adding directed edges. Generally, the GNN is defined as:
\begin{equation}
    \mathbf{h}^{(k)}_{i}= \text{Aggregation}^{(k)}_{v_j \in N(v_i)}(\text{Attention}^{(k)}(v_i, e_{i, j}, v_j) \text{Message}^{(k)}(v_i, e_{i, j}, v_j)),
\end{equation}
where $N(v_i)$ denotes the incoming nodes. We will briefly discuss three basic meta-relation based operations: attention, message message passing, and aggregation, as well as the relative temporal encoding strategy.
\begin{itemize}
    \item \textbf{Attention operation}. The $\text{Attention}(\cdot, \cdot, \cdot)$ operation is the mutual attention that calculates the weight of two connected nodes grounded by their meta-relations. Specifically, they employ multi-head attention based on meta-relations, which is formulated as:
\begin{equation}
    \begin{split}
        &\text{Attn\_head}^{i}(v_i, e_{i, j}, v_j) = \\ &\text{f\_linear}_{\tau(v_i)}^{i}(\mathbf{h}_{i}) \mathbf{W}^{ATT}_{\phi(e_{i, j})} \text{g\_linear}_{\tau(v_j)}^{i}(\mathbf{h}_{j})^T \frac{E_{\Phi_{v_i, e_{i, j} v_j}}}{\sqrt{d}}, \\
         &\text{Attention}(v_i, e_{i, j}, v_j) = softmax_{v_j \in N(v_i)}(||_{h=1}^{H} \text{Attn\_head}^{i}(v_i, e_{i, j}, v_j) )
    \end{split}
\end{equation}
where $H$ is the number of heads, $\text{f\_linear}^i_{\tau(\cdot)}, \text{g\_linear}^i_{\tau(\cdot)}: \mathbb{R}^{d/H} \rightarrow \mathbb{R}^{d/H}$ are the node-type-specific transformation functions for source nodes and target nodes respectively, $\mathbf{W}^{ATT}_{\phi(\cdot)}$ is the edge-type-specific matrix, and $E_{\Phi_{v_i, e_{i, j} v_j}}$ is the meta-path-specific scalar weight.

\item\textbf{Message passing operation}. The $\text{Message}(\cdot)$ is the heterogeneous message passing function. Similar to the $\text{Attention}(\cdot, \cdot, \cdot)$ above, they incorporate the meta-relations into the message passing process as follows:
\begin{equation}
    \begin{split}
        \text{msg\_head}^i(v_i, e_{i, j}), v_j) &= \text{m\_linear}^i_{\tau(v_i)}(\mathbf{h}_{i}) \mathbf{W}_{\phi(e_{i, j})}^{MSG}, \\
        \text{Message}(v_i, e_{i, j}, v_j) &= ||_{h=1}^{H} \text{msg\_head}^h(v_i, e_{i, j}, v_j)
    \end{split}
\end{equation}
where $\text{m\_linear}(\cdot): \mathbb{R}^{d/H} \rightarrow \mathbb{R}^{d/H}$ is the node-type-specific transformation, and $\mathbf{W}^{MSG}_{\phi(\cdot)}$ is the edge-type-specific matrix. 

\item\textbf{Aggregation operation}. For aggregation operation, since the $\text{Attention}()$ function's results have been normalized by softmax function, they simply use average function as $\text{Aggregation}(\cdot)$. Finally, they employ meta-path-specific projection followed by residual connection to learn the final representation of each nodes as follows:
\begin{equation}
    \begin{split}
        \mathbf{z}_i^{(k)} &= \sum_{v_j \in N(v_i)}(\text{Attention}^{(k)}(v_i, e_{i, j}, v_j) \text{Message}^{(k)}(v_i, e_{i, j}, v_j)), \\
        \mathbf{h}_{i}^{(k)} &= \text{A\_linear}_{\Phi_{v_i, e_{i, j} v_j}} (\sigma(\mathbf{z}_i^{(k)})) + \mathbf{h}^{(k-1)},
    \end{split}
\end{equation}
where $\text{A\_linear}_{\Phi_{v_i, e_{i, j} v_j}}(\cdot): \mathbb{R}^d \rightarrow \mathbb{R}^d$ is the meta-relation-specific projection. 

\item\textbf{Relative Temporal Encoding}
To tackle the graph's time dependency, they propose the Relative Temporal Encoding mechanism to each node's embedding. Technically, they calculate the timestamp difference of target and source nodes as $\delta_{i, j} = T(v_j) - T(v_i)$, where $T(\cdot)$ is the timestamp of the node. Then they project the time gap to the specific embedding space. This temporal encoding is added to the source nodes' representation before GNN encoding.
\end{itemize}

\section{GNN Based Encoder-Decoder Models}
\label{sec:GNN Based Encoder-Decoder Models}

Encoder-decoder architecture is one of the most widely used machine learning framework in the NLP field, such as the Sequence-to-Sequence (Seq2Seq) models\citep{DBLP:conf/nips/SutskeverVL14,DBLP:conf/emnlp/ChoMGBBSB14}. Given the great power of GNNs for modeling graph-structured data, very recently, many research efforts have been made to develop GNN-based encoder-decoder frameworks including Graph-to-Tree~\citep{li-etal-2020-graph-tree,zhang-etal-2020-graph-tree} and Graph-to-Graph~\citep{guo2019deep,shi2020graph} models.
In this section, we will first introduce the typical Seq2Seq models, and then discuss various graph-based encoder-decoder models for various NLP tasks.



\begin{figure}[ht!]
\centering
\vspace{-2mm}
\includegraphics[width=12.0cm]{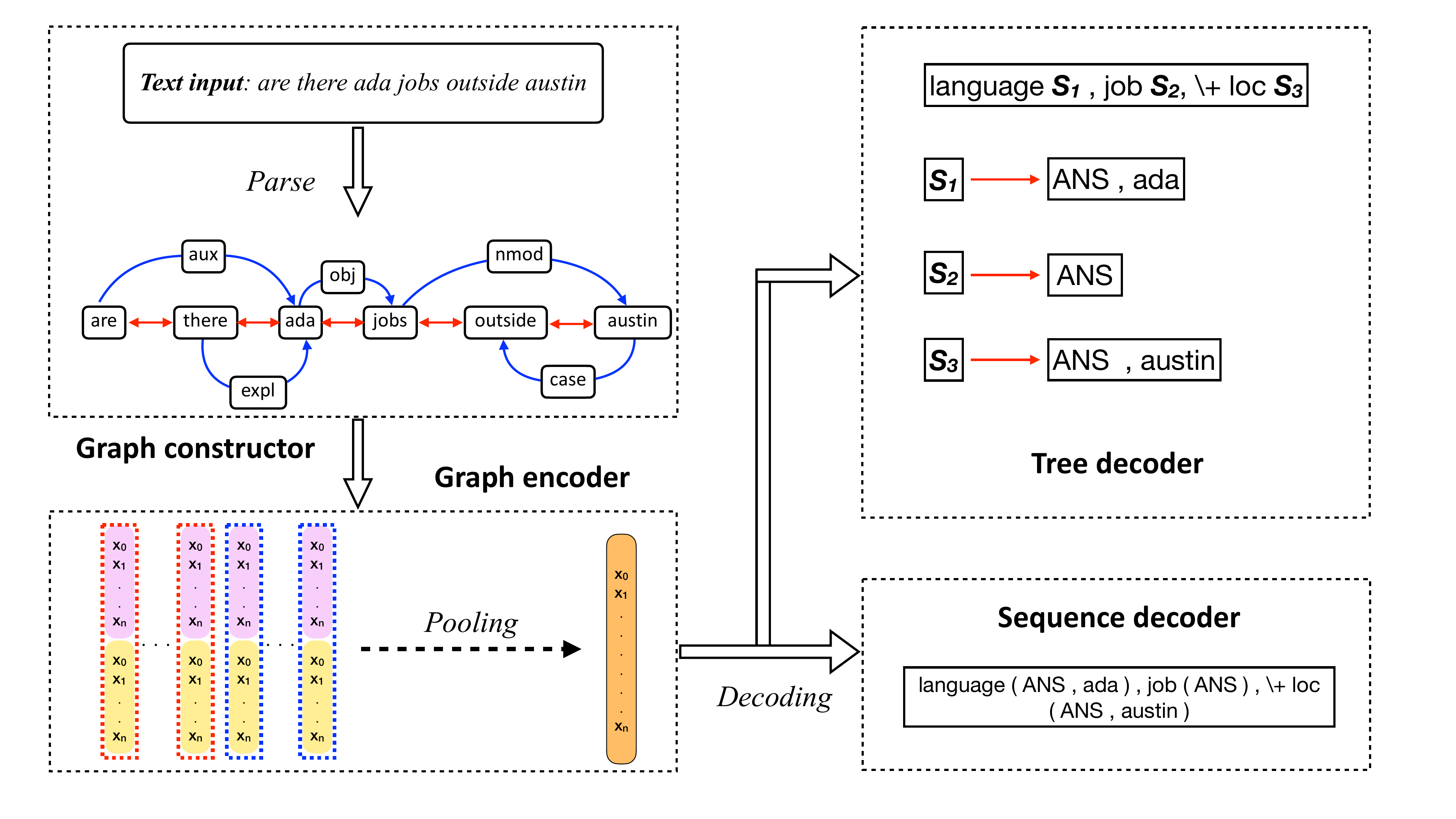}
\vspace{-8mm}
\caption{Overall architecture for graph based encoder-decoder model which contains both the Graph2Seq and Graph2Tree models. Input and output are from \textbf{JOBS640} dataset~\citep{luke2005ze} . Nodes like $\textit{\textbf{S}}_1$, $\textit{\textbf{S}}_2$ stand for sub-tree nodes, which new branches are generated from.}
\label{fig:graph-encoder-decoder-sample}
\vspace{-0mm}
\end{figure}

\subsection{Sequence-to-Sequence Models}


Sequence-to-Sequence (Seq2Seq) learning~\citep{DBLP:conf/nips/SutskeverVL14,DBLP:conf/emnlp/ChoMGBBSB14} is one of the most widely used machine learning paradigms in the NLP field.
In this section, we first give a brief overview of Seq2Seq learning and introduce some typical Seq2Seq techniques. Then we pinpoint some known limitations of Seq2Seq learning as well as its solutions, namely, incorporating more structured encoder-decoder models as alternatives to Seq2Seq models so as to encode more complex data structures.

\subsubsection{Overview}

Sequence-to-Sequence (Seq2Seq) models were originally developed by \citet{DBLP:conf/nips/SutskeverVL14} and \citet{DBLP:conf/emnlp/ChoMGBBSB14} for 
solving general sequence-to-sequence problems (e.g., machine translation).
The Seq2Seq model is an end-to-end encoder-decoder framework which learns to map a variable-length input sequence to a variable-length output sequence.
Basically, the idea is to use an RNN-based encoder to read the input sequence (i.e., one token at a time), to build up a fixed-dimensional vector representation, and then use an RNN-based decoder to generate the output sequence (i.e., one token at a time) conditioned on the encoder output vector. 
The decoder is essentially a RNN language model except that it is conditioned on the input sequence.
One of the most common Seq2Seq variants is to apply a Bidirectional LSTM encoder to encode the input sequence, and apply a LSTM decoder to decode the output sequence~\citep{DBLP:conf/nips/SutskeverVL14}.
Other Seq2Seq variants replace LSTM with Gated Recurrent Units (GRUs)~\citep{DBLP:conf/emnlp/ChoMGBBSB14}, Convolutional Neural Networks (CNNs)~\citep{gehring2017convolutional} or Transformer models~\citep{vaswani2017attention}.

Despite the promising achievements in many NLP applications such as machine translation, the original Seq2Seq models suffer a few issues such as the information bottleneck of the fixed-dimensional intermediate vector representation, and the exposure bias of cross-entropy based sequence training.
In the original Seq2Seq architecture, the intermediate vector representation becomes an information bottleneck because it summarizes the rich information of the input sequence as a fixed-dimensional embedding, which serves as the only knowledge source for the decoder to generate a high-quality output sequence.
In order to increase the learning capacity of the original Seq2Seq models, many effective techniques have been proposed.

\subsubsection{Approach}
The attention mechanism~\citep{bahdanau2015neural,luong2015effective} was developed to learn the soft alignment between the input sequence and output sequence. Specifically, at each decoding step $t$, an attention vector indicating a probability distribution over the source words is computed as
\begin{equation}\label{eq:attn}
\begin{aligned}
e^t_i &= \text{f}(\vec{h}_i, \vec{s}_t)\\
\vec{a}^t &= \text{softmax}(\vec{e}^t),
\end{aligned}
\end{equation}
where $\text{f}$ can be arbitrary neural network computing the relatedness between the decoder state $\vec{s}_t$ and encoder hidden state state $\vec{h}_i$. One common option is to apply an additive attention mechanism $\text{f}(\vec{h}_i, \vec{s}_t) = \vec{v}^T \text{tanh}(\vec{W}_h \vec{h}_i + \vec{W}_s \vec{s}_t + b)$ where $\vec{v}$, $\vec{W}_h$, $\vec{W}_s$ and $b$ are learnable weights. Given the attention vector $\vec{a}^t$ at the $t$-th decoding step, the context vector can be computed as a weighted sum of the encoder hidden states, formulated as
\begin{equation}
\begin{aligned}
\vec{h}_t^* = \sum_i a^t_i \vec{h}_i,
\end{aligned}
\end{equation}
The computed context vector will be concatenated with the decoder state, and fed through some neural network for producing a vocabulary distribution.

The copying mechanism~\citep{vinyals2015pointer,gu2016incorporating} was introduced to directly copy tokens from the input sequence to the output sequence in a learnable manner. This can be very helpful in some scenarios where the output sequence refers to some named entities or out-of-vocabulary tokens in the input sequence. Specifically, at each decoding step $t$, a generation probability will be calculated for deciding whether to generate a token from the vocabulary or copy a token from the input sequence by sampling from the attention distribution $\vec{a}^t$. The generation probability can be computed as
\begin{equation}
\begin{aligned}
p_{\text{gen}} = \sigma(\vec{w}_{h^*}^T \vec{h}_t^* + \vec{w}_s^T \vec{s}_t + \vec{w}_x^T \vec{x}_t + b_\text{ptr})),
\end{aligned}
\end{equation}
where $\vec{w}_{h^*}$, $\vec{w}_s$, $\vec{w}_x$ and $b_\text{ptr}$ are learnable weights, $\sigma$ is a sigmoid function, and $p_{\text{gen}}$ is a scalar between 0 and 1.

The coverage mechanism~\citep{tu2016modeling} was proposed to encourage the full utilization of different tokens in the input sequence. This can be useful in some NLP tasks such as machine translation. Specifically, at each decoding step $t$, a coverage vector $\vec{c}^t$ which is the aggregated attention vectors over all previous decoding steps will be computed as
\begin{equation}
\begin{aligned}
\vec{c}^t = \sum_{t'=0}^{t-1} \vec{a}^{t'}.
\end{aligned}
\end{equation}
In order to encourage better utilization of those source tokens that have not received enough attention scores so far,
the above coverage vector will be used as extra input to the aforementioned attention mechanism~\cref{eq:attn}, that is, 
\begin{equation}
\begin{aligned}
e^t_i &= \text{f}(\vec{h}_i, \vec{s}_t, c_i^t)
\end{aligned}
\end{equation}
To avoid generating repetitive text, a coverage loss is calculated at each decoding step to penalize repeatedly attending to the same locations, formulated as,
\begin{equation}
\begin{aligned}
\text{covloss}_t = \sum_i \text{min}(a_i^t, c_i^t)
\end{aligned}
\end{equation}
The above coverage loss essentially penalizes the overlap between the attention vector and the coverage vector, and is bounded to $\sum_i a_i^t = 1$. It will be reweighted and added to the overall loss function.


The exposure bias occurs when during the training phase, the ground-truth token is used as the input (i.e., for better supervision) to the decoder for predicting the next token, while in the inference phase, the decoder's prediction from the previous time step is used as the input for next step prediction (due to no access to the ground-truth token).
In order to reduce this gap between training and inference phases and thus increase the generalization ability of the original Seq2Seq models, scheduled sampling~\citep{DBLP:conf/nips/BengioVJS15} was proposed to alleviate this issue by taking as input either the decoder's prediction from the previous time step or the ground truth with some probability for next step prediction, and gradually decreasing the probability of feeding in the ground truth at each iteration of training.
The celebrated Seq2Seq models equipped with the above effective techniques have achieved great successes in a wide range of NLP applications such as neural machine translation~\citep{bahdanau2015neural,luong2015effective,gehring2017convolutional}, 
text summarization~\citep{nallapati2016abstractive,see2017get,paulus2018deep},
text generation~\citep{song2017amr},
speech recognition~\citep{zhang2017very},
and dialog systems~\citep{serban2016building,serban2017hierarchical}.

\subsubsection{Discussions}
Seq2Seq models were originally developed to solve sequence-to-sequence problems, that is, to map a sequential input to a sequential output.
However, many NLP applications naturally admit graph-structured input data such as dependency graphs~\citep{fu2019graphrel,chen2020reinforcement}, constituency graphs~\citep{li-etal-2020-graph-tree,marcheggiani-titov-2020-graph}, AMR graphs~\citep{beck-etal-2018-graph,song2019semantic}, IE graphs~\citep{DBLP:conf/naacl/CaoFT19,huang-etal-2020-knowledge} and knowledge graphs~\citep{nathani-etal-2019-learning,wu2019relation}.
In comparison with sequential data, graph-structured data is able to encode rich syntactic or semantic relationships among objects.
Moreover, even if the raw input is originally represented in a sequential form, it can still benefit from explicitly incorporating rich structural information (e.g., domain-specific knowledge) to the sequence.
The above situations essentially call for an encoder-decoder framework for learning a graph-to-X mapping where X can stand for a sequence, tree or even graph.
Existing Seq2Seq models face a significant challenge in learning an accurate mapping from graph to the appropriate target due to its incapability of modeling complex graph-structured data.

Various attempts have been made in order to extend Seq2Seq models to handle Graph-to-Sequence problems where the input is graph-structured data.
A simple and straightforward approach is to directly linearize the structured graph data into the sequential data~\citep{iyer2016summarizing,gomez2018automatic,liu2017retrosynthetic}, and apply the Seq2Seq models to the resulting sequence. However, this kind of approaches suffer significant information loss, which leads to downgraded performance.
The root cause of RNN's incapability of modeling complex structured data is because it is a linear chain.
In light of this, some research efforts have been devoted to 
extend Seq2Seq models.
For instance, Tree2Seq~\citep{eriguchi2016tree} extends Seq2Seq models by adopting Tree-LSTM~\citep{tai2015improved} which is a generalization of chain-structured LSTM to tree-structured network topologies.
Set2Seq~\citep{DBLP:journals/corr/VinyalsBK15} is an extension of Seq2Seq models that goes beyond sequences and handles the input set using the attention mechanism.
Although these Seq2Seq extensions achieve promising results on certain classes of problems, none of them can model arbitrary graph-structured data in a principled way.



\subsection{Graph-to-Sequence Models} 


\subsubsection{Overview}

To address the aforementioned limitations of Seq2Seq models on encoding rich and complex data structures, recently, a number of graph-to-sequence encoder-decoder models for NLP tasks have been proposed~\citep{bastings-etal-2017-graph,beck-etal-2018-graph,song-etal-2018-graph,xu2018graph2seq}.
This kind of Graph2Seq models typically adopt a GNN based encoder and a RNN/Transformer based decoder.
Compared to the Seq2Seq paradigm, the Graph2Seq paradigm is better at capturing the rich structure information of the input text and can be applied to arbitrary graph-structured data.
Graph2Seq models have shown superior performance in comparison with Seq2Seq models in a wide range of NLP tasks including 
neural machine translation~\citep{bastings-etal-2017-graph,marcheggiani-etal-2018-exploiting,beck-etal-2018-graph,song2019semantic,Xu2020DocumentGF,yao2020heterogeneous,yin-etal-2020-novel,Cai_Lam_2020},
AMR-to-text~\citep{beck-etal-2018-graph,song-etal-2018-graph,damonte-cohen-2019-structural,ribeiro-etal-2019-enhancing,zhu-etal-2019-modeling,ijcai2020-542,guo-etal-2019-densely,yao2020heterogeneous,wang2020amr,Cai_Lam_2020,bai-etal-2020-online,song-etal-2020-structural,zhao-etal-2020-line,zhang-etal-2020-lightweight,jin-gildea-2020-generalized},
text summarization~\citep{fernandes2018structured,xu2020discourse,huang-etal-2020-knowledge,zhang-etal-2020-summarizing},
question generation~\citep{chen2020reinforcement,DBLP:conf/coling/WangXLZS20},
KG-to-text~\citep{koncel-kedziorski-etal-2019-text},
SQL-to-text~\citep{xu-etal-2018-sql},
code summarization~\citep{liu2021retrieval},
and semantic parsing~\citep{xu-etal-2018-exploiting}.

\subsubsection{Approach}

Most proposed Graph2Seq models were designed for tackling particular NLG tasks. In the followings, we will discuss some common techniques adopted in a wide rage of Graph2Seq variants, which include both graph-based encoding techniques and sequential decoding techniques.


\paragraph{Graph-based Encoders}

Early Graph2Seq methods and their follow-up works~\citep{bastings-etal-2017-graph,marcheggiani-etal-2018-exploiting,damonte-cohen-2019-structural,guo-etal-2019-densely,xu2020discourse,Xu2020DocumentGF,zhang-etal-2020-summarizing,zhang-etal-2020-lightweight} mainly used some typical GNN variants as the graph encoder inclduing GCN, GGNN, GraphSAGE and GAT.
Since the edge direction in a NLP graph often encodes critical information about the semantic relations between two vertices, it is often extremely helpful to capture the bidirectional information of text~\citep{devlin-etal-2019-bert}.
In the literature of Graph2Seq paradigm, some efforts have been made to extend the existing GNN models to handle directed graphs.
The most common strategy is to introduce separate model parameters for different edge directions (i.e., incoming/outgoing/self-loop edges) when performing neighborhood aggregation~\citep{marcheggiani-etal-2018-exploiting,song-etal-2018-graph,song2019semantic,Xu2020DocumentGF,yao2020heterogeneous,ijcai2020-542,guo-etal-2019-densely}.

Besides the edge direction information, many graphs in NLP applications are actually multi-relational graphs where the edge type information is very important for the downstream task.
In order to encode edge type information, some works~\citep{simonovsky2017dynamic,DBLP:conf/coling/ChenTLY18,ghosal-etal-2020-kingdom,wang_relational_2020,schlichtkrull2018modeling,teru2020inductive} have extended them by having separate model parameters for different edge types (i.e., similar ideas have been used for encoding edge directions).
However, in many NLP applications (e.g., KG-related tasks), the total number of edge types is large, hence the above strategy can have severe scalability issues.
To this end, some works~\citep{beck-etal-2018-graph,koncel-kedziorski-etal-2019-text,yao2020heterogeneous,ribeiro-etal-2019-enhancing,guo-etal-2019-densely,chen2020toward} proposed to bypass this problem by converting a multi-relational graph to a Levi graph~\citep{levi1942finite} and then utilize existing GNNs designed for homogeneous graphs as encoders.
Another commonly adopted technique is to explicitly incorporate edge embeddings into the message passing mechanism~\citep{marcheggiani-etal-2018-exploiting,song-etal-2018-graph,song2019semantic,zhu-etal-2019-modeling,wang2020amr,Cai_Lam_2020,ijcai2020-542,song-etal-2020-structural,liu2021retrieval,jin-gildea-2020-generalized}.

Besides the above widely used GNN variants, some Graph2Seq works also explored other GNN variants such as GRN~\citep{song-etal-2018-graph,song2019semantic} and GIN~\citep{ribeiro-etal-2019-enhancing}.
Notably, GRN is also capable of handling multi-relational graphs by explicitly including edge embeddings in the LSTM-style message passing mechanism.


\paragraph{Node \& Edge Embeddings Initialization}
For GNN based approaches, initialization of nodes and edges is extremely critical.
While both CNNs and RNNs are good at capturing local dependencies among consecutive words in text, GNNs do well in capturing local dependencies among neighboring nodes in a graph.
Many works on Graph2Seq have shown benefits of initializing node and/or edge embeddings by applying CNNs~\citep{bastings-etal-2017-graph,marcheggiani-etal-2018-exploiting} or bidirectional RNNs (BiRNNs)~\citep{bastings-etal-2017-graph,marcheggiani-etal-2018-exploiting,fernandes2018structured,xu2018graph2seq,xu-etal-2018-sql,koncel-kedziorski-etal-2019-text,Cai_Lam_2020,DBLP:conf/coling/WangXLZS20,chen2020reinforcement,liu2021retrieval} to the word embedding sequence before applying the GNN based encoder.
Some works also explored to initialize node/edge embeddings with BERT embeddings+BiRNNs~\citep{xu2020discourse,chen2020reinforcement} or RoBERTa+BiRNNs~\citep{huang-etal-2020-knowledge}.


\paragraph{Sequential Decoding Techniques}
Since the main difference between Seq2Seq and Graph2Seq models is on the encoder side, common decoding techniques used in Seq2Seq models such as attention mechanism~\citep{bahdanau2015neural,luong2015effective}, copying mechanism~\citep{vinyals2015pointer,gu2016incorporating}, coverage mechanism~\citep{tu2016modeling},
and scheduled sampling~\citep{DBLP:conf/nips/BengioVJS15}
can also be adopted in Graph2Seq models with potential modifications.

Some efforts have been made to adapt common decoding techniques to the Graph2Seq paradigm.
For example, in order to copy the whole node attribute containing multi-token sequence from the input graph to the output sequence, \citet{chen2020toward} extended the token-level copying mechanism to the node-level copying mechanism.
To combine the benefits of both sequential encoder and graph encoder, \citet{pan2020semantic,sachan2020stronger} proposed to fuse their outputs to a single vector before feeding it to a decoder. \citet{huang-etal-2020-knowledge} designed separate attention modules for sequential encoder and graph encoder, respectively.
\begin{equation}
\begin{aligned}
a_i^v &= \text{attn\_v}(\vec{s}_t, \vec{h}_i^v)\\
\vec{c}^v &= \sum_i a_i^v \vec{h}_i^v\\
a_j^s &= \text{attn\_s}(\vec{s}_t, \vec{h}_j^s, \vec{c}^v)\\
\vec{c}^s &= \sum_i a_j^s \vec{h}_j^s\\
\vec{c} &= \vec{c}^v || \vec{c}^s
\end{aligned}
\end{equation}
where $\vec{h}_i^v$ and $\vec{h}_j^s$ are the graph encoder outputs and sequential encoder outputs, respectively.
$\vec{c}$ is the concatenation of the graph context vector $\vec{c}^v$ and sequential context vector $\vec{c}^s$.

In order to tackle the limitations (e.g., exposure bias and discrepancy between the training and inference phases) of cross-entropy based sequential training, \citet{chen2020reinforcement} proposed to train the Graph2Seq system by minimizing a hybrid loss combining both cross-entropy loss and reinforcement learning~\citep{williams1992simple} loss.
While LSTM or GRU based decoders are the most commonly used decoder in Graph2Seq models, some works also employed a Transformer based decoder~\citep{koncel-kedziorski-etal-2019-text,zhu-etal-2019-modeling,yin-etal-2020-novel,wang2020amr,Cai_Lam_2020,bai-etal-2020-online,ijcai2020-542,song-etal-2020-structural,zhao-etal-2020-line,jin-gildea-2020-generalized}.

\subsubsection{Discussions}

There are some connections and differences between Graph2Seq models and Transformer-based Seq2Seq models, and many of them have already been discussed above when we talk about the connections and differences between GNNs and Transformer models. It is worth noting that there is a recent trend in combining the benefits of the both paradigms, thus making them less distinguishable.
Many recent works designed various graph transformer based generation models (as we discussed above) which employ a graph-based encoder combining both the benefits of GNNs and Transformer, and a RNN/Transformer based decoder.

Despite the great success of Graph2Seq models, there are some open challenges.
Many of these challenges are essentially the common challenges of applying GNNs for graph representation learning, including how to better model multi-relational or heterogeneous graphs, how to scale to large-scale graphs such as knowledge graphs, how to conduct joint graph structure and representation learning, how to tackle the over-smoothing issue and so on.
In addition, Graph2Seq models also inherit many challenges that Seq2Seq models have, e.g., how to tackle the limitations of cross-entropy based sequential training (e.g., exposure bias and discrepancy between the training and inference phases).

\begin{figure}[ht!]
\centering
\vspace{0mm}
\includegraphics[width=12.0cm]{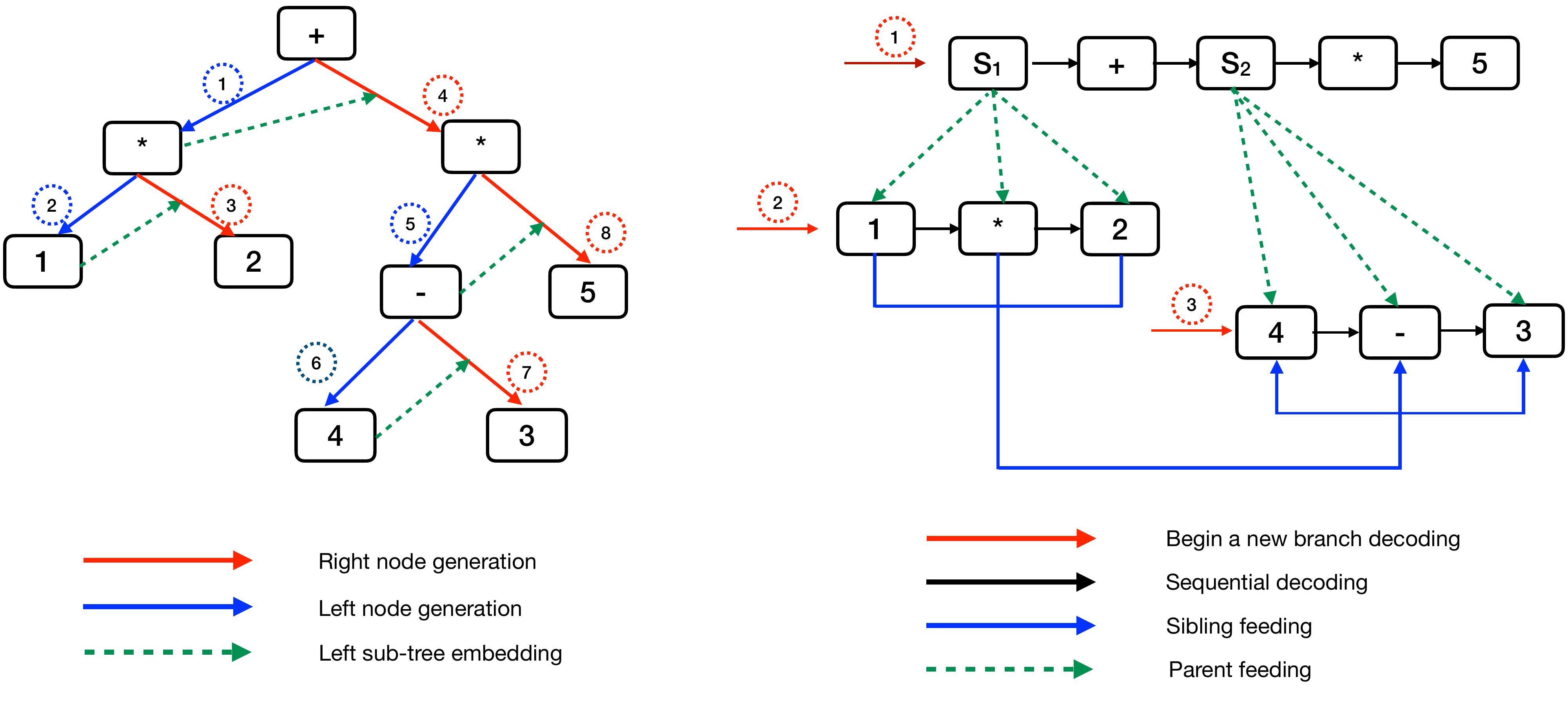}
\vspace{-4mm}
\caption{Equation: ( 1 * 2 ) + ( 4 - 3 ) * 5. Left: a DFS-based tree decoder example, the number stands for the order of the decoding actions. Right: a BFS based tree decoder example. Nodes like $S_1, S_2$ stand for sub-tree nodes, and once a sub-tree node generated, decoder will start a new branch for a new descendant decoding process. The number stands for the order of different branching decoding processes.}
\label{fig:tree-decoder-sample}
\vspace{-2mm}
\end{figure}

\subsection{Graph-to-Tree Models} 

\subsubsection{Overview}

Compared to Graph2Seq model, which considers the structural information in the input side, many NLP tasks also contain outputs represented in a complex structured, such as trees, which are also rich in structural information at the output side, e.g., syntactic parsing\citep{ji-etal-2019-graph}\citep{yang2020strongly}, semantic parsing\citep{li-etal-2020-graph-tree}\citep{xu-etal-2018-exploiting}, math word problem solving\citep{li-etal-2020-graph-tree}\citep{zhang-etal-2020-graph-tree}. It is a natural choice for us to consider the structural information of these outputs. 
To this end, some Graph2Tree models are proposed to incorporate the structural information in both the input and output side, which make the information flow in the encoding-decoding process more complete. 

\subsubsection{Approach}

To illustrate how the Graph2Tree model works, we will introduce how different components of the Graph2Tree model operate here, including: graph construction, graph encoder, attention mechanism, and tree decoder.

\paragraph{Graph construction}
The graph construction module, which is usually highly related to specific tasks, could be divided into two categories: one with auxiliary information and one without auxiliary information. For the former, \citet{li-etal-2020-graph-tree} use syntactic graph in both semantic parsing and math word problem solving tasks, which consists of the original sentence and the syntactic pasing tree (dependency and constituency tree). And the input graph in \citep{yin2018learning} considers the graph composed of the abstract syntax tree (AST) of a fragment of source code. For the latter, the input can usually form a task graph itself without auxiliary information. For example, \citet{zhang-etal-2020-graph-tree} employ the relationship between different numbers in the math word problem in the graph construction module.

\paragraph{Graph encoder}
Graph encoder is used for embedding the input graph into the latent representation. To implementing the graph encoder module, several graph2tree models use relatively simple GNN models, such as GCN\citep{kipf2016semi}, GGNN\citep{li2015gated}, and GraphSAGE\citep{hamilton2017inductive}. For \citep{li-etal-2020-graph-tree}, it uses a bidirectional variant of the GraphSage model, and \citet{zhang-etal-2020-graph-tree} exploit the GCN model before a transformer layer. And \citet{yin2018learning} simply adopt the GGNN model as its neural encoder.

\paragraph{Attention}
The attention module is a key component in an encoder-decoder framework, which carry the important information for bridging the input and output semantics. In the graph2tree model, the input graph often contains different types of nodes\citep{li-etal-2020-graph-tree}\citep{zhang-etal-2020-graph-tree}, while the traditional attention module can not distinguish between these nodes. In \citep{li-etal-2020-graph-tree}, the author uses the separate attention module to calculate the attention vector for different nodes in the input graph where some nodes is from the original sentence, and others are composed of the nodes in parsing trees generated by the external parser. It has been validated that distinguishing these two types of nodes could facilitate better learning process than the original attention module. This idea is similar to the application of Tree2Seq\citep{eriguchi2016tree} attention module in machine translation.

Specifically in \citep{li-etal-2020-graph-tree}, the decoder generates the tree structure by representing some branching nodes as non-terminal nodes, i.e., node $S_1$ in Figure \ref{fig:tree-decoder-sample}. Once these nodes generated, the decoder will start a new sequential decoding process. 
The decoder hidden state $\mathbf{s}_t$ at time step $t$ is calculated a
\vspace{-2mm}
\begin{equation}
  \textbf{s}_t = f_{decoder}(y_{t-1}, \textbf{s}_{t-1};\textbf{s}_{par};\textbf{s}_{sib}),
\end{equation}
where the $\textbf{s}_{par}, \textbf{s}_{sib}$ stand for the parent node hidden state and sibling node hidden state as illustrated in Figure \ref{fig:tree-decoder-sample}. After the current hidden state generated, the output module including attention layer is calculated as follows:
\begin{equation}
\alpha_{t(v)} = \frac{\exp(score( {\textbf{z}_v},\textbf{s}_t))}{\exp(\sum_{k=1}^{V_1} score({\textbf{z}_{k}},\textbf{s}_t))}, \forall v \in \mathcal{V}_1
\end{equation}
\vspace{-2mm}

\begin{equation}
\beta_{t(v)} = \frac{\exp(score( {\textbf{z}_v},\textbf{s}_t))}{\exp(\sum_{k=1}^{V_2} score({\textbf{z}_{k}},\textbf{s}_t))}, \forall v \in \mathcal{V}_2
\end{equation}
\vspace{0mm}

\vspace{-4mm}
\begin{align}
\textbf{c}_{v_1} &= \sum \alpha_{t(v)} \textbf{z}_v, \forall v \in \mathcal{V}_1 \\
\textbf{c}_{v_2} &= \sum \beta_{t(v)} \textbf{z}_v, \forall v \in \mathcal{V}_2
\end{align}
where $\mathbf{z}_v$ denotes to the learned node embedding for node $\mathbf{v}$, $\mathcal{V}_1$ denotes to the node set including all words from original sentences, and $\mathcal{V}_2$ denotes to another node set including all other nodes.
We then concatenate the context vector $\textbf{c}_{v_1}$, context vector $\textbf{c}_{v_2}$ and decoder hidden state $\textbf{s}_t$ to compute the final attention hidden state at this time step as:

\vspace{-1mm}
\begin{equation}
    \tilde{\textbf{s}_t} = \tanh(W_c \cdot [\textbf{c}_{v_1};\textbf{c}_{v_2};\textbf{s}_t]+b_c),
\end{equation}
where $W_c$ and $b_c$ are learnable parameters. The final context vector $\bm{\tilde{s_{t}}}$ is further fed to the output layer which is a softmax function after a feed-forward layer.

\paragraph{Tree decoder}
The output of some applications (i.e., semantic parsing, code generation, and math word problem) contain structural information, for example, the output in math word problem is a mathematical equation, which can be expressed naturally by the data structure of the tree. To generate these kinds of outputs, tree decoders are widely used in these tasks. Tree decoders can be divided into two main parts as shown in Figure \ref{fig:tree-decoder-sample}, namely, dfs (depth first search) based tree decoder, and bfs (breadth first search) based tree decoder.

For bfs-based decoders\citep{li-etal-2020-graph-tree, dong-lapata-2016-language,alvarez2016tree}, the main idea is to represent all the sub-trees in the tree as non-terminal nodes, and then use sequence decoding to generate intermediate results. If the results contains non-terminals, then we start branching (begin a new decoding process) with this node as the new root node, until the entire tree is expanded. 

For dfs-based decoders\citep{zhang-etal-2020-graph-tree,yin2018learning}, they regards the entire tree generation process as a sequence of actions. For example, in the generation of a binary tree (mathematical equation) in \citep{zhang-etal-2020-graph-tree}, the root node is generated in priority at each step, following by the generation of the left child node. After all the left child nodes are generated, a bottom-up manner is adopted to begin the generation of the right child nodes. 

In addition, the tree decoder is constantly evolving, and some techniques are proposed to collect more information during the decoding process or leverage the information from the input or output, such as parent feeding\citep{dong-lapata-2016-language}, sibling feeding\citep{li-etal-2020-graph-tree}, sub-tree copy\citep{yin2018learning}, tree based re-ranking\citep{do-rehbein-2020-neural} and other techniques. At the same time, the wide application of the transformer model also brings about many transformer based tree decoders\citep{sun2020treegen}\citep{li2020auto}, which proves the wide application of tree decoder and Graph2tree model.

    
    
    
    

\subsection{Graph-to-Graph Models}
The graph-to-graph models that are typically utilized for solving graph transformation problem as a graph encoder-decoder model. The graph encoder generates the latent representation of each node in the graph or generate one graph-level latent representation for the whole graph via the GNNs. The graph decoder then generates the output target graphs based on the node-level or graph-level latent representations from the encoder. 
In this section, we first introduce graph-to-graph transformation problem and the typical NLP applications that can be formalized as graph-to-graph transformation problems. Then, we introduce the specific techniques for a Graph-to-graph model for information extraction.

\subsubsection{Overview}
\paragraph{Graph-to-graph transformation}
Graph-to-graph models aims to deal with the problem of deep graph transformation~\citep{guo2018deep}. The goal of graph transformation is to transform an input graph in the source domain to the corresponding output graphs in the target domain via deep learning. Emerging as a new while important problem, deep graph transformation has multiple applications in many areas, such as molecule optimization~\citep{shi2020graph, zhou_neural_2020, do2019graph} and malware confinement in cyber security~\citep{guo2019deep}. Considering the entities that are being transformed during the translation process, there are three categories of sub-problems: node transformation, edge transformation, and node-edge-co-transformation. For node transformation, only the node set or nodes' attributes in the input graph can change during the transformation process. For edge transformation, only the graph topology or edge' attributes in the input graph can change during the transformation process. While for node-edge-co-transformation, both the attributes of nodes and edges can change.

\paragraph{Graph-to-Graph for NLP}
Since the natural language or information knowledge graphs can be naturally formalized as graphs with a set of nodes and their relationships, many generation tasks in the domain of NLP can be formalized as a graph transformation problem, which can further be solved by the graph-to-graph models. In this way, the semantic structural information of both the input and output sentences can be fully utilized and captured. Here, two important NLP tasks (i.e., information extraction and semantic parsing), which can be formalized as the graph-to-graph problems, are introduced as follows.

\textbf{Graph Transformation for Information Extraction}.
Information extraction is to extract the structured information from a text, which usually consists of name entity recognition, relation extraction and co-reference linking. The problem of information extraction can be formalized as a graph transformation problem, where the input is the dependency or constituency graph of a text and the output is the information graph. In input dependency or constituency graph, each node represents a word token and each edge represent the dependency relationship between two nodes. In output information graph, each node represent a name entity and each edge represent the either the semantic relation the co-reference link between two entities. In this way, the information extraction is about generating the output information graph given the input dependency or constituency graph.

\textbf{Graph Transformation for Semantic Parsing}.
The task of semantic parsing is about mapping natural language to machine interpretable meaning representations, which in turn can be expressed in many different formalisms, including lambda calculus, dependency-based compositional semantics, frame semantics, abstract meaning representations (AMR), minimal recursion semantics, and discourse representation theory~\citep{fancellu2019semantic}. Explicitly or implicitly, a representation in any of these formalisms can be expressed as a directed acyclic graph (DAG). Thus, semantic parsing can also be formalized as a graph transformation problem, where the input is the dependency or constituency graph and the output is the directed acyclic graph for semantics. For example, the semantic formalism for AMR can be encoded as a rooted, directed, acyclic
graph, where nodes represent concepts, and labeled directed edges represent the relationships between them~\citep{flanigan2014discriminative, fu2021end}.

Sequence-to-graph transformation can be regarded as a special case of the graph-to-graph, where the input sequence is a line-graph. Sequence-to-graph models are popularly utilized for AMR parsing tasks, where the goal is to learning the mapping from a sentence to its AMR graph~\citep{zhang2019amr}. To generate the AMR tree with indexed node, the approach to parsing is formalized as a two-stage process: node prediction and edge prediction. The whole process is implemented by an pointer-network, where the encoder is a multi-layer bi-direction-RNN and the nodes in the target graphs are predicted in sequence. After this, the edges among each pair of nodes are predicted based on the learnt embedding of the ending nodes.

\subsubsection{Approach}
In this subsection, we introduce an example graph-to-graph model in dealing with the task of information extraction by describing its challenges and methodologies. 


\textbf{Challenges for Graph-to-Graph IE}.
There are three main challenges in solving the graph-to-graph IE problem: (1) Different resolution between the input and output graphs. The nodes in the input dependency graph represent word tokens, while the nodes in the output information graph represent name entities; (2) Difficult to involve both the sentence-level and word-level. To learn the word embedding in the graph encoder, it is important to consider both the word interactions in a sentence and the sentence interactions in a text; and (3) Difficult to model the dependency between entity relations and co-reference links in the graph decoder. The generation process of entity relation and co-reference links are dependent on each other. For example, if words “Tom” and “He” in two separate sentences have a co-reference link, and “Tom” and “London” has the relation of “born\_In”, then “He” and “London” should also have the relation of “born\_In”.

\textbf{Methodology.}
To solve the above mentioned challenges for the graph-to-graph IE task, here we introduce an end-to-end encoder-decoder based graph-to-graph transformation model, which transforms the input constructed graphs of text into the information graphs which contains name entities as well as co-reference links and relations among entities. The whole model consists of a hierarchy graph encoder for span embedding and a parallel decoder with co-reference link and entity relation generation. 

First, to construct the initial graphs, the dependency information are formalized into a heterogeneous graph which consists of nodes representing word tokens (i.e., word-level nodes) and nodes representing sentences (i.e., sentence-level nodes). There are also three types of edges in the graph. One type of edges represent the dependency relationships between word-level nodes (i.e., dependency edges). One type of edges represent the adjacent relationship between sentence-level nodes (i.e., adjacent edges). The last type of edges represent the belongingness between the word-level and sentence-level nodes (i.e., interactive edges).

Second, the constructed heterogeneous graph is inputted into the encoder, which is based on a hierarchy graph convolution neural network. Specifically, for each layer, the conventional message passing operations are first conducted along the dependency and adjacent edges to update the embedding of word-level and sentence-level nodes. Then, the conventional message passing operations are conducted along the interactive edges based on the newly updated word-level and sentence-level nodes' embedding. After several layers of propagation, the embedding of word-evel nodes will contains both the dependency and sentence-level information. 

Third, based on the words' embedding from the encoder, the name entities can be first extracted via BIO Tagging~\citep{marquez2005semantic}. Thus, the entity-level embedding are then constructed by summing all of the embedding of the words it contains. Given the entity embedding, to model the dependency between the co-reference links and relations between entities, a parallel graph decoder~\citep{guo2018deep} that involves both co-reference link and entity relation generation processes is utilized. 
Specifically, given the entity embedding $h_i$ and $h_j$ of a pair of name entities $v_i$ and $v_j$, the initial generated latent representation of co-reference link $\textbf{c}^{0}_{i,j}$ is computed as:
\begin{equation}
\label{eq:n2edeconv_direct}
 \textbf{c}^{0}_{i,j}=\sum\nolimits^{C}_{m=1}(\sigma (h_i^{m}\Bar{\mu}_j)+\sigma(h_j^{m}\Bar{\nu}_i)),
\end{equation}
where $\sigma (h_i^{m}\Bar{\mu}_j)$ means the deconvolution contribution of node $v_i$ to its edge representations with node $v_j$, which is made by the $m$-th entry of its node representations, and $\Bar{\mu}_j$ represents one entry of the deconvolution filter vector $\Bar{\mu}\in\mathbb{R}^{N \times 1}$ that is related to node $v_j$. The initial relation latent representation $e^{0}_{i,j}$ between a pair of name entities $v_i$ and $v_j$ can also be computed in the same way. $C$ refers to the length of name entity embedding. 

Given the initial latent representation of co-reference links and relations, the co-reference link representation $\textbf{c}^{l}_{i,j}$ at the $l$-th layer is computed as follows:
\begin{equation}
\label{eq:e2edeconv_direct}
 \textbf{c}^{l}_{i,j}=\sigma (\Bar{\phi}^{l-1}_j\sum\nolimits_{k_1=1}^{N}[\textbf{c}^{l-1};\textbf{e}^{l-1}]^{l-1}_{i,k_1}x_{k_1})+\sigma(\Bar{\psi}^{l}_i\sum\nolimits_{k_2=1}^{N}[\textbf{c}^{l-1};\textbf{e}^{l-1}]^{l-1}_{k_2,j}x_{k_1}),
\end{equation}\normalsize 
where $\Bar{\phi}^{l-1}_j\sum_{k_1=1}^{N}[\textbf{c}^{l-1};\textbf{e}^{l-1}]^{l-1}_{i,k_1}x_{k_1}$ can be interpreted as the decoded contribution of node $v_i$ to its edge representations with node $v_j$, and $\Bar{\phi}^{l-1}_j$ refers to the element of deconvolution filter vector that is related to node $v_j$. The output of the last ``edge'' deconvolution layer denotes the probability of the existence of an edge in the target graph. All the symbols $\sigma$ refers to the activation functions. $[\textbf{c}^{l-1};\textbf{e}^{l-1}]$ refers to the concatenation of all the co-reference and relation representations at $(l-1)$-th layer.

\section{Applications}
\label{sec:Applications}

In this chapter, we will discuss a large variety of typical NLP applications using GNNs, including natural language generation, machine reading comprehension, question answering, dialog system, text classification, text matching, topic modeling, sentiment classification, knowledge graph, information extraction, semantic and syntactic parsing, reasoning, and semantic role labelling. We also provide the summary of all the applications with their sub-tasks and evaluation metrics in Table~\ref{tab:app-table}.

\begin{table}[]
\caption{Typical NLP applications and relevant works using GNNs}
\resizebox{\textwidth}{!}{%
\begin{tabular}{|c|c|c|l|}
\hline
\textbf{Application} & \textbf{Task}  & \textbf{Evaluation} & \textbf{References} \\ \hline
\multirow{11}{*}{NLG} 
& {Neural}  & \multirow{3}{*}{BLEU} & \citet{bastings-etal-2017-graph, beck2018graph, cai2020graph} \\ 
& Machine & & \citet{guo-etal-2019-densely, marcheggiani-etal-2018-exploiting, shaw-etal-2018-self} \\ 
& Translation& & \citet{song2019semantic, xiao-etal-2019-lattice, Xu2020DocumentGF, yin-etal-2020-novel} \\ \cline{2-4} 
& \multirow{4}{*}{Summarization} & \multirow{4}{*}{ROUGE}  &  \citet{xu2020discourse, wang2019heterogeneous, li2020leveraging}  \\ 
&  &  & \citet{fernandes2018structured, wang-etal-2020-heterogeneous} \\
&  &  & \citet{cui2020enhancing, jia-etal-2020-neural, zhao-etal-2020-improving} \\
&  &  & \citet{jin2020semsum, yasunaga-etal-2017-graph, leclair2020improved} \\ \cline{2-4} 
&  & \multirow{4}{*}{BLEU, METEOR} &  \citet{bai-etal-2020-online, jin-gildea-2020-generalized, xu-etal-2018-sql} \\
& Structural-data &  & \citet{beck2018graph, cai-lam-2020-amr,zhu2019modeling} \\
& to Text &  & \citet{cai2020graph, ribeiro-etal-2019-enhancing, song-etal-2020-structural} \\
&  &  & \citet{wang2020amr, yao2018exploring, zhang-etal-2020-lightweight} \\ \cline{2-4}
& Natural Question  & BLEU, METEOR, &  \citet{chen2020toward, liu2019learning, pan2020semantic}    \\ 
& Generation & ROUGE & \citet{wang-etal-2020-answer, sachan2020stronger, su-etal-2020-multi} \\ \hline
\multirow{9}{*}{MRC and QA} &  & \multirow{4}{*}{F1, Exact Match}  & \citet{de2018question, DBLP:conf/naacl/CaoFT19, chen2020graphflow} \\
& Machine Reading &  & \citet{qiu-etal-2019-dynamically, schlichtkrull2018modeling, tang2020multi} \\
& Comprehension &  & \citet{tu-etal-2019-multi, song2018exploring} \\ 
& &  & \citet{fang2020hierarchical, zheng2020srlgrn} \\  \cline{2-4} 
& Knowledge Base & \multirow{2}{*}{F1, Accuracy}  &  \citet{feng2020scalable, sorokin2018modeling} \\
& Question Answering &  & \citet{ santoro2017simple,yasunaga2021qa} \\  \cline{2-4} 
& Open-domain & \multirow{2}{*}{Hits@1, F1}  & \multirow{2}{*}{\citet{han-etal-2020-open, sun-etal-2019-pullnet, sun-etal-2018-open}} \\
& Question Answering &  &  \\  \cline{2-4} 
& Community & \multirow{2}{*}{nDCG, Precision} & \multirow{2}{*}{\citet{hu2019hierarchical, hu2020multi}} \\ 
& Question Answering &  &  \\  \hline
\multirow{4}{*}{Dialog Systems}  & Dialog State Tracking  & Accuracy & \citet{DBLP:conf/coling/ChenTLY18,DBLP:conf/aaai/0002LWZT020} \\ \cline{2-4}
& Dialog Response & BLEU, METEOR,   & \multirow{2}{*}{\citet{DBLP:conf/ijcai/HuCL0MY19, bai2021semantic}} \\
& Generation & ROUGE &  \\  \cline{2-4} 
& Next Utterance Selection  & Recall@K & \citet{liu2021graph}\\ \hline
\multicolumn{2}{|c|}{\multirow{2}{*}{Text Classification}} & \multirow{2}{*}{Accuracy} & \citet{chen2020iterative, defferrard2016convolutional, henaff2015deep} \\
\multicolumn{2}{|c|}{} &  & \citet{DBLP:conf/emnlp/HuangMLZW19, hu2020heterogeneous,DBLP:conf/aaai/LiuYZWL20} \\ \hline
\multicolumn{2}{|c|}{Text Matching} & Accuracy, F1  &  \citet{chen2017neural,DBLP:conf/acl/LiuNWGHLX19}\\ \hline
\multicolumn{2}{|c|}{\multirow{2}{*}{Topic Modeling}}   & \multirow{2}{*}{Topic Coherence Score}  & \citet{long_graph_2020, yang_graph_2020}  \\
\multicolumn{2}{|c|}{}  & & \citet{zhou_neural_2020, zhu_graphbtm_2018} \\ \hline
\multicolumn{2}{|c|}{\multirow{4}{*}{Sentiment Classification}} & \multirow{4}{*}{Accuracy, F1} &  \citet{zhang-qian-2020-convolution, pouran_ben_veyseh_improving_2020}  \\ 
\multicolumn{2}{|c|}{}  & & \citet{chen-etal-2020-aspect, tang-etal-2020-dependency} \\
\multicolumn{2}{|c|}{}  & & \citet{sun-etal-2019-aspect, wang_relational_2020, zhang_aspect-based_2019} \\ 
\multicolumn{2}{|c|}{}  & & \citet{ghosal-etal-2020-kingdom, huang_syntax-aware_2019, chen2020inducing} \\ \hline
\multirow{6}{*}{Knowledge Graph} & Knowledge & \multirow{6}{*}{Hits@N} & \citet{malaviya2020commonsense, nathani-etal-2019-learning, teru2020inductive} \\ 
& Graph &  & \citet{bansal2019a2n, schlichtkrull2018modeling, shang2019end} \\
& Completion &  & \citet{wang-etal-2019-incorporating, wang2019robust, zhang2020relational} \\ \cline{2-2} \cline{4-4}
& Knowledge & & \citet{cao2019multi, li2019semi, sun2020knowledge}\\ 
& Graph &  & \citet{wang2018cross, wang-etal-2020-knowledge-graph, ye2019vectorized} \\
& Alignment &  & \citet{xu-etal-2019-cross-lingual, wu2019relation} \\ \hline
\multirow{5}{*}{Information Extraction}  & Named Entity & \multirow{5}{*}{Precision, Recall, F1}  & \citet{luo-zhao-2020-bipartite, ding-etal-2019-neural, gui-etal-2019-lexicon}\\
& Recognition &  & \citet{jin-etal-2019-fine, sui2019leverage} \\ \cline{2-2} \cline{4-4}
& \multirow{2}{*}{Relation Extraction} &  & \citet{qu2020few, zeng-etal-2020-double, sahu-etal-2019-inter} \\ 
&  &  & \citet{guo-etal-2019-attention, zhu-etal-2019-graph} \\ \cline{2-2} \cline{4-4}
& Joint Learning Models &  &  \citet{fu2019graphrel, luan2019general, sun-etal-2019-joint} \\ \hline
\multirow{3}{*}{Parsing} & Syntax-related  & \multirow{3}{*}{Accuracy} &  \citet{do-rehbein-2020-neural, ji-etal-2019-graph, yang2020strongly} \\ \cline{2-2} \cline{4-4} 
& \multirow{2}{*}{Semantics-related} &  & \citet{bai-etal-2020-online, zhou-etal-2020-amr}  \\ 
& &  & \citet{Shao_Gong_Qi_Cao_Ji_Lin_2020, bogin-etal-2019-representing, bogin-etal-2019-global}  \\ \hline
\multirow{6}{*}{Reasoning} & Math Word & \multirow{6}{*}{Accuracy} & \citet{li-etal-2020-graph-tree, Lee2020Mathematical, wu-etal-2020-knowledge} \\ 
& Problem Solving &  & \citet{ zhang-etal-2020-graph-tree, ferreira-freitas-2020-premise} \\ \cline{2-2} \cline{4-4}
& Natural Language & & \multirow{2}{*}{\citet{Kapanipathi2020InfusingKI, Wang_Kapanipathi_Musa_Yu_Talamadupula_Abdelaziz_Chang_Fokoue_Makni_Mattei_Witbrock_2019}} \\ 
& Inference &  &  \\ \cline{2-2} \cline{4-4}
& Commonsense &  & \multirow{2}{*}{\citet{zhou2018commonsense, lin2019kagnet,lin-etal-2019-kagnet}} \\
& Reasoning &  &  \\ \hline
\multicolumn{2}{|c|}{\multirow{2}{*}{Semantic Role Labelling}} & {Precision, Recall,}   &  \citet{marcheggiani-titov-2020-graph, xia2020semantic, zhang2020syntax} \\ 
\multicolumn{2}{|c|}{} & {F1}   &  \citet{li2018unified, marcheggiani2017encoding, fei2020cross} \\ \hline
\end{tabular}%
}
\label{tab:app-table}
\end{table}

\subsection{Natural Language Generation}
Natural language generation (NLG) aims to generate high-quality, coherent and understandable natural languages given various form of inputs like text, speech and etc while we only focus on the linguistic form. Modern natural language generation methods usually take the form of encoder-decoder, which encodes the input sequences into latent space and predicts a collection of words based on the the latent representation. Most modern NLG pipelines can be divided into two steps: Encoding and Decoding, which are processed by two module: encoder and decoder. In this section, we provide a comprehensive overview of the auto-regressive graph-based methodologies which exploit graph structures in encoder in this thriving area covering 1) neural machine translation, 2) summarization, 3) question generation, 4) structural-data to text.

\subsubsection{Neural Machine Translation}
\paragraph{Background and Motivation} The classic neural machine translation (NMT) system aims to map the source language's sentences into the target language without changing the semantic meaning. Most prior works~\citep{bahdanau2015neural,luong2015effective} adopt the attention-based sequence-to-sequence learning diagram, especially the RNN-based language model. Compared with the traditional machine translation models, these methods can produce much better performance without specific linguistic knowledge. However, these methods suffer from the long-dependency problem. With the development of attention mechanism, fully-attention-based models such as Transformer~\citep{vaswani2017attention}, which captures the implicit correlations by self-attention, have made a breakthrough and achieved a new state-of-art. Although these works achieve great success, they rarely take the structural information into account, such as the syntactic structure. Recently, with the help of powerful GNNs, many researchers further boost the performance by mining the structural knowledge contained in the unstructured texts.



\paragraph{Methodologies} Most GNN-based NMT methods cast the conventional seq2seq diagram to the Graph2Seq architecture. They firstly convert the input texts to graph-structured data, and then employ the GNN-based encoder to exploit the structural information. In this section, we introduce and summarize some representative GNN-related techniques adopted in recent NMT approaches regarding graph construction and representation learning.

\begin{itemize}
    \item \textbf{Graph Construction}. Various static graphs have been introduced to the NMT task to tackle corresponding challenges. \citet{bastings-etal-2017-graph,beck2018graph,cai2020graph,guo-etal-2019-densely} first converted the given texts into syntactic dependency graph. Such structure doesn't take semantic relations of words into account. Intuitively, it is beneficial to represent the redundant sentences by high-level semantic structure abstractions. To this end, \citet{marcheggiani-etal-2018-exploiting} construct the semantic-role-labeling based dependency graph for the given texts. What's more, \citet{beck2018graph,song2019semantic} construct the AMR graph for the sentences which can cover more semantic correlations. Besides the classic graph types, some specifically designed graphs (app-driven graphs) are proposed to address the unique challenges. Although source sentences in NMT are determined, either word-level or subword-level segmentations have multiple choices to split a source sequence with different word segments or different subword vocabulary sizes. Such a phenomenon is proposed to affect the performance of NMT~\citep{xiao-etal-2019-lattice}. They propose the lattice graph, which incorporates different segmentation of source sentences. \citet{shaw-etal-2018-self} construct the \textit{relative position graph} to explicitly model the relative position feature. \citet{yin-etal-2020-novel} build the multi-modal graph to introduce visual knowledge to NMT, which presents the input sentences and corresponding images in a unified graph to capture the semantic correlations better. 
    Despite the single-type static graph, \citet{Xu2020DocumentGF} construct the hybrid graph considering multiple relations for document-level NMT to address the severe long-dependency issue. In detail, they construct the graph considering both intra-sentential and inter-sentential relations. For intra-sentential relations, they link the words with sequential and dependency relations. For inter-sentential relations, they link the words in different sentences with lexical (repeated or similar) and coreference correlations.

\item\textbf{Graph Representation Learning}. Most of the constructed graphs in this line are heterogeneous graphs, which contain multiple node or edge types and can't be exploited directly by typical GNNs. Thus, researchers adopt various heterogeneous graph representation techniques. \citet{bastings-etal-2017-graph,marcheggiani-etal-2018-exploiting} regard the dependency graphs as multi-relational graphs and apply directed-GCN to learn the graph representation. Similarly, \citet{beck2018graph} firstly convert the constructed multi-relational graph to levi-graph and apply relational GGNN, which employs edge-type-specific parameters to exploit the rich structure information. \citet{Xu2020DocumentGF} regard the edge as connectivity and treat the edge direction as edge types, such as "in", "out", and "self". Then they apply relational GCN to encode the document graph. \citet{guo-etal-2019-densely} convert the heterogeneous graph to levi-graph and adopt the densely connected GCN to learn the embedding. \citet{song2019semantic} propose a special type-aware heterogeneous GGNN to learn the node embedding and edge representation jointly. Specifically, they first learn the edge representation by fusing both the source node and edge type's embeddings. Then for each node, they aggregate the representation from its incoming and outgoing neighbors and utilize a RNN based module to update the representation.

Besides the extension of traditional GNNs, Transformer is further explored to learn from the structural inputs in NMT. Unlike traditional Transformer which adopt absolute sinusoidal position embedding to ensure the self-attention learn the position-specific feature, \citet{shaw-etal-2018-self, xiao-etal-2019-lattice} adopt position-based edge embedding to capture the position correlations and make the transformer learn from the graph-based inputs. \citet{cai2020graph} learn the bidirectional path-based relation embedding and add it to the node embedding when calculating self-attention. They then find the shortest path from the given graph for any two nodes and apply bidirectional GRU to further encode the path to get the relation representation. \citet{yin-etal-2020-novel} apply graph-transformer-based encoder to learn the multi-modal graph. Firstly, for text modal's nodes, they get the initial embedding by summing up the word embedding and position embedding. As for visual nodes, they apply a MLP layer to project them to the unified space as text nodes. Secondly for each modal, they apply multi-head self-attention to learn the intra-modal representation. Thirdly, they employ GAT-based cross-modal fusion to learn the cross-modal representation.

\item\textbf{Special Techniques}. In order to allow information flow from both directions, some technqies are designed for incorporating direction information. For example, \citet{bastings-etal-2017-graph,marcheggiani-etal-2018-exploiting,beck2018graph} add the corresponding reverse edge as an additional edge type "reverse". The self-loops edge type are also added as type "self". For another example,  \citet{guo-etal-2019-densely} first add a global node and the edges from this global node to other nodes are marked with type "global". In addition, they further add bidirectional sequential links with type "forward" and "backward" between nodes existing in the input texts.

\end{itemize}

\paragraph{Benchmarks and Evaluation}

Common benchmarks for NMT from text include News Commentary v11, WMT14, WMT16, WMT19 for training, newstest2013, newstest2015, newstest2016, newsdev2019, newstest2019 for evaluation and testing. As for multi-modal NMT task, Multi30K dataset~\citep{elliott-etal-2016-multi30k} is widely used by previous works. As for evaluation metrics, BLEU is the a typical metric to evaluate the similarity between the generated and real output texts.





\subsubsection{Summarization}
\paragraph{Background and Motivation}
Automatic summarization is the task of producing a concise and fluent summary while preserving key information content and overall meaning~\citep{allahyari2017text}. It is a well-noticed but challenging problem due to the need of searching in overwhelmed textural data in real world. Broadly, there are two main classic settings in this task: 1) extractive summarization and 2) abstractive summarization. Extractive summarization task focus on selecting sub-sentences from the given text to reduce redundancy, which is formulated as a classification problem. In contrast, abstractive summarization follows the neural language generation task. It normally adopts the encoder-decoder architecture to generate the textual summary. Compared to the extractive summarization, the abstractive summarization setting is more challenging but more attractive since it can produce non-existing expressions. Traditional approaches~\citep{gehrmann2018bottom,zhou2018neural,liu2019fine} simply regard the inputs as sequences and apply the encoder like LSTM, Transformer, etc. to learn the latent representation, which fail to utilize the rich structural information implicitly existing in the natural inputs. Many researchers find that structural knowledge is beneficial to address some troublesome challenges, e.g., long-dependency problem, and thus propose the GNN-based techniques~\citep{wang-etal-2020-heterogeneous,fernandes2018structured} to explicitly leverage the structural information to boost the performance.

\paragraph{Methodologies} Most GNN-based summarization approaches firstly construct the graph to represent the given natural texts. Then they employ GNN-based encoders to learn the graph representation. After that, for extractive summarization models, they adopt the classifier to select candidate subsentences to compose the final summary. As for abstractive summarization, they mostly adopt the language decoder with maximizing the outputs' likelihood to generate the summary. In the following, we introduce some representative GNN-related techniques from the recent summarization methods.

\begin{itemize}
\item\textbf{Graph Construction}.
Here we introduce the different ways to construct suitable and effective graph for different types of inputs, including sign-documents, multi-documents and codes.

\textbf{\textit{Single-document based.}} \citet{fernandes2018structured} construct the hybrid graph, including sequential and coreference relation. To tackle the issue such as semantic irrelevance and deviation, \citet{Jin_Wang_Wan_2020} construct the semantic dependency graph and cast it as the multi-relational graph for the given texts. To capture the typical long-dependency in document-level summarization, \citet{xu2020discourse} construct the hybrid graph. They first construct the discourse graph by RST parsing and then add co-reference edges between co-reference mentions in the document.
To better capture the long-dependency relation in sentence-level and enrich the semantic correlations, \citet{wang-etal-2020-heterogeneous} regards both the sentences and the containing words as nodes and construct a similarity graph to model the semantic relations.
To model the redundant relation between sentences, \citet{jia-etal-2020-neural} propose to construct the hybrid heterogeneous graph containing three types of nodes: 1) named entity, 2) word, and 3) sentence as well as four types of edges: 1) sequential, 2) containing, 3) same, and 4) similar.
However, the methods above are mostly focused on the cross-sentence relations and overlook the inter-sentence, especially the topic information.
To this end, \citet{cui2020enhancing} and \citet{zhao-etal-2020-improving} construct the topic graph by introducing additional topic words to discover the latent topic information. On top of that, \citet{zhao-etal-2020-improving} mine the sub-graph of non-topic nodes to represent the original texts while preserving the topic information.

\textbf{\textit{Multi-document based.}}
\citet{yasunaga-etal-2017-graph} decompose the given document clusters into sentences and construct the discourse graph by Personalized Discourse Graph algorithm (PDG). \citet{li2020leveraging} split the documents into paragraphs and constructs three individual graphs: 1) similarity graph, 2) discourse graph, and 3) topic graph to investigate the effectiveness.

\textbf{\textit{Code based.}}
To fully represent the code information in the code summarization task, 
\citet{fernandes2018structured} construct the specific code graph for the given program clips. They first break up the identifier tokens (i.e., variables, methods, etc.) into sub-tokens by programming language heuristics. Then they construct the graph to organize the sub-tokens according to sequential positions and lexically usage. \citet{leclair2020improved} propose another way by firstly parsing the given programs into abstract syntax trees (AST) and then converting them to program graphs.

\item\textbf{Graph Representation Learning}

In the literature of summarization tasks, both homogeneous GNNs and heterogeneous GNNs have been explored to learn the graph representation. 
For homogeneous graphs, \citet{li2020leveraging} apply self-attention-based GAT to learn the representation on the fully-connected graph. Specifically, they introduce Gaussian kernel to mine the edge importance between nodes from the graph's topology. \citet{zhao-etal-2020-improving} adopt the GAT-based graph transformer, which regards the similarity learned by self-attention as edge weight.
For heterogeneous graphs, some researchers cast the heterogeneous graphs to homogeneous graphs by special techniques. For example, some works\citep{leclair2020improved,yasunaga-etal-2017-graph,xu2020discourse} ignore both the edges and nodes' types by treating the edge as connectivity.  \citet{cui2020enhancing} project the nodes to the unified embedding space to diminish the heterogeneity. After that, some classic GNNs are employed such as GCN~\citep{xu2020discourse,leclair2020improved,yasunaga-etal-2017-graph}, GAT~\citep{cui2020enhancing}. For example, \citet{fernandes2018structured} employ the relational GGNN to learn type-specific relations between nodes. \citet{wang-etal-2020-heterogeneous, jia-etal-2020-neural} firstly split the heterogeneous graph into two sub-graphs according to nodes' type (i.e., words graph and sentence graph) and then apply GAT-based cross-attention on two sub-graphs to learn the representation iteratively.

\item\textbf{Embedding Initialization}
The quality of the initial node embedding plays an important role in the overall performance of GNN-based methods. For graphs whose nodes are words, most approaches adopt the pre-trained word embeddings such as BERT~\citep{li2020leveraging,xu2020discourse,cui2020enhancing}, ALBERT~\citep{jia-etal-2020-neural}. Besides, since the topic graph~\citep{cui2020enhancing} introduces additional topic nodes, they initialize them by the latent representation of topic modeling. \citet{Jin_Wang_Wan_2020} apply Transformer to learn the contextual-level node embedding.
For nodes such as sentence-level nodes, which are composed of words, \citet{yasunaga-etal-2017-graph} adopt the GRU to learn the sentences' embedding (i.e., the node embeddings) from the corresponding word sequences. They adopt the last hidden state as the sentences' representation. Similarly, \citet{wang-etal-2020-heterogeneous} adopt CNN to capture the fine-grained n-gram feature and then employ Bi-LSTM to get the sentences' feature vectors.
\citet{jia-etal-2020-neural} apply the average pooling function to the ALBERT's encoder outputs to represent the sentence nodes, while
\citet{zhao-etal-2020-improving} initialize the nodes (utterances) by CNN and the topic words by LDA. 
\end{itemize}

\paragraph{Benchmarks and Evaluation} Common benchmarks for automatic summarization from documents include CNN/DailyMail~\citep{see2017get}, NYT~\citep{AB2/GZC6PL_2008}, WikiSum~\citep{liu2018generating}, MultiNews~\citep{fabbri2019multi}. As for code based summarization, Java~\citep{alon2018code2seq} and Python~\citep{barone2017parallel} are widely used. As for evaluation metrics, BLEU, ROUGE and human evaluation are commonly used.

\subsubsection{Structural-data to Text}

\paragraph{Background and Motivation}
Despite the natural texts, many NLP applications evolve the data which is represented by explicit graph structure, such as SQL queries, knowledge graphs, AMR, etc. The task of structural-data is to generate the natural language from structural-data input. Traditional works~\citep{pourdamghani2016generating,pourdamghani-etal-2014-aligning} apply the linearization mechanisms which map the structural-data to sequential data and adopt the Seq2Seq architecture to generate the texts. To fully capture the rich structure information, recent efforts focus on GNN-based techniques to handle this task. In the following, we introduce GNN techniques for three typical cases, namely AMR-to-text generation, SQL-to-text generation and RDF-to-text generation.


\paragraph{Methodologies} Most GNN-based AMR-to-text and SQL-to-text approaches typically construct domain-specific graphs such as AMR graphs and SQL-parsing-based graphs to organize the inputs. RDF-to-text generation often uses the graph structure inherent in the RDF triples. Following that, they apply Graph2Seq consisting of GNN encoders and sequential decoders to generate neural language outputs. This section summarizes various graph construction methods and the techniques employed to exploit the informative graphs.

\begin{itemize}
\item\textbf{Graph Construction}. 
Regarding the AMR-to-text Generation, 
the input AMRs can be normally represented as  directed heterogeneous graphs according to the relations~\citep{damonte-cohen-2019-structural,song-etal-2020-structural,bai-etal-2020-online,zhu-etal-2019-modeling,zhang-etal-2020-lightweight,yao2020heterogeneous,beck2018graph,cai2020graph,jin-gildea-2020-generalized,ribeiro-etal-2019-enhancing,ijcai2020-542,wang2020amr,song-etal-2018-graph}. To incorporate the conventional GNNs specializing in homogeneous-graph learning, 
\citet{damonte-cohen-2019-structural,yao2020heterogeneous,beck2018graph,cai-lam-2020-amr,ribeiro-etal-2019-enhancing} convert the AMR graphs to levi-graph. In addition, for each edge, they~\citep{damonte-cohen-2019-structural,beck2018graph,yao2020heterogeneous,cai-lam-2020-amr} add the reverse edges and self-loops to allow information flows in both directions. 
Besides the default, reverse, and self-loop edges, \citet{yao2020heterogeneous} also introduces fully-connected edges to model indirect nodes and connected edges, which treat original edges as connectivity without direction to model connection information. 
\citet{zhao-etal-2020-line} split the given AMR graph $\mathcal{G}_{AMR}$ into two directed sub-graphs: 1) concept graph $\mathcal{G}_{c}$, and 2) line graph $\mathcal{G}_{l}$. They firstly treat the edge as connectivity to get the concept graph. Then for each edge in $\mathcal{G}_{AMR}$, they create a node in $\mathcal{G}_{l}$. Two nodes in $\mathcal{G}_{l}$ are connected if they share the same nodes in $\mathcal{G}_{AMR}$. The two sub-graphs are connected by original connections in $\mathcal{G}_{AMR}$. To leverage multi-hop connection information, they preserve the $1-K$ order neighbors in the adjacency matrices.
Regarding the SQL inputs, the SQL queries can be parsed by many SQL tools\footnotemark[1] into many sub-clauses without loss, which naturally contain rich structure information. 
\footnotetext[1]{\href{http://www.sqlparser.com}{http://www.sqlparser.com}.}
\citet{xu-etal-2018-sql,xu2018graph2seq} construct the directed and homogeneous SQL-graph based on the sub-clauses by some hand-craft rules.
Regarding the RDF triple inputs, \citet{marcheggiani-perez-beltrachini-2018-deep,ijcai2020-419} treat the relation in a triple as an additional node in the graph connecting to the subject and object entity nodes.

\item\textbf{Graph Representation Learning}.
\citet{ribeiro-etal-2019-enhancing, ijcai2020-419} treat the obtained levi-graphs as directed homogeneous graphs and learn the representation by bidirectional GNNs. \citet{ribeiro-etal-2019-enhancing} also proposes a bidirectional embedding learning framework that traverses the directed graphs in the original and the reversal direction. 
\citet{xu-etal-2018-sql} apply classic graph2seq architecture~\citep{xu2018graph2seq} with bidirectional GraphSage methods to learn the embedding of SQL graph via two ways, including 1) pooling-based mechanism and 2) node-based mechanism, which means add a supernode connecting to other nodes, to investigate the influence of graph embedding. 
Some approaches directly employ multi-relational GNN to encode the obtained multi-relational graphs. For example, \citet{damonte-cohen-2019-structural} adopt directed-GCN to exploit the AMR graphs considering both heterogeneity and parameter-overhead. \citet{beck2018graph} propose relational GGNN to capture diverse semantic correlations. \citet{song-etal-2018-graph} employ a variance of GGNN to exploit the multi-relational AMR graphs by aggregating the bidirectional node and edge features and then fusing them via a LSTM network. \citet{zhao-etal-2020-line} propose a heterogeneous GAT to exploit the AMR graphs in different grains. Firstly, they apply GAT to each sub-graph to learn the bidirectional representation separately. Then they apply cross-attention to explore the dependencies between the two sub-graphs. \citet{zhang-etal-2020-lightweight} propose the multi-hop GCN, which dynamically fuses the $1-K$ order neighbors' features to control the information propagate in a range of orders. \citet{ijcai2020-542} apply relational GAT with bidirectional graph embedding mechanism by incorporating the edge types into the attention procedure to learn type-specific attention weights.

Transformer architectures are also utilized to encode the AMR or SQL graphs. \citet{yao2018exploring} firstly apply GAT-based graph Transformer in each homogeneous sub-graphs and then concatenate sub-graphs representation to feed the feed-forward layer. Some works~\citet{zhu-etal-2019-modeling,song-etal-2020-structural,bai-etal-2020-online,cai-lam-2020-amr,jin-gildea-2020-generalized} adopt the structure-aware graph transformer~\citep{zhu-etal-2019-modeling,cai2020graph}, which injecting the relation embedding learned by shortest path to the self-attention to involve the structure features. Specifically, \citet{jin-gildea-2020-generalized} explore various shortest path algorithms to learn the relation representation of arbitrary two nodes. Similarly, \citet{wang2020amr} employ the graph Transformer, which leverages the structure information by incorporating the edge types into attention-weight learning formulas.

\item\textbf{Special Mechanisms}.
\citet{damonte-cohen-2019-structural} apply the Bi-LSTM encoder following the GNN encoder to further encode the sequential information. Despite the language generation procedure,  to better preserve the structural information, \citet{zhu-etal-2019-modeling,bai-etal-2020-online,ijcai2020-542} introduce the graph reconstruction on top of the latent graph representation generated by graph transformer encoder.

\end{itemize}

\paragraph{Benchmarks and Evaluation}

Common benchmarks for AMR-to-text generation task include LDC2015E85, LDC2015E86, LDC2017T10, and LDC2020T02. As for the SQL-to-text generation task, WikiSQL~\citep{zhong2017seq2sql} and Stackoverflow~\citep{iyer2016summarizing} are widely used by previous works. The RDF-to-text generation task often uses WebNLG~\citep{gardent2017creating} and New York Times (NYT)~\citep{riedel2010modeling}. As for evaluation metrics, the AMR-to-text generation task mostly adopts BLEU, Meteor, CHRF++, and human evaluation including meaning similarity and readability. While BLEU-4 are widely used for SQL-to-text task. The RDF-to-text generation task uses BLEU, Meteor and TER.

\subsubsection{Natural Question Generation}

\paragraph{Background and Motivation}
The natural question generation (QG) task aims at generating natural language questions from certain form of data, such as KG~\citep{kumar2019difficulty,chen2020toward}, tables~\citep{bao2018table}, text~\citep{du2017learning,song2018leveraging} or images~\citep{li2018visual}, where the generated questions need to be answerable from the input data.
Most prior work~\citep{du2017learning,song2018leveraging,kumar2019difficulty} adopts a Seq2Seq architecture which regards the input data as sequential data without considering its rich structural information.
For instance, when encoding the input text, 
most previous approaches~\citep{du2017learning,song2018leveraging} typically ignore the hidden structural information associated with a word sequence such as the dependency parsing tree.
Even for the setting of QG from KG, most approaches~\citep{kumar2019difficulty} typically linearize the KB subgraph to a sequence and apply a sequence encoder.
Failing to utilize the graph structure of the input data may limit the effectiveness of QG models. 
As for the multi-hop QG from text setting which requires reasoning over multiple paragraphs or documents, it is beneficial to capture the relationships among different entity mentions across multiple paragraphs or documents.
In summary, modeling the rich structures of the input data is important for many QG tasks.
Recently, GNNs have been successfully applied to the QG tasks~\citep{liu2019learning,chen2020reinforcement,DBLP:conf/coling/WangXLZS20}.

\paragraph{Methodologies}
Most GNN-based QG approaches adopt a Graph2Seq architecture where a GNN-based encoder is employed to model the graph-structured input data, and a sequence decoder is employed to generate a natural language question.
In this section, we introduce and summarize some representative GNN-related techniques adopted in recent QG approaches.

\begin{itemize}
\item\textbf{Graph Construction}.
Different graph construction strategies have been proposed to suit the various needs of different QG settings by prior GNN-based approaches. 
Some works~\citep{liu2019learning,DBLP:conf/coling/WangXLZS20,chen2020reinforcement,pan2020semantic} converted the passage text to a graph based on dependency parsing or semantic role labeling for QG from text. 
As for multi-hop QG from text, in order to model the relationships among entity mentions across multiple paragraphs or documents, an entity graph is often constructed. For instance, \citet{su-etal-2020-multi} constructed an entity graph with the named entities in context as nodes and edges connecting the entity pairs appearing in the same sentence or paragraph. In addition, an answer-aware dynamic entity graph was created on the fly by masking out entities irrelevant to the answers.
\citet{sachan2020stronger} built a so-called context-entity graph containing three types of nodes (i.e., named-entity mentions, coreferent entities, and sentence-ids) and added edges connecting them.
Unlike the above approaches that build a static graph based on prior knowledge, 
\citet{chen2020reinforcement} explored dynamic graph construction for converting the passage text to a graph of word nodes by leveraging the attention mechanism. As for QG from KG, graph construction is not needed since the KG is already provided. A common option is to extract a k-hop subgraph surrounding the topic entity as the input graph when generating a question~\citep{chen2020toward}.

\item\textbf{Graph Representation Learning}.
Common GNN models used by existing QG approaches include GCN~\citep{liu2019learning,su-etal-2020-multi}, GAT~\citep{DBLP:conf/coling/WangXLZS20}, GGNN~\citep{chen2020reinforcement,chen2020toward,pan2020semantic}, and graph transformer~\citep{sachan2020stronger}.
In order to model the edge direction information, \citet{chen2020reinforcement} and \citet{chen2020toward} extended the GGNN model to handle directed edges. 
In order to model multi-relational graphs,
\citet{chen2020toward} explored two graph encoding strategies: i) converting a multi-relational graph to a Levi graph~\citep{levi1942finite} and applying a regular GGNN model, or ii) extending the GGNN model by incorporating the edge information in the message passing process.
\citet{pan2020semantic} also extended the GGNN model by bringing in the attention mechanism from GAT and introducing edge type aware linear transformations for message passing between node pairs.
\citet{sachan2020stronger} proposed a graph-augmented transformer model employing a relation-aware multi-head attention mechanism similar to \citet{zhu2019modeling,cai2020graph}.
\citet{pan2020semantic,sachan2020stronger} found it beneficial to additionally model the sequential information in the input text besides the graph-structured information.
\citet{pan2020semantic} separately applied a sequence encoder to the document text, and a graph encoder to the semantic graph representation of the document constructed from semantic role labeling or dependency parsing. 
The outputs of the sequence encoder and graph encoder would then be fused and fed to a sequence decoder for question generation. The model was jointly trained on question decoding and content selection sub-tasks. 
\citet{sachan2020stronger} ran both the structure-aware attention network on the input graph and the standard attention network on the input sequence, and fused their output embeddings using some non-linear mapping function to learn the final embeddings for the sequence decoder. During the training, a contrastive objective was proposed to predict supporting facts, serving as a regularization term in addition to the main cross-entropy loss for sequence generation.

\end{itemize}

\paragraph{Benchmarks and Evaluation}
Common benchmarks for QG from text include 
SQuAD~\citep{rajpurkar2016squad},
NewsQA~\citep{trischler2017newsqa}, and
HotpotQA~\citep{yang2018hotpotqa}.
As for QG from KG, WebQuestions~\citep{kumar2019difficulty} and PathQuestions~\citep{kumar2019difficulty} are widely used by previous works.
As for evaluation metrics, BLEU-4, METEOR, ROUGE-L and human evaluation (e.g., syntactically correct, semantically correct, relevant) are common metrics.
Complexity is also used to evaluate the performance of multi-hop QG systems.

\subsection{Machine Reading Comprehension and Question Answering}

\subsubsection{Machine Reading Comprehension}
\paragraph{Background and Motivation}
The task of Machine Reading Comprehension (MRC) aims to answer a natural language question using the given passage.
Significant progress has been made in the MRC
task thanks to the development of various (co-)attention mechanisms that capture the interaction between the question and context~\citep{hermann2015teaching,cui2017attention,DBLP:conf/iclr/SeoKFH17,DBLP:conf/iclr/XiongZS17}.
Considering that the traditional MRC setting mainly focuses on one-hop reasoning which is relatively simple, recently, more research efforts have been made to solve more challenging MRC settings.
For instance, the multi-hop MRC task is to answer a natural language question using multiple passages or documents, which requires the multi-hop reasoning capacity.
The conversational MRC task is to answer the current natural language question in a conversation given a passage and the previous questions and answers, which requires the capacity of modeling conversation history.
The numerical MRC task requires the capacity of performing numerical reasoning over the passage.
These challenging MRC tasks call for the learning capacity of modeling complex relations among objects.
For example, it is beneficial to model relations among multiple documents and the entity mentions within the documents for the multi-hop MRC task.
Recently, GNNs have been successfully applied to various types of MRC tasks including multi-hop MRC~\citep{song2018exploring,DBLP:conf/naacl/CaoAT19,qiu-etal-2019-dynamically,DBLP:conf/naacl/CaoFT19,DBLP:conf/emnlp/FangSGPWL20,DBLP:conf/ijcai/TangSMXYL20,zheng2020srlgrn,DBLP:conf/acl/TuWHTHZ19,DBLP:conf/acl/DingZCYT19}, conversational MRC~\citep{chen2020graphflow}, and numerical MRC~\citep{DBLP:conf/emnlp/RanLLZL19}.

\paragraph{Methodologies}
GNN-based MRC approaches typically operate by first constructing an entity graph or hierarchical graph capturing rich relations among nodes in the graph, and then applying a GNN-based reasoning module for performing complex reasoning over the graph. Assuming the GNN outputs already encode the semantic meanings of the node itself and its neighboring structure, a prediction module will finally be applied for predicting answers.
The graph construction techniques and graph representation techniques developed for solving the MRC task vary between different approaches.
In this section, we introduce and summarize some representative GNN-related techniques adopted in recent MRC approaches.

\begin{itemize}

\item\noindent\textbf{Graph Construction}.
In order to apply GNNs for complex reasoning in the MRC task, one critical step is graph construction. 
Building a high-quality graph capturing rich relations among useful objects (e.g., entity mentions, paragraphs) is the foundation for conducting graph-based complex reasoning.
Most GNN-based MRC approaches conduct static graph construction by utilizing domain-specific prior knowledge. 
Among all existing GNN-based MRC approaches, the most widely adopted strategy for static graph construction is to construct an entity graph using carefully designed rules.
These approaches~\citep{song2018exploring,DBLP:conf/naacl/CaoAT19,qiu-etal-2019-dynamically,DBLP:conf/naacl/CaoFT19,DBLP:conf/ijcai/TangSMXYL20,zheng2020srlgrn,DBLP:conf/emnlp/RanLLZL19} usually extract entity mentions from questions, paragraphs and candidate answers (if given) as nodes, and connect the nodes with edges capturing different types of relations such as exact match, co-occurrence, coreference and semantic role labeling.
Edge connectivity with different granularity levels in terms of context window (e.g., sentence, paragraph and document) might also be distinguished for better modeling performance~\citep{qiu-etal-2019-dynamically,DBLP:conf/naacl/CaoFT19}.
For instance, \citet{DBLP:conf/naacl/CaoFT19} distinguished cross-document edge and within-document edge when building an entity graph.
As for the numerical MRC task, the most important relations are probably the arithmetic relations. In order to explicitly model numerical reasoning, \citet{DBLP:conf/emnlp/RanLLZL19} constructed a graph containing numbers in the question and passage as nodes, and added edges to capture various arithmetic relations among the numbers.
Besides building an entity graph capturing various types of relations among entity mentions, some approaches~\citep{DBLP:conf/acl/TuWHTHZ19,DBLP:conf/emnlp/FangSGPWL20,DBLP:conf/acl/ZhengWLDCJZL20} opt to build a hierarchical graph containing various types of nodes including entity mentions, sentences, paragraphs and documents, and connect these nodes using predefined rules.
For example, \citet{DBLP:conf/acl/ZhengWLDCJZL20} constructed a hierarchical graph that contains edges connecting token nodes and sentence nodes, sentence nodes and paragraph nodes as well as paragraph nodes and document nodes.

Very recently, dynamic graph construction techniques without relying on hand-crafted rules have also been explored for the MRC task and achieved promising results. Unlike static graph construction techniques that have been widely explored in the MRC literature, dynamic graph construction techniques are less studied.
In comparison to static graph construction, dynamic graph construction aims to build a graph on the fly without relying on domain-specific prior knowledge, and is typically jointly learned with the remaining learning modules of the system. 
Recently, \citet{chen2020graphflow} proposed a GNN-based model for the conversational MRC task, which is able to dynamically build a question and conversation history aware passage graph containing each passage word as a node at each conversation turn by leveraging the attention mechanism. 
A kNN-style graph sparsification operation was conducted so as to further extract a sparse graph from the fully-connected graph learned by the attention mechanism.
The learned sparse graph will be consumed by the subsequent GNN-based reasoning module, and the whole system is end-to-end trainable.

\item\textbf{Graph Representation Learning}.
Most GNN-based MRC approaches rely on a GNN model for performing complex reasoning over the graph. 
In the literature of the MRC task, both homogeneous GNNs and multi-relational GNNs have been explored for node representation learning.
Even though most GNN-based MRC approaches construct a multi-relational or heterogeneous graph, some of them still apply a homogeneous GNN model such as GCN~\citep{zheng2020srlgrn,DBLP:conf/acl/DingZCYT19}, GAT~\citep{qiu-etal-2019-dynamically,DBLP:conf/emnlp/FangSGPWL20,DBLP:conf/acl/ZhengWLDCJZL20} and Graph Recurrent Network (GRN)~\citep{song2018exploring}.
Unlike other works that apply a GNN model to a single graph, 
\citet{chen2020graphflow} proposed a Recurrent Graph Neural Network (RGNN) for processing a sequence of passage graphs for modeling conversational history.
The most widely used multi-relational GNN model in the MRC task is the RGCN model~\citep{schlichtkrull2018modeling}.
Many approaches~\citep{DBLP:conf/naacl/CaoAT19,DBLP:conf/naacl/CaoFT19,DBLP:conf/acl/TuWHTHZ19,DBLP:conf/emnlp/RanLLZL19} adopt a gating RGCN variant which in addition introduces a gating mechanism regulating how much of the update message propagates to the next step.
\citet{DBLP:conf/ijcai/TangSMXYL20} further proposed a question-aware gating mechanism for RGCN, that is able to regulate the aggregated message according to the question, and even bring the question information into the update message.


 




\item\textbf{Node Embedding Initialization}.
Many studies have shown that the quality of the initial node embeddings play an important role in the overall performance of GNN-based models.
Most approaches use pre-trained word embeddings such as GloVe~\citep{pennington2014glove}, ELMo~\citep{peters2018deep}, BERT~\citep{devlin-etal-2019-bert} and RoBERTa~\citep{liu2019roberta} to initialize tokens.
Some works~\citep{DBLP:conf/naacl/CaoFT19,chen2020graphflow} also concatenated linguistic features to word embeddings to enrich the semantic meanings.
On top of the initial word embeddings, most approaches choose to further apply some transformation functions such as MLP for introducing nonlinearity~\citep{DBLP:conf/ijcai/TangSMXYL20}, 
BiLSTM for capturing local dependency of the text~\citep{DBLP:conf/naacl/CaoAT19,chen2020graphflow,DBLP:conf/emnlp/FangSGPWL20}, co-attention layer for fusing questions to passages~\citep{qiu-etal-2019-dynamically,DBLP:conf/acl/TuWHTHZ19,chen2020graphflow,DBLP:conf/emnlp/FangSGPWL20}.


\item\textbf{Special Techniques}
 In order to increase the richness of the supervision signals, some approaches adopt the multi-tasking learning strategy to predict not only the answer span, but also the supporting paragraph/sentence/fact and answer type~\citep{qiu-etal-2019-dynamically,DBLP:conf/emnlp/FangSGPWL20,zheng2020srlgrn,chen2020graphflow}.
 
\end{itemize}






\paragraph{Benchmarks and Evaluation}
Common multi-hop MRC benchmarks include HotpotQA~\citep{yang2018hotpotqa}, WikiHop~\citep{welbl2018constructing} and ComplexWebQuestions~\citep{talmor2018web}.
Common conversational MRC benchmarks include CoQA~\citep{reddy2019coqa}, QuAC~\citep{choi2020quac} and DoQA~\citep{campos2019conversational}.
DROP~\citep{dua2019drop} is a benchmark created for the numerical MRC task.
As for evaluation metrics, F1 and EM (i.e., exact match) are the two most widely used evaluation metrics for the MRC task.
Besides, the Human Equivalence Score (HEQ)~\citep{choi2020quac,campos2019conversational} is used to judge whether a system performs as well as an average human. HEQ-Q and HEQ-D are accuracies at the question level and dialog level, respectively.

\subsubsection{Knowledge Base Question Answering}

\paragraph{Background and Motivation}
Knowledge Base Question Answering (KBQA) has emerged as an important research topic in the past few years ~\citep{yih2015semantic,zhang2018variational,chen2019bidirectional}.
The goal of KBQA is to automatically find answers from the KG given a natural language question.
Recently, due to its nature capability of modeling relationships among objects, GNNs have been successfully applied for performing the multi-hop KBQA task which requires reasoning over multiple edges of the KG to arrive at the right answer.
A relevant task is open domain QA~\citep{sun-etal-2018-open,sun-etal-2019-pullnet} which aims to answer open domain questions by leveraging hybrid knowledge sources including corpus and KG. 
Here we only focus on the QA over KG setting, while the GNN applications in the open domain QA task will be introduced in other sections.

\paragraph{Methodologies}
In this section, we introduce and summarize some representative GNN-related techniques adopted in the recent KBQA research.

\begin{itemize}
    \item\textbf{Graph Construction}.
Semantic parsing based KBQA methods~\citep{yih2015semantic} aims at converting natural language questions to a semantic graph which can be further executed against the KG to find the correct answers.
In order to better model the structure of the semantic graph, 
\citet{DBLP:conf/coling/SorokinG18} proposed to use GNNs to encode the candidate semantic graphs. Specifically, they used a similar procedure as \citet{yih2015semantic} to construct multiple candidate semantic graphs given the question, and chose the one which has the highest matching score to the question in the embedding space.
\citet{DBLP:conf/emnlp/FengCLWYR20,yasunaga2021qa} focused on a different multi-choice QA setting which is to select the correct answer from the provided candidate answer set given the question and the external KG.
For each candidate answer, \citet{DBLP:conf/emnlp/FengCLWYR20} proposed to extract from the external KG a ``contextualized'' subgraph according to the question and candidate answer.
This subgraph serves as the evidence for selecting the corresponding candidate answer as the final answer.
Specifically, they first recognized all the entity mentions in the question and candidate answer set, and linked them to entities in the KG. 
Besides these linked entities in the KG, any other 
entities that appeared in any two-hop paths between pairs of mentioned entities in the KG as well as the corresponding edges were also added to the subgraph.
\citet{yasunaga2021qa} constructed a joint graph by regarding the QA context as an additional node (QA context node) and connecting it to the topic entities in the KG subgraph. Specifically, they introduced two new relation types $r_{z,q}$ and $r_{z,a}$ for capturing the relationship between the QA context node and the relevant entities in the KG. The specific relation type is determined by whether the KG entity is found in the question portion or the answer portion of the QA context.


\item\textbf{Graph Representation Learning}
In order to better model the constructed multi-relational or heterogeneous graphs, basic GNNs need to be extended to handle edge types or node types.
To this end, \citet{DBLP:conf/coling/SorokinG18} extended GGNN~\citep{DBLP:conf/acl/ZhangYCWWW20} to include edge embeddings in message passing. After learning the vector representations of both the question and every candidate semantic graph, they used a simple reward function to select the best semantic graph for the question.
The final node embedding of the question variable node (q-node) in each semantic graph was extracted and non-linearly transformed to obtain the graph-level representation.
\citet{DBLP:conf/emnlp/FengCLWYR20} designed a Multi-hop Graph Relation Network (MHGRN) to unify both GNNs and path-based models.
Specifically, they considered both node type and edge type information of the graph by introducing node type specific linear transformation, and node type and relation type aware attention in message passing.
In addition, instead of performing one hop message passing at each time, inspired by path-based models~\citep{santoro2017simple,lin2019kagnet}, they proposed to pass messages directly over all the paths of lengths up to $K$.
Graph-level representations were obtained via attentive pooling over the output node embeddings, and would be concatenated with the text representation of question and each candidate answer to compute the plausibility score.
Similarly, \citet{yasunaga2021qa} extended GAT by introducing node type and edge type aware message passing to handle multi-relational graphs. They in addition employed a pre-trained language model for KG node relevance scoring in the initial stage and final answer selection stage. 
\end{itemize}

\paragraph{Benchmarks and Evaluation}
Common benchmarks for KBQA include 
WebQuestionsSP~\citep{yih2016value},
MetaQA~\citep{zhang2018variational},
QALD-7~\citep{usbeck20177th},
CommonsenseQA~\citep{talmor2019commonsenseqa}, and
OpenbookQA~\citep{mihaylov2018can}.
F1 and accuracy are common metrics for evaluating KBQA methods.

\subsubsection{Open-domain Question Answering}
\paragraph{Background and Motivation} The task of open-domain question answering aims to identify answers to the natural question given a large scale of open-domain knowledge (e.g. documents, knowledge base and etc.). Untill recent times, the open-domain question answering~\citep{bordes2015large,zhang2018variational} has been mostly exploited through knowledge bases such as Personalized PageRank~\citep{10.1145/511446.511513}, which actually closely related to the Knowledge based Question Answering task (KBQA) in techniques. The knowledge based methods benefit from obtaining external knowledge easily through graph structure. However, these methods limit in the missing information of the knowledge base and fixed schema. Other attempts have been made to answer questions from massive and unstructured documents~\citep{chen2017reading}. Compared to the KB based methods, these methods can fetch more information but suffer from the difficulty of retrieve relevant and key information from redundant external documents.

\paragraph{Methodologies} In this section, we introduce and summarize some representative GNN-related techniques in the recent open-domain question answering research.

\begin{itemize}
    \item\textbf{Graph Construction}. Most of the GNN based methods address the mentioned challenges by constructing a heterogeneous graph with both knowledge base and unstructured documents~\citep{han-etal-2020-open,sun-etal-2018-open,sun-etal-2019-pullnet}. \citet{han-etal-2020-open,sun-etal-2018-open} firstly extract the subgraph from external knowledge base named Personalized PageRank~\citep{10.1145/511446.511513}. Then they fetch a relevant text corpus from Wikipedia and fuse them to the knowledge graph. Specifically, they represent the documents by words' encoding and link the nodes (the nodes in the knowledge graph are entities) which appear in the document. \citet{sun-etal-2019-pullnet} propose a iteratively constructed heterogeneous graph method from both knowledge base and text corpus. Initially, the graph depends only on the question. Then for each iteration, they expand the subgraph by choosing nodes from which to "pull" information about, from the KB or corpus as appropriate. 

\item\textbf{Graph Representation Learning} \citet{sun-etal-2018-open,sun-etal-2019-pullnet} first initialize the nodes' embedding with pre-trained weight for entities and LSTM encoding for documents. They further propose different update rule for both entities and documents. For entities, they apply R-GCN~\citep{schlichtkrull2018modeling} on the sub-graph only from the knowledge base and then take average of the linked words' feature in the connected documents. The entities' representation is the combination of: 1) the previous entities themselves' representation, 2) question encoding, 3) knowledge-subgraph's aggregation results, and 4) related documents' aggregation results. For documents' update operation, they aggregate the features from connected entities. \citet{sun-etal-2018-open} adopt similar idea for heterogeneous graph representation learning. Technically, before encoding entities, they incorporate the connected words' embedding in the documents to the entities. Then for nodes, they propose GCN with attention weight to aggregate neighbor entities. Note that the question is employed in the attention mechanism to guide the learning process. The documents' updating process is in the same pattern.
 
\end{itemize}

\paragraph{Benchmarks and Evaluation}
Common benchmarks for Open-domain Question answering include WebQuestionsSP~\citep{yih2016value}, MetaQA~\citep{zhang2018variational}, Complex WebQuestions 1.1 (Complex WebQ)~\citep{talmor2018web}, and WikiMovies-10K~\citep{miller-etal-2016-key}. Hits@1 and F1 scores are the common evaluation metrics for this task~\citep{sun-etal-2018-open,sun-etal-2019-pullnet,han-etal-2020-open}.

\subsubsection{Community Question Answering}
\paragraph{Background and Motivation} The task of community question answering aims to retrieve the relevant answer from QA forums such as Stack Overflow or Quora. Different from the traditional MRC (QA) task, CQA systems are able to harness tacit knowledge (embedded in their diverse communities) or explicit knowledge (embedded in all resolved questions) in answering of an enormous number of new questions posted each day. Nevertheless, the growing number of new questions could make CQA systems without appropriate collaboration support become overloaded by users’ requests.

\paragraph{Methodologies} In this section, we introduce and summarize some representative GNN-related techniques adopted in the recent CQA research.
\begin{itemize}
    \item\textbf{Graph Construction}. Most of the GNN-based methods construct a multi-modal graph for existing question/answer pairs~\citep{10.1145/3343031.3350966,10.1145/3394171.3413711}. For the given q/a pair (q, a), both of them construct the question/answer $\mathcal{G}^q/\mathcal{G}^a$ graph separately. Since in real community based forums, the question/answer pairs may contain both visual and text contents, they employ a multi-modal graph to represent them jointly. \citet{10.1145/3343031.3350966} firstly employ object detection models such as YOLO3~\citep{redmon2018yolov3} to fetch visual objects. The objects are represented by their labels (visual words more accurately). The visual objects are treated as words in the answers which are modeled with textural contents equally. Then they regard each textural words as vertex and link them with undirected occurrence edges. \citet{10.1145/3394171.3413711} adopt the same idea as~\citep{10.1145/3343031.3350966} for building occurrence graph for both textural contents and visual words. But for extracting visual words from images, they employ unsupervised Meta-path Link Prediction for Visual Labeling. Concretely, they define the meta-path over image and words and build the heterogeneous image-word graph. 

\item\textbf{Graph Representation Learning}. Most of the GNN-based community question answering models adapt the GNN models to capture structure information. Given the question/answer pair (q, a), \citet{10.1145/3343031.3350966} stacks the graph pooling network to capture the hierarchical semantic-level correlations between nodes. Conceptually, the graph pooling network extract the high-level semantic representation for both question and answer graphs. Formally, it consists of two GCN-variant APPNP~\citep{klicpera2018combining} encoders. Generally, one APPNP is employed to learn the high-level semantic cluster distribution for each vertex. The other APPNP network is used to learn the immediate node representation. The final node representation is the fusion of the two encoders' results.  \citet{10.1145/3394171.3413711} employ the APPNP to learn the importance of each vertex's neighbors. 

\end{itemize}

\paragraph{Benchmarks and Evaluation} Common benchmarks for Community Question Answering include Zhihu and Quora released by MMAICM~\citep{10.1145/3240508.3240626}. The normalized discounted cumulative gain (nDCG) and precision are common metrics for evaluating Community Question Answering methods~\citep{10.1145/3240508.3240626,10.1145/3394171.3413711,10.1145/3343031.3350966}.

\subsection{Dialog Systems}

\paragraph{Background and Motivation}

Dialog system~\citep{williams2014dialog,chen2017survey} is a computer system that can continuously converse with a human.
In order to build a successful dialog system, it is important to model the dependencies among different interlocutors or utterances within a conversation.
Due to the ability of modeling complex relations among objects, recently, GNNs have been successfully applied to various dialog system related tasks including dialog state tracking~\citep{DBLP:conf/coling/ChenTLY18,DBLP:conf/aaai/0002LWZT020} which aims at estimating the current dialog state given the conversation history, dialog response generation~\citep{DBLP:conf/ijcai/HuCL0MY19} which aims at generating the dialog response given the conversation history, and 
next utterance selection~\citep{liu2021graph} which aims at selecting the next utterance from a candidate list given the conversation history.


\paragraph{Methodologies}
In this section, we introduce and summarize some representative GNN-related techniques adopted in the recent dialog systems research.

\begin{itemize}
\item\textbf{Graph Construction}.
Building a high-quality graph representing a structured conversation session is challenging.
A real-world conversation can have rich interactions among speakers and utterances. 
Here, we introduce both static and dynamic graph construction techniques used in recent GNN-based approaches.
For static graphs, most GNN-based dialog systems rely on prior domain knowledge to construct a graph.
For instance, in order to apply GNNs to model  multi-party dialogues (i.e., involving multiple interlocutors), \citet{DBLP:conf/ijcai/HuCL0MY19} converted utterances in a structured dialogue session to a directed graph capturing response relationships between utterances. Specifically, they created an edge for every pair of utterances from the same speaker following the chronological order of the utterances.
\citet{DBLP:conf/coling/ChenTLY18} built a directed heterogeneous graph according to the domain ontology that consists of edges among slot-dependent nodes and slot-independent nodes.
\citet{DBLP:conf/aaai/0002LWZT020} constructed three types of graphs including a token-level schema graph according to the original ontology scheme, a utterance graph according to the dialogue utterance, and a domain-specific slot-level schema graph connecting two slots from the same domain or share the same candidate values.
\citet{liu2021graph} constructed a graph connecting utterance nodes that are adjacent or belong to dependent topics.
Regarding the dynamic graph construction,
unlike most GNN-based approaches that rely on prior knowledge for constructing static graph, \citet{DBLP:conf/coling/ChenTLY18} jointly optimized the graph structure and the parameters of GNN by approximating posterior probability of the adjacency matrix (i.e., modeled as a latent variable following a factored Bernoulli distribution) via variational inference~\citep{hoffman2013stochastic}.

\item\textbf{Graph Representation Learning}
Various GNN models have been applied in dialog systems. For instance, \citet{liu2021graph} applied GCN to facilitate reasoning over all utterances.
\citet{DBLP:conf/aaai/0002LWZT020} proposed a graph attention matching network to learn the representations of ontology schema and dialogue utterance simultaneously, and a recurrent attention graph neural network which employs a GRU-like gated cell for dialog state updating.
Inspired by the hierarchical sequence-based HRED model for dialog response generation, \citet{DBLP:conf/ijcai/HuCL0MY19} proposed an utterance-level graph-structured encoder which is a gated GNN variant, and is able to control how much the new information (from the preceding utterance nodes) should be considered when updating the current state of the utterance node. They also designed a bi-directional information flow algorithm to allow both forward and backward message passing over the directed graph.
In order to model multi-relational graphs,
\citet{DBLP:conf/coling/ChenTLY18} designed a R-GCN like GNN model employing edge type specific weight matrices.



\item\textbf{Node Embedding Initialization}
In terms of node embedding initialization for GNN models, \citet{DBLP:conf/ijcai/HuCL0MY19} applied a BiLSTM to first encode the local dependency information in the raw text sequence.
\citet{liu2021graph} used the state-of-the-art pre-trained ALBERT embeddings~\citep{lan2019albert} to initialize the node embeddings.
\citet{DBLP:conf/aaai/0002LWZT020} included token embeddings, segmentation embeddings as well as position embeddings to capture the rich semantic meanings of the nodes.

\end{itemize}


\paragraph{Benchmarks and Evaluation}
Common dialog state tracking benchmarks include
PyDial~\citep{casanueva2017benchmarking} and MultiWOZ~\citep{budzianowski2018multiwoz,eric2020multiwoz}.
Ubuntu Dialogue Corpus~\citep{lowe2015ubuntu} and MuTual~\citep{cui2020mutual} are often used for evaluating dialog response generation and next utterance selection systems.
As for evaluation metrics, BLEU, METEOR and ROUGE-L are common metrics for evaluating dialog response generation systems. Besides automatic evaluation, human evaluation (e.g., grammaticality, fluency, rationality) is often conducted.
Accuracy is the most widely used metric for evaluating dialog state tracking systems.
Recall at k is often used in the next utterance selection task.




\subsection{Text Classification}

\paragraph{Background and Motivation}
Traditional text classification methods heavily rely on feature engineering (e.g., BOW, TF-IDF or more advanced graph path based features) for text representation.
In order to learn ``good'' representations from text, various unsupervised approaches have been proposed for word or document representation learning, including word2vec~\citep{DBLP:conf/nips/MikolovSCCD13}, GloVe~\citep{pennington2014glove}, topic models~\citep{blei2003latent,larochelle2012neural}, autoencoder~\citep{miao2016neural,chen2017kate}, and
doc2vec~\citep{le2014distributed,kiros2015skip}.
These pre-trained word or document embeddings can further be consumed by a MLP~\citep{joulin2017bag}, CNN~\citep{DBLP:conf/emnlp/Kim14} or LSTM~\citep{liu2016recurrent,zhang2018sentence} module for training a supervised text classifier.
In order to better capture the relations among words in text or documents in corpus, various graph-based approaches have been proposed for text classification.
For instance, \citet{peng2018large} proposed to first construct a graph of words, and then apply a CNN to the normalized subgraph.
\citet{tang2015pte} proposed a network embedding based approach for text representation learning in a semi-supervised manner by converting a partially labeled text corpora to a heterogeneous text network.
Recently, given the strong expressive power, GNNs have been successfully applied to both semi-supervised~\citep{DBLP:conf/aaai/YaoM019,DBLP:conf/aaai/LiuYZWL20,DBLP:conf/emnlp/HuYSJL19} and supervised~\citep{defferrard2016convolutional,DBLP:conf/emnlp/HuangMLZW19,DBLP:conf/acl/ZhangYCWWW20} text classification.






\paragraph{Methodologies}
GNN-based text classification approaches typically operate by first constructing a document graph or corpus graph capturing rich relations among nodes in the graph, and then applying a GNN to learn good document embeddings which will later be fed into a softmax layer for producing a probabilistic distribution over a class of labels. 
The graph construction techniques and graph representation techniques developed for solving the text classification task vary between different approaches.
In this section, we introduce and summarize some representative GNN-related techniques adopted in recent text classification approaches.

\begin{itemize}
    \item\textbf{Graph Construction}.
Semi-supervised text classification leverages a small amount of labeled data with a large amount of unlabeled data during training.
Utilizing the relations among labeled and unlabeled documents is essential for performing well in this semi-supervised setting.    
Regarding the static graph construction, recently, many GNN-based semi-supervised approaches~\citep{DBLP:conf/aaai/YaoM019,DBLP:conf/aaai/LiuYZWL20,DBLP:conf/emnlp/HuYSJL19} have been proposed for text classification to better model the relations among words and documents in the corpus.
These approaches typically construct a single heterogeneous graph for the whole corpus containing word nodes and document nodes, and connect the nodes with edges based on word co-occurrence and document-word relations~\citep{DBLP:conf/aaai/YaoM019,DBLP:conf/aaai/LiuYZWL20}.
\citet{DBLP:conf/emnlp/HuYSJL19} proposed to enrich the semantics of the short text with additional information (i.e., topics and entities), and constructed a Heterogeneous Information Network (HIN) containing document, topic and entity nodes with document-topic, document-entity and entity-entity edges based on several predefined rules.
One limitation of semi-supervised text classification is its incapability of handling unseen documents in the testing phase.
In order to handle the inductive learning setting, some GNN-based approaches~\citep{defferrard2016convolutional,DBLP:conf/emnlp/HuangMLZW19,DBLP:conf/acl/ZhangYCWWW20} proposed to instead build an individual graph of unique words for each document by leveraging word similarity or co-occurrence between words within certain fixed-sized context window.
In comparison to static graph construction, dynamic graph construction does not rely on domain-specific prior knowledge, and the graph structure can be jointly learned with the remaining learning modules of the system. 
\citet{henaff2015deep} proposed to jointly learn a graph of unique words for each input text using a Gaussian kernel.
\citet{chen2020iterative} proposed to regard each word in text as a node in a graph, and dynamically build a graph for each document.










\item\textbf{Graph Representation Learning}
Early graph-based text classification approaches~~\citep{henaff2015deep,defferrard2016convolutional} were motivated by extending CNNs to graph CNNs which can directly model graph-structured textual data.
With the fast growth of the GNN research, recent work started to explore various GNN models for text classification including GCN~\citep{DBLP:conf/aaai/YaoM019,chen2020iterative}, GGNN~\citep{DBLP:conf/acl/ZhangYCWWW20} and message passing mechanism (MPM)~\citep{DBLP:conf/emnlp/HuangMLZW19}.
\citet{DBLP:conf/aaai/LiuYZWL20} introduced a TensorGCN which first performs intra-graph convolution propagation and then performs inter-graph convolution propagation.
\citet{DBLP:conf/emnlp/HuYSJL19} proposed a Heterogeneous GAT (HGAT) based on a dual-level (i.e., node-level and type-level) attention mechanism.









\item\textbf{Node Embedding Initialization}
Node embedding initialization is critical for the GNN performance.
Interestingly, \citet{DBLP:conf/aaai/YaoM019} observed in their experiments that by using only one-hot representation, a vanilla GCN~\citep{DBLP:conf/aaai/YaoM019} without any external word embeddings or knowledge already outperformed state-of-the-art methods for text classification.
Nevertheless, most GNN-based approaches~\citep{defferrard2016convolutional,DBLP:conf/emnlp/HuYSJL19,DBLP:conf/emnlp/HuangMLZW19,DBLP:conf/aaai/LiuYZWL20,DBLP:conf/acl/ZhangYCWWW20} still use pre-trained word embeddings to initialize node embeddings.
\citet{chen2020iterative} further applied a BiLSTM to a sequence of word embeddings to capture the contextual information of text for node embedding initialization.





\item\textbf{Special Techniques} 
As a common trick used in text classification, some GNN-based approaches removed stop words during preprocessing~\citep{henaff2015deep,DBLP:conf/aaai/YaoM019,DBLP:conf/acl/ZhangYCWWW20}.

\end{itemize}

\paragraph{Benchmarks and Evaluation}
Common benchmarks for evaluating text classification methods include
20NEWS~\citep{lang1995newsweeder},
Ohsumed~\citep{hersh1994ohsumed},
Reuters~\citep{lewis2004rcv1},
Movie Review (MR)~\citep{pang2004sentimental},
AGNews~\citep{zhang2015character},
Snippets~\citep{phan2008learning},
TagMyNews~\citep{vitale2012classification}, and
Twitter\footnote{\url{https://www.nltk.org/}}.
Accuracy is the most common evaluation metric.

\subsection{Text Matching}

\paragraph{Background and Motivation}
Most existing text matching approaches operate by mapping each text into a latent embedding space via some neural networks such as CNNs~\citep{hu2014convolutional,pang2016text} or RNNs~\citep{wan2016deep}, and then computing the matching score based on the similarity between the text representations.
In order to model the rich interactions between two texts at different granularity levels, sophisticated attention or matching components are often carefully designed~\citep{lu2013deep,palangi2016deep,yang2016anmm}.
Recently, there are a few works~\citep{DBLP:conf/acl/ChenZLJCZY20,DBLP:conf/acl/LiuNWGHLX19} successfully exploring GNNs for modeling the complicated interactions between text elements in the text matching literature.


\paragraph{Methodologies}
In this section, we introduce and summarize some representative GNN-related techniques adopted in recent text matching approaches.

\begin{itemize}
    \item\textbf{Graph Construction}
Chinese short text matching heavily relies on the quality of word segmentation. 
Instead of segmenting each sentence into a word sequence during preprocessing which can be erroneous, ambiguous or inconsistent, \citet{DBLP:conf/acl/ChenZLJCZY20} proposed to construct a word lattice graph from all possible segmentation paths. Specifically, the word lattice graph contains all character subsequences that match words in a lexicon as nodes, and adds an edge between two nodes if they are adjacent in the original sentence. As a result, the constructed graph encodes multiple word segmentation hypotheses for text matching.
In order to tackle long text matching, \citet{DBLP:conf/acl/LiuNWGHLX19} proposed to organize documents into a graph of concepts (i.e., a keyword or a set of highly correlated keywords in a document), and built a concept interaction heterogeneous graph that consists of three types of edges including keyword-keyword, keyword-concept and sentence-concept edges.
Specifically, they first constructed a keyword co-occurrence graph, and based on that, they grouped keywords into concepts by applying community detection algorithms on the keyword co-occurrence graph.
Finally, they assigned each sentence to the most similar concept.


\item\textbf{Graph Representation Learning}
\citet{DBLP:conf/acl/LiuNWGHLX19} applied a GCN model to learn meaningful node embeddings in the constructed graph.
\citet{DBLP:conf/acl/ChenZLJCZY20} designed a GNN-based graph matching module which allows bidirectional message passing across nodes in both text graphs.
In order to obtain graph-level embeddings from the learned node embeddings, max pooling~\citep{DBLP:conf/acl/LiuNWGHLX19} or attentive pooling~\citep{DBLP:conf/acl/ChenZLJCZY20} techniques were adopted.

\item\textbf{Node Embedding Initialization}
As for node embedding initialization, \citet{DBLP:conf/acl/ChenZLJCZY20} used the BERT embeddings while \citet{DBLP:conf/acl/LiuNWGHLX19} first computed a match vector for each node in the graph.

\end{itemize}

\paragraph{Benchmarks and Evaluation}
Common benchmarks for text matching include 
LCQMC~\citep{liu2018lcqmc},
BQ~\citep{chen2018bq},
CNSE~\citep{DBLP:conf/acl/LiuNWGHLX19}, and
CNSS~\citep{DBLP:conf/acl/LiuNWGHLX19}.
Accuracy and F1 are the most widely used evaluation metrics.

\subsection{Topic Modeling}

\paragraph{Background and Motivation}
The task of topic modeling aims to discover the abstract ``topics'' that emerge in a corpus. Typically, a topic model learns to represent a piece of text as a mixture of topics where a topic itself is represented as a mixture of words from a vocabulary.
Classical topic models include graphical model based methods~\citep{blei2003latent,blei2010nested},
autoregressive model based methods~\citep{larochelle2012neural}, and autoencoder based methods~\citep{miao2016neural,chen2017kate,isonuma-etal-2020-tree}.
Recent works~\citep{zhu_graphbtm_2018,zhou_neural_2020,yang_graph_2020} have explored GNN-based methods for topic modeling by explicitly modeling the relationships between documents and words.


\paragraph{Methodologies}
In this section, we introduce and summarize some representative GNN-related techniques adopted in recent topic modeling approaches.

\begin{itemize}
    \item\textbf{Graph Construction}
How to construct a high-quality graph which naturally captures useful relationships between documents and words is the most important for GNN applications in the topic modeling task. Various graph construction strategies have been proposed for GNN-based topic models.
In order to explicitly model the word co-occurrence, \citet{zhu_graphbtm_2018} extracted the biterms (i.e., word pairs) within a fixed-length text window for every document from a sampled mini-corpus, and built an undirected biterm graph where each node represents a word, and each edge weight indicates the frequency of the corresponding biterm in the mini-corpus.
\citet{zhou_neural_2020} built a graph containing documents and words in the corpus as nodes, and added edges to connect document nodes and word nodes based on co-occurrence information where the edge weight matrix is basically the TF-IDF matrix.
\citet{yang_graph_2020} converted a corpus to a bi-partite graph containing document nodes and word nodes, and the edge weight indicates the frequency of the word in the document.

\item\textbf{Graph Representation Learning}
Given a graph representation of the corpus, \citet{zhu_graphbtm_2018} designed a GCN-based autoencoder model to reconstruct the input biterm graph. In addition, residual connections were introduced to the GCN architecture so as to avoid oversmoothing when stacking many GCN layers. 
Similarly, \citet{zhou_neural_2020} designed a GCN-based autoencoder
model to restore the original document representations. 
Notably, \citet{zhu_graphbtm_2018,zhou_neural_2020} reused the adjacency matrix as the node feature matrix which captures the word co-occurrence information.
During the inference time, for both of the autoencoder-based methods, the weight matrix of the decoder network can be interpreted as the (unnormalized) word distributions of the learned topics.
Given the observations that Probabilistic Latent Semantic Indexing (pLSI)~\citep{hofmann1999probabilistic} can be interpreted as stochastic block model (SBM)~\citep{abbe2017community} on a specific bi-partite graph, and GAT can be interpreted as the semi-amortized inference of SBM, \citet{yang_graph_2020} proposed a GAT-based topic modeling network to model the topic structure of non-i.i.d documents.
As for node embedding initialization, they used pre-trained word embeddings to initialize word node features, and term frequency vectors to initialize document node features.

\end{itemize}

\paragraph{Benchmarks and Evaluation}
Common benchmarks for the topic modeling task include
20NEWS ~\citep{lang1995newsweeder}, All News~\citep{thompson2017all}, Grolier~\citep{wang2019atm}, NYTimes~\citep{wang2019atm}, and Reuters~\citep{lewis2004rcv1}.
As for evaluation metrics, since it is challenging to annotate the ground-truth topics for a document, topic coherence score and case study on learned topics are typical means of judging the quality of the learned topics. 
Besides, with the topic representations of text output by a topic model, the performance on downstream tasks such as text classification can also be used to evaluate topic models.

\subsection{Sentiment Classification}

\paragraph{Background and Motivation}

The sentiment classification task aims to detect the sentiment (i.e., positive, negative or neutral) of a piece of text~\citep{pang2002thumbs}.
Unlike general sentiment classification, aspect level sentiment classification aims at identifying the sentiment polarity of text regarding a specific aspect, and has received more attention~\citep{pontiki-etal-2014-semeval}.
While most works focus on sentence level and single domain sentiment classification, some attempts have been made on document level~\citep{chen-etal-2020-aspect} and cross-domain~\citep{ghosal-etal-2020-kingdom} sentiment classification. 
Early works on sentiment classification heavily relied on feature engineering~\citep{jiang2011target}.
Recent attempts~\citep{tang2016effective,huang-carley-2018-parameterized} leveraged the expressive power of various neural network models such as LSTM~\citep{hochreiter1997long}, CNN~\citep{lecun1998convolutional} or Memory Networks~\citep{sukhbaatar2015end}.
Very recently, more attempts have been made to leverage GNNs to better model syntactic and semantic meanings of text for the sentiment classification task.


\paragraph{Methodologies}
GNN-based sentiment classification approaches typically operate by first constructing a graph representation (e.g., dependency tree) of the text, and then applying a GNN to learn good text embeddings which will be used for predicting the sentiment polarity. 
The graph construction techniques and graph representation techniques developed for solving the sentiment classification task vary between different approaches.
In this section, we introduce and summarize some representative GNN-related techniques adopted in recent sentiment classification approaches.

\begin{itemize}
    \item\textbf{Graph Construction}
Most GNN-based approaches~\citep{zhang_aspect-based_2019,sun-etal-2019-aspect,huang_syntax-aware_2019,pouran_ben_veyseh_improving_2020,tang-etal-2020-dependency,wang_relational_2020} for sentence level sentiment classification used a dependency tree structure to represent the input text.
Besides using a dependency graph for capturing syntactic information, \citet{zhang-qian-2020-convolution} in addition constructed a global lexical graph to encode the corpus level word co-occurrence information, and further built a concept hierarchy on both the syntactic and lexical graphs.
\citet{ghosal-etal-2020-kingdom} constructed a subgraph from ConceptNet~\citep{speer2017conceptnet} using seed concepts extracted from text.
To capture the document-level sentiment preference information, \citet{chen-etal-2020-aspect} built a bipartite graph with edges connecting sentence nodes to the corresponding aspect nodes for capturing the intra-aspect consistency, and a graph with edges connecting sentence nodes within the same document for capturing the inter-aspect tendency.




\item\textbf{Graph Representation Learning}
Both the design of GNN models and quality of initial node embeddings are critical for the overall performance of GNN-based sentiment classification methods.
Common GNN models adopted in the sentiment classification task include GCN~\citep{zhang_aspect-based_2019,sun-etal-2019-aspect,pouran_ben_veyseh_improving_2020,zhang-qian-2020-convolution}, GAT~\citep{huang_syntax-aware_2019,chen-etal-2020-aspect} and Graph Transformer~\citep{tang-etal-2020-dependency}.
To handle multi-relational graphs, R-GCN~\citep{ghosal-etal-2020-kingdom} and R-GAT~\citep{wang_relational_2020} were also applied to perform relation-aware message passing over graphs.
Most approaches used GloVe+BiLSTM~\citep{sun-etal-2019-aspect,tang-etal-2020-dependency,wang_relational_2020,tang-etal-2020-dependency} or BERT~\citep{huang_syntax-aware_2019,pouran_ben_veyseh_improving_2020,chen-etal-2020-aspect,wang_relational_2020,tang-etal-2020-dependency} to initialize node embeddings.



\item\textbf{Special Techniques}
One common trick used in aspect level sentiment classification is to include position weights or embeddings~\citep{zhang_aspect-based_2019,zhang-qian-2020-convolution} to emphasize more on tokens closer to the aspect phase.





\end{itemize}

\paragraph{Benchmarks and Evaluation}
Common benchmarks for evaluating sentiment classification methods include
Twitter~\citep{dong2014adaptive},
SemEval sentiment analysis datasets~\citep{pontiki-etal-2014-semeval,pontiki2015semeval,pontiki2016semeval},
MAMS~\citep{jiang2019challenge}, and
Amazon-reviews~\citep{blitzer2007biographies}.
Accuracy and F1 are the most common evaluation metrics.

\subsection{Knowledge Graph}
Knowledge graph (KG), which represents the real world knowledge in a structured form, has attracted a lot of attention in academia and industry. KG can be denoted as a set of triples of the form $\left \langle subject, relation, object \right \rangle$.
There are three main tasks in term of KG, namely, knowledge graph embedding (KGE), knowledge graph completion (KGC), and Knowledge Graph Alignment (KGA). KGE aims to map entities and relations into low-dimensional vectors, which usually regarded as the sub-task in KGC and KGA. In this section, we will give a overview of the graph-based approaches to KGC and KGA.

\subsubsection{Knowledge Graph Completion}
\paragraph{Background and Motivation}
The purpose of KGC is to predict new triples on the basis of existing triples, so as to further extend KGs. KGC is usually considered as a link prediction task. Formally, the knowledge graph is represented by $\mathcal{G} = (\mathcal{V}, \mathcal{E}, \mathcal{R})$, in which entities $v_i \in \mathcal{V}$, edges $(v_s, r, v_o) \in \mathcal{E}$, and $r \in \mathcal{R}$ is a relation type. This task scores for new facts (i.e. triples like $\left \langle subject, relation, object \right \rangle$) to determine how likely those edges are to belong to $\mathcal{E}$.

\paragraph{Methodologies.} KGC can be solved with an encoder-decoder framework. To encode the local neighborhood information of an entity, the encoder can be chosen from a variety of GNNs such as GCN\citep{malaviya2020commonsense, shang2019end}, R-GCN \citep{schlichtkrull2018modeling, teru2020inductive} and Attention-based GNNs \citep{nathani2019learning, bansal2019a2n, zhang2020relational, wang-etal-2019-incorporating}. 
Then, the encoder maps each entity (subject entity and object entity) $v_i \in \mathcal{V}$ to a real-valued vector $e_i \in \mathbb{R}^d$. Relation can be represented as an embedding $e_r$ or a matrix $M_r$. Following the framework concluded by \citet{wang2019robust}, the GNN encoder in a multi-relational graph (such as KG) can be formulated as: 
\begin{equation}
\label{KGC_eq}
\begin{aligned}
    a_v^{(l)} &= AGGREGATE_l(h_{r,u}^{(l-1)}, \forall u \in \mathcal{N}_v^r) \\
    h_v^{(l)} &= COMBINE_l(h_{r0, v}^{(l-1)}, a_v^{(l)})
\end{aligned}
\end{equation}
where $h_{r,u}^{(l-1)}$ denotes the message passing from the neighbor node $u$ under relation $r$ at the l th layer. For example, RGCN \citep{schlichtkrull2018modeling} sets $h_{r,u}^{(l-1)} = W_r^{(l-1)}h_u^{(l-1)}$ and $AGGREGATE(\cdot)$ be mean pooling. 
Since the knowledge graph is very large, the update of the node representation Eq.\ref{KGC_eq} is efficiently implemented by using sparse matrix multiplications to avoid explicit summation over neighborhoods in practice.

The decoder is a knowledge graph embedding model and can be regarded as a scoring function. The most common decoders of knowledge graph completion includes translation-based models (TransE \citep{bordes2013translating}), tensor factorization based models (DistMult \citep{yang2014embedding}, ComplEx\citep{trouillon2016complex}) and neural network base models (ConvE \citep{dettmers2018conve}). In Table~\ref{tab:my_kgc} , we summarize these common scoring functions following \citet{ji2020survey}. $\text{Re}(\cdot)$ denotes the real part of  a vector, $*$ denotes convolution operator, $\omega$ denotes convolution filters and $g(\cdot)$ is a non-linear function. For example, RGCN uses DistMult as a scoring function, and DistMult can perform well on the standard link prediction benchmarks when used alone. In DistMult, every relation r is represented by a diagonal matrix $M_r \in \mathbb{R}^{d \times d}$ and a triple is scored as $f(s, r, o) = e_s^T M_r e_o$.

At last, the model is trained with negative sampling, which randomly corrupts either the subject or the object of each positive example. To optimize KGC models, cross-entropy loss \citep{schlichtkrull2018modeling, wang2019robust, zhang2020relational, malaviya2020commonsense} and margin-based loss \citep{teru2020inductive, nathani-etal-2019-learning} are common loss functions used for optimizing KGC models.

\begin{table}[]
    \centering
    \small
    \caption{KGC Scoring Function.}
    \resizebox{\textwidth}{!}{
    \begin{tabular}{cccc}
    \hline
Model & Ent. embed. & Rel. embed. & Scoring Function $f(s, r, o)$ \\ \hline
{DistMult} & \multirow{4}{*}{$e_s, e_o \in \mathbb{R}^d$} & \multirow{4}{*}{$M_r \in \mathbb{R}^{d \times d}$} & \multirow{4}{*}{$e_s^T M_r e_o$} \\ 
\citep{schlichtkrull2018modeling} & & & \\
\citep{wang2019robust} & & & \\
\citep{bansal2019a2n} & & & \\ \hline

ComplEx & \multirow{2}{*}{$e_s, e_o \in \mathbb{C}^d$} & \multirow{2}{*}{$e_r \in \mathbb{C}^d$} & \multirow{2}{*}{$\text{Re}(e_r, e_s, \Bar{e_t})=\text{Re}(\sum_{k=1}^K e_r e_s\Bar{e_t})$} \\
\citep{wang2019robust} & & & \\ \hline

ConvKB & \multirow{2}{*}{$e_s, e_o \in \mathbb{R}^d$} & \multirow{2}{*}{$e_r \in \mathbb{R}^d$} & \multirow{2}{*}{$\text{concat}(\sigma([e_s, e_r, e_o]*\omega))\cdot w$}\\
\citep{nathani-etal-2019-learning} & & & \\ \hline

ConvE & \multirow{2}{*}{$M_s \in \mathbb{R}^{d_w \times d_h}, e_o \in \mathbb{R}^d$} & \multirow{2}{*}{$M_r \in \mathbb{R}^{d_w \times d_h}$} &  \multirow{2}{*}{$\sigma(vec(\sigma([M_s; M_r]*\omega))W)e_o$} \\ 
\citep{wang-etal-2019-incorporating} & & & \\ \hline


Conv-TransE & \multirow{3}{*}{$e_s, e_o \in \mathbb{R}^d$} & \multirow{3}{*}{$e_r \in \mathbb{R}^d$} & \multirow{3}{*}{$g(vec(M(e_s, e_r))W)e_o$} \\
\citep{shang2019end} & & & \\
\citep{malaviya2020commonsense} & & & \\
\hline
    \end{tabular}}
    \label{tab:my_kgc}
\end{table}

\paragraph{Benchmarks and Evaluation} Common KGC benchmark datasets include FB15k-237 \citep{dettmers2018convolutional}, WN18RR \citep{toutanova2015representing}, NELL-995  \citep{xiong2017deeppath} and Kinship  \citep{lin2018multi}. Two commonly used evaluation metrics are \textit{mean reciprocal rank} (MRR) and \textit{Hits at $n$} (H@n), where $n$ is usually 1, 3, or 10. 

\subsubsection{Knowledge Graph Alignment} 

\paragraph{Background and Motivation}.
KGA aims at finding corresponding nodes or edges referring to the same entity or relationship in different knowledge graphs. KGA, such as cross-lingual knowledge graphs alignment, is useful for constructing more complete and compact KGs.
Let $G_1 = (\mathcal{V}_1, \mathcal{E}_1, \mathcal{R}_1)$ and $G_2 = (\mathcal{V}_2, \mathcal{E}_2, \mathcal{R}_2)$ be two different KGs, and $S = \{(v_{i_1}, v_{i_2})|v_{i_1} \in \mathcal{V}_1, v_{i_2} \in \mathcal{V}_2\}$ be a set of pre-aligned entity pairs between G1 and G2. The core task of KGA is entity or relation alignment, which is defined as finding new entity or relation alignments based on the existing ones.


\paragraph{Methodologies.} GNN-based KGA or entity alignment approaches mostly use GNN models to learn the representations of the entities and relations in different KGs. Then, entity/relation alignment can be performed by computing the distance between two entities/relations. GCN is widely used in \citep{wang2018cross, xu-etal-2019-cross-lingual, wu2019jointly}. To further capture the relation information existing in multi-relational KGs, \citet{wu2019relation} proposed a Relation-aware Dual-Graph Convolutional Network (RDGCN), which also applied a graph attention mechanism. Similarly, \citet{ye2019vectorized} also introduced relation information by proposing a vectorized relational graph convolutional network (VR-GCN). \citet{cao2019multi} proposed
a Multi-channel Graph Neural Network
model (MuGNN) containing a KG self-attention module and a cross-KG attention module to encode two KGs via multiple channels. GAT is another common model, which is applied in \citep{li2019semi, sun2020knowledge, wang-etal-2020-knowledge-graph}. Moreover, \citet{sun2020knowledge, wu2019relation, wu2019jointly} also introduced a gating mechanism to control the aggregation of neighboring information.

Entity/relation alignments are predicted by the distance between the entity/relation embeddings. The distance measuring functions are mainly based on L1 norm \citep{ye2019vectorized, wu2019relation, wang2018cross, wu2019jointly}, L2 norm \citep{cao2019multi, li2019semi, sun2020knowledge}, cosine similarity \citep{xu-etal-2019-cross-lingual}, and feed-forward neural network \citep{xu-etal-2019-cross-lingual, wang-etal-2020-knowledge-graph}.

\paragraph{Benchmarks and Evaluation.} Common KGA benchmarks datasets include~$DBP15K$ \citep{sun2017cross} and $DWY100K$ \citep{sun2018bootstrapping}. $DBP15K$ contains three cross-lingual datasets: $DBP_{ZH-EN}$ (Chinese to English), $DBP_{JA-EN}$ (Japanese to English), and $DBP_{FR-EN}$ (French to English). $DWY100K$ is composed of two large-scale cross-resource datasets: $DWY-WD$ (DBpedia to Wikidata) and $DWY-YG$ (DBpedia to YAGO3). Hits@N, which is calculated by measuring the proportion of correctly aligned entities/relations in the top N list, is used as evaluation metric to assess the performance of the models.

\subsection{Information Extraction}

\paragraph{Background and Motivation}
Information Extraction (IE) aims to extract entity pairs and their relationships of a given sentence or document. IE is a significant task because it contributes to the automatic knowledge graph construction from unstructured texts. With the success of deep neural networks, NN-based methods have been applied to information extraction. However, these methods often ignore the non-local and non-sequential context information of the input text \citep{qian2019graphie}. Furthermore, the prediction of overlapping relations, namely the relation prediction of pairs of entities sharing the same entities, cannot be solved properly \citep{fu2019graphrel}. To these ends, GNNs have been widely used to model the interaction between entities and relations in the text.

\paragraph{Methodologies}
Information extraction composed of two sub-tasks: named entity recognition (NER) and relation extraction (RE).
NER predicts a label for each word in a sentence, which is often regarded as a sequence tagging task \citep{qian2019graphie}. RE predicts a relation type for every pair of entities in the text. When the entities are annotated in the input text, the IE task degrades into an RE task \citep{sahu-etal-2019-inter, christopoulou-etal-2019-connecting, guo-etal-2019-attention, zhu-etal-2019-graph, zhang-etal-2019-long, zhang-etal-2018-graph, vashishth-etal-2018-reside, zeng-etal-2020-double, song-etal-2018-n}.
GNN-based IE approaches typically operate via a pipeline approach. First, a text graph is constructed. Then, the entities are recognized and  the relationships between entity pairs are predicted. Very recently, researchers starts to jointly learn the NER and RE to take advantage of the interaction between these two sub-tasks \citep{fu2019graphrel, luan2019general, sun-etal-2019-joint}. Followings are the introduction of different GNN-based techniques.

\begin{itemize}
    \item\textbf{Graph Construction} Most GNN-based information extraction methods design specific rules to construct static graphs. Because the input of IE task is usually a document containing multiple sentences, the nodes in the constructed graph can be words, entity spans and sentences and the corresponding edges are word-level edges, span-level edges and sentence-level edges. These nodes can be connected by syntactic dependency edges \citep{fu2019graphrel, guo-etal-2019-attention, zhang-etal-2018-graph, vashishth-etal-2018-reside, song-etal-2018-n, sahu-etal-2019-inter}, co-reference edges \citep{luan2019general, zeng-etal-2020-double, sahu-etal-2019-inter}, re-occurrence edges \citep{qian2019graphie}, co-occurrence edges \citep{christopoulou-etal-2019-connecting, zeng-etal-2020-double}, adjacent word edges \citep{qian2019graphie, luan2019general, sahu-etal-2019-inter} and adjacent sentence edge \citep{sahu-etal-2019-inter}. Recently, dynamic graph construction has also been successfully applied in IE tasks. \citep{luan2019general} proposed a general IE framework using dynamically constructed span graphs, which selected the most confident entity spans from the input document and linked these span nodes with co-references and confidence-weighted relation types. \citep{sun-etal-2019-joint} first constructs a static entity-relation bipartite graph and then investigates the dynamic graph for pruning redundant edges.

\item\textbf{Graph Representation Learning}
To better capture non-local information of the input document, a variety of GNN models are applied in the NER task and the RE task. In addition, joint learning is a critical technique to reduce error propagation along the pipeline.
For the name entity recognition task, common GNN models such as GCN \citep{qian2019graphie, luo-zhao-2020-bipartite} are applied. GCN is the most common GNN models used in the relation extraction task \citep{zhang-etal-2019-long, zhang-etal-2018-graph, vashishth-etal-2018-reside, zeng-etal-2020-double}. To learn edge type-specific representations, \citet{sahu-etal-2019-inter} introduces a labelled edge GCN to keep separate parameters for each edge type. Inspired by the graph attention mechanism, \citet{guo-etal-2019-attention} proposes attention guided GCN to prune the irrelevant information from the dependency trees.
Recently, many joint learning models have been proposed to relieve the error propagation in the pipeline IE systems and leverage the interaction between the NER task and the RE task. \citet{fu2019graphrel} proposes a GraphRel model containing 2 phases prediction of the entities and relations. \citet{luan2019general} introduces a general framework to couple multiple information extraction sub-tasks by sharing entity span representations which are refined using contextualized information from relations and co-references. \citet{sun-etal-2019-joint} develops a paradigm that first detected entity spans, and then performed a joint inference on entity types and relation types.

\end{itemize}

\paragraph{Benchmarks and Evaluation.}
Common IE benchmark datasets contain NYT \citep{riedel2010modeling}, WebNLG \citep{gardent2017creating}, ACE2004, ACE2005, SciERC\citep{luan2018multi}, TACRED \citep{zhang2017position} and etc. Precision, recall and F1 are the most common evaluation metrics for IE.

\subsection{Semantic and Syntactic Parsing}
In this section, we mainly discuss applications of GNN for parsing, including syntax related and semantics related parsing. For syntax related parsing, GNN has been employed in tasks of dependency parsing\citep{ji-etal-2019-graph}\citep{do-rehbein-2020-neural} and constituency parsing\citep{yang2020strongly}. For semantics related parsing, we will briefly introduce semantic parsing and AMR (Abstract Meaning Representation) parsing.

\subsubsection{Syntax Related Parsing} 

\paragraph{Background and motivation}
The tasks related to syntax are mainly dependency parsing and constituency parsing. Both of them aim to generate a tree with syntactic structure from natural language sentences, conforming to predefined formal grammar rules. Dependency parsing focuses on the dependency relationship between words in sentence. Constituency parsing focuses on the compositional relationship between different components in a sentence. Traditional approaches can be divided into two directions: transition-based and graph-based. Transition-based methods\citep{andor-etal-2016-globally}\citep{ma-etal-2018-stack} usually formalize this problem as a series of decisions on how to combine different words into a syntactic structure. Graph-based methods\citep{kiperwasser-goldberg-2016-simple}\citep{dozat2016deep}\citep{ji-etal-2019-graph} firstly score all word pairs in a sentence on the possibility of holding valid dependency relationship, and then exploit decoders to generate the parse trees. 

\paragraph{Methodologies}
Here, we mainly focus on graph-based parsers where graph neural network plays the role of extracting high-order neighbor features.

\begin{itemize}
    \item \textbf{Dependency parsing}.  
In graph-based parsers, we take each word as a node in a graph and the key task is to learn a low-dimensional node representation with a neural encoder. 
To incorporate more dependency structure information, \citet{ji-etal-2019-graph} proposes to employ GNN as a encoder to incorporate high-order information. Their encoder contains both GNN and Bi-LSTM, where the GNN accepts all node embeddings from Bi-LSTM and take them as node embeddings in a complete graph. The constructed graphs are dynamic graphs where edge weight can change consistently during training. There are two kinds of loss functions: 1) the first one considers both tree structure and dependency relation labels; 2) the second one are applied after each GNN layer where only tree structure is considered. 
Other than the generation of dependency parsing trees, some other works focus on how to do reranking among different candidates to choose a best parsing tree. \citet{do-rehbein-2020-neural} demonstrate that GNN can also work well as a encoder for dependency parsing trees in a neural reranking model.

\item\textbf{Constituency parsing}. Most approaches for constituency parsing are transition-based~\citep{sagae-lavie-2005-classifier, dyer-etal-2016-recurrent, liu-zhang-2017-order, yang2020strongly} which generate the constituency parsing tree by executing an action sequences. \citet{yang2020strongly} proposes to use GNN to encode the partial tree in the decoding process which can generate one token per step. Other methods usually generate the final parsing tree by combining different sub-trees in a shift-reduce way. The authors believe that this strongly incremental way is more closer to the way of human thinking.

\end{itemize}

\paragraph{Benchmark and Evaluation}
For syntactic parsing, two becnmark datasets are commonly used, namely, PTB 3.0\citep{taylor2003penn} and UD 2.2\citep{11234/1-2837}. As for evaluation, Accuracy, including exact match accuracy and execution accuracy, and Smatch score\citep{cai-knight-2013-smatch} are commanly used.

\subsubsection{Semantics Related Parsing}

\paragraph{Background and Motivation}
For semantics related tasks, we will introduce two popular applications: SQL parsing and AMR parsing. Semantic parsing aims to generate machine-interpretable representations from natural language, like SQL queries. AMR parsing is another young research field. AMR is represented as a rooted labeled directed acyclic graph form, and the goal of AMR parsing aims to provide sentence-level semantic representations. It is widely used in many NLP tasks like text summarization, machine translation and question answering\citep{zhou-etal-2020-amr}.  

\paragraph{Methodologies}
Here, we provide a summary of the techniques for two typical semantic related parsing tasks, namely, SQL parsing and AMR parsing.

\begin{itemize}

\item\textbf{SQL parsing}. The main purpose of SQL parsing is to convert natural language into SQL queries that can be successfully executed.
Most of the traditional methods~\citep{jia-liang-2016-data, alvarez2016tree, dong-lapata-2016-language} are sequential encoder based, which however, lost some other useful information at the source side, such as syntax information and DB schema information. Thus, many GNN-based models are proposed.
For syntactic information, \citet{li-etal-2020-graph-tree, xu-etal-2018-exploiting} use external parser to perform syntactic parsing (i.e., constituency parsing and dependency parsing) on the raw sentence. Then they exploit the syntactic parsing tree instead of the source sentence as input, and use GNN to learn the syntactic structure and dependent information in this "tree" graph. It has been proved experimentally that the additional syntactic information is helpful for semantic parsing tasks.
SQL parsing problem becomes more complex if the DB schema of the training and testing set are different~\citep{yu-etal-2018-spider}. To this end, some works propose to model these schema information to achieve better results. For example, \citet{bogin-etal-2019-representing} takes the DB schema as a graph and use GGNN to learn the node representation. Then they incorporate schema information on both the encoder and decoder to generate the final results. \citet{bogin-etal-2019-global} employs GNN to globally select the overall structure of the output query which could decrease the ambiguity of DB constants choice.

After the SQL queries are generated, reranking can be utilized to further improve the performance. Reranking the candidates predicted by the model is helpful to reduce the likelihood of picking some sub-optimal results. SQL queries are structured and it is a reasonable way to use GNN to encode the SQL queries in the reranking model. For example, \citet{do-rehbein-2020-neural} employs graph-based transformer to rearrange the results generated by the neural semantic parser and achieved good results.

\item\textbf{AMR parsing}.
Similar to \citep{li-etal-2020-graph-tree}\citep{xu-etal-2018-exploiting}, syntactic information, especially dependency relation information, are also employed in AMR parsing. \citet{zhou-etal-2020-amr} considers both the dependency syntactic graph and the latent probabilistic graph. Specifically, by learning a vector representation for the two graph structures and then fusing them together, their model leverages the structure information in the source side and achieve better performance compared to seq-to-graph-like models.

\end{itemize}

\paragraph{Benchmark and Evaluation} 
For SQL parsing, three benchmark datasets are commanly used, including ATIS\citep{dahl-etal-1994-expanding}, GEO\citep{luke2005ze}, WikiSQL\citep{zhong2017seq2sql}, SPIDER\citep{yu-etal-2018-spider}. For AMR parsing, AMR annotation release\citep{knight2014abstract, knight2017abstract} is a well-recognized dataset. For evaluation metrics, accuracy, including exact match accuracy and execution accuracy, as well as Smatch score\citep{cai-knight-2013-smatch}.) are commonly used.

\subsection{Reasoning}
Reasoning is a significant research direction for NLP. In recent years, GNN begins to play an important role in NLP reasoning tasks, such as math word problem solving~\citep{li-etal-2020-graph-tree, zhang-etal-2020-graph-tree}, natural language inference~\citep{Kapanipathi2020InfusingKI, Wang_Kapanipathi_Musa_Yu_Talamadupula_Abdelaziz_Chang_Fokoue_Makni_Mattei_Witbrock_2019}, common sense reasoning~\citep{lin-etal-2019-kagnet, zhou2018commonsense} and so on. In this subsection, we will give a brief introduction for the three tasks and how graph neural networks are employed in these methods.

\subsubsection{Math word problem solving}
\paragraph{Background and Motivation}
Math word problem solving aims to infer reasonable equations from given natural language problem descriptions. It is important for exploring automatic solutions to mathematical problems and improving the reasoning ability of neural networks. Most of the traditional methods are based on the seq2seq~\citep{wang2017deep} framework to generate the corresponding equation directly from the source sentence in an end-to-end manner. This kind of methods ignore some important information in natural sentences, such as 1) the relationship information between different mathematical elements (numbers) in the question, 2) the syntax information in the question sentence, 3) external knowledge, and so on. Thus, GNN-based models are proposed as a very good way to incorporate this information.

\paragraph{Methodologies}
\citet{li-etal-2020-graph-tree} is the first to introduce GNN into math word problem solving. Graph2tree considers both the input and output structure information. At the input side, GNN is used to encode the input syntactic tree. After all the input nodes embedding are generated, on the output side, considering the hierarchy of the equation, a BFS-based tree decoder is used to generate the final equation result in a coarse-to-fine way.
\citet{zhang-etal-2020-graph-tree} is another MWP automatic solving model that uses graph data structure to model 1) the relationship between the numbers in the problem, and 2) the relationship between different numbers with their corresponding descriptors. 
In addition, some works introduce the external knowledge information in another way. For example, \citet{wu-etal-2020-knowledge} first connects the entities in the problem description into graphs based on external global knowledge information, and then uses GAT as encoder. This method can enhance the ability of modeling the relationship between the entities in the problem, and has obtained good results.

\paragraph{Benchmarks and Evaluation}
For math word problem, three benchmark datasets are commonly used, including MAWPS~\citep{koncel2016mawps}, MATH23K~\citep{wang2017deep}, and MATHQA~\citep{amini2019mathqa}.

\subsubsection{Natural language inference}

\paragraph{Background and Motivation}
Natural language inference (NLI) is another fundamental reasoning task. This task aims to predict the relationship between premise and hypothesis, and is often formalized as a three-classification problem (contradict, entails, neutral). 

\paragraph{Methodologies}
Traditional methods are mostly based on neural encoder with attention, and most of them are RNN models\citep{chen2017neural}. Considering the rich information contained in the external knowledge base, some works try to use external information to improve the accuracy of the model. For example, \citet{Wang_Kapanipathi_Musa_Yu_Talamadupula_Abdelaziz_Chang_Fokoue_Makni_Mattei_Witbrock_2019} uses graph-based attention model to incorporate the information from introduced external knowledge source. Their experiments demonstrate that adding the learned knowledge graph representation to the classifier help to obtain good results. Considering the introduced graph can have noisy information, \citet{Kapanipathi2020InfusingKI} employs a encoder with a subgraph filtering module using Personalized PageRank before a GCN layer where the filtering module can help to select context relevant sub-graphs from introduced knowledge graph to reduce noisy information.

\paragraph{Benchmarks and Evaluation}
For NLI task, three benchmark datasets are commonly used, including SNLI\citep{bowman2015large}, MultiNLI\citep{Williams_2018}, and SciTail\citep{khot2018scitail}.

\subsubsection{Commonsense reasoning}

\paragraph{Background and Motivation}
Commonsense reasoning helps neural models incorporate the "common sense" or world knowledge during inference. Take the commonsense QA as example, we aim to obtain a neural model tended to generate the answer which is more consistent with commonsense from multiple answers that all logically fit the requirements. In fact, large-scale pre-trained models such as GPT-2\citep{radford2019language}, BERT\citep{devlin-etal-2019-bert} with simple fine-tuning can achieve very good results. However, some external knowledge sources can help the model to better characterize the question and the concepts in the answer, which will definitely help the overall performance.

\paragraph{Methodologies}
\citet{lin-etal-2019-kagnet} introduces graph neural networks to the common sense reasoning task. The model first retrieves the concepts in the questions and options into an external knowledge base to obtain a schema graph, and then uses GCN to incorporate information from this retrieved graph to learned features. The learnt features would be fed to a simple score module for each QA pair. Experiments on large benchmarks dataset, e.g., CommonsenseQA~\citep{talmor2019commonsenseqa}, demonstrate the effectiveness of the external knowledge base introduced by GNN.

\paragraph{Benchmarks and Evaluation}
We introduce some benchmark datasets for commonsense reasoning here: CommonsenseQA~\citep{talmor2019commonsenseqa}; Event2Mind\citep{rashkin2018event2mind}; SWAG\citep{zellers2018swag}; Winograd Schema Challenge\citep{levesque2012winograd}; ReCoRD\citep{zhang2018record}.


\subsection{Semantic Role Labelling}

\paragraph{Background and Motivation}
The problem of semantic role labeling (SRL) aims to recover the predicate-argument structure of a sentence, namely, to determine essentially “who did what to whom”, “when”, and “where. More formally, for every predicate, the SRL model must identify all argument spans and label them with their semantic roles. Such high-level structures can be used as semantic information for supporting a variety of downstream tasks, including dialog systems, machine reading and translation~\citep{shen2007using,liu2010semantic,gao2011corpus}. Recent SRL works can mostly be divided into two categories, i.e., syntax-aware~\citep{xia2020semantic,marcheggiani-titov-2020-graph} and syntax-agnostic~\citep{he2017deep,he2018jointly} approaches according to whether incorporating syntactic knowledge or not. Most syntax-agnostic works employ deep BiLSTM or self-attention encoder to encode the contextual
information of natural sentences, with various kinds of scorers to predict the probabilities of BIO-based semantic roles~\citep{he2017deep} or predicate-argument-role tuples~\citep{he2018jointly}. Motivated by the strong interplay between syntax and semantics, researchers explore various approaches to integrate syntactic knowledge into syntax-agnostic models considering that the semantic representations are closely related
to syntactic ones. For example, one can observe that many arcs in the syntactic dependency graph are mirrored in the semantic dependency graph. Given these similarities and the availability
of accurate syntactic parser for many languages,
it seems natural to exploit syntactic information
when predicting semantics. 

However, the last
generation of SRL models powered by deep learning models put syntax aside in favor of neural sequence models, namely LSTMs~\citep{zhou_neural_2020,marcheggiani2017simple} due to the challenges that (1) it is difficult to effectively incorporate syntactic information into neural SRL models, due to the sophisticated tree structure of syntactic relation; and (2) the syntactic parsers are unreliable on account of the risk of erroneous syntactic input, which may lead to error propagation and an unsatisfactory SRL performance. Given this situation, GNNs are emerging as powerful tools to capture and incorporate the syntax patterns into deep neural network-based SRL models. The nature property of GNN in capturing the complex relationship patterns in the structured data makes it a good fit for modeling syntactic dependency and constituency structures of sentences.

\paragraph{Methodologies}
The problem solved by the GNN-based SRL models can be divided into two categories. One is about argument prediction given the predicates in a sentence~\citep{marcheggiani-titov-2020-graph,marcheggiani2017encoding, li2018unified}. Formally, SRL can be cast as a sequence labeling problem where given an input sentence, and the position of the predicate in the sentence, the goal is to predict a BIO sequence of semantic roles for the words in sentences; Another is about end-to-end semantic role triple extraction which aims to  detect all the possible predicates and their corresponding arguments in one shot~\citep{fei2020cross,xia2020semantic}. Technically, given a sentence, the SRL model predicts a set of labeled predicate-argument-role triplets, while each triple contains a possible predicate token and two candidate tokens. Both of the above mentioned problems can be solved based on the GNN-based SRL models, which consists of two parts, namely, graph construction and graph representation learning.

\begin{itemize}
    \item\textbf{Graph Construction}.
The graphs are constructed based on the syntax information, which can be extracted from two sources, one is syntactic dependency information and another is syntactic constituents information. Most of the existing GNN-SRL models~\citep{li2018unified,fei2020cross,marcheggiani2017encoding,zhang2020syntax,xia2020semantic} have relied on syntactic dependency representations. In these methods, information from dependency trees are injected into word representations using GNN or self-attention mechanisms. Recently, Marcheggiani et al~\citep{marcheggiani-titov-2020-graph} incorporated the constituency syntax into SRL models by conducting the message passing on a graph where nodes represent constituents. 
Based on the syntax information, the graphs constructed in the current SRL models are divided into three main categories: (1) directed homogeneous graphs; (2) heterogeneous graphs; and (3) probability weighted graphs. Most of the works~\citep{marcheggiani2017encoding,li2018unified,fei2020cross} represent the syntax information as a directed homogeneous graph where all the nodes are input word tokens and directed with dependent edges. Other work~\citep{xia2020semantic} enhances SRL with heterogeneous syntactic knowledge by combining various syntactic treebanks that follow different annotation guidelines and domains. Liu et al.~\citep{liu2019contextualized} also construct a heterogeneous syntactic graph by incorporating several types of edges, including lexical relationships, syntactic dependency, co-occurrence relationships. Some work~\citep{zhang2020syntax} utilizes the probability matrix of all dependency arcs for constructing an edge-weighted directed graph to eliminate the influences of the error from the parsing results.

\item\textbf{Graph Representation Learning}. 
As described in Section 6, various GNN models can be utilized for graph representation learning. Here, we introduce the different roles that GNNs play in different SRL models. In most of the works~\citep{zhang2020syntax, liu2019contextualized,marcheggiani2017encoding, marcheggiani-titov-2020-graph}, GNN is utilized as an encoder to learn the final representations of words which follows a typical word embedding layer, such as BiLSTM. While in some works~\citep{xia2020semantic, fei2020cross},
GNN is utilized to extract the initial words' embedding, which are regarded as inputs of the encoder. For example, Xia et al.~\citep{xia2020semantic} combines the syntax embedding extracted from GNN with the word embedding and character embedding as the input. Fei~\citep{fei2020cross} utilizes GNN to refine the initial word embedding which consists of word representation and part-of-speech (POS) tags, and then input the refined word embedding into the BiLSTM encoder.

\end{itemize}

\paragraph{Benchmarks and Evaluation}
There are two main benchmark datasets for the evaluation in the domain of SRL: (1) CoNLL dataset concerns the recognition of semantic roles for the English language, based on PropBank predicate-argument structures. Given a sentence, the task consists of analyzing the propositions expressed by some target verbs of the sentence. In particular, for each target verb all the constituents in the sentence which fill a semantic role of the verb have to be recognized. (2) Chinese Proposition Bank 1.0 (CPB1.0) which creates a corpus of text annotated with information about basic semantic propositions (i.e., predicate-argument relations).
The typical evaluation metrics in SLR task are about metrics for classification problem, such as precision, recall and F1 of the correctly predicted arguments.

\subsection{Related Libraries and Codes}
Open-source implementations facilitate the research works of baseline experiments in graph neural networks for NLP. Besides various paper codes were released individually, there is a recently released library called \textit{Graph4NLP}~\footnote{The codes and details of \textit{Graph4NLP} library are provided at~\url{https://github.com/graph4ai/graph4nlp}.}, which is an easy-to-use library for R\&D at the intersection of Deep Learning on Graphs and Natural Language Processing. It provides both full implementations of state-of-the-art models mentioned above for several NLP applications including text classification, semantic parsing, machine translation, KG completion, and natural language generation. \textit{Graph4NLP} also provides flexible interfaces to build customized models for researchers and developers with whole-pipeline support. Built upon highly-optimized runtime libraries including \textit{DGL} and \textit{Pytorch}, Graph4NLP has both high running efficiency and great extensibility. The architecture of \textit{Graph4NLP} consists of four different layers: 1) Data Layer, 2) Module Layer, 3) Model Layer, and 4) Application Layer. There are also some other related GNN-based libraries. 
Noticeably, \citet{fey2019fast} published a geometric learning library in \textit{PyTorch} named \textit{PyTorch Geometric}, which implements many GNNs. The \textit{Deep Graph Library (DGL)}~\citep{wang2019deep} was released which provides a fast implementation of many GNNs on top of popular deep learning platforms such as \textit{PyTorch} and \textit{MXNet}. The \textit{Dive into Graphs} \citep{liu2021dig} was released recently as a research-oriented library that integrates unified and extensible implementations of common graph deep learning algorithms for several advanced tasks.

\section{General Challenges and Future Directions}
\label{sec:General Challenges and Future Directions}

In this chapter, we will discuss various general challenges of GNNs for NLP and pinpoint the future research directions. We believe putting more research efforts in these directions will further unleash the great potential of GNNs in the NLP field, and result in fruitful research outcomes.



\subsection{Dynamic Graph Construction}
As we see in~\cref{sec:Applications}, 
many NLP problems can be tackled from a graph perspective, and GNNs are naturally applicable to and good at handling graph-structured data. 
Thus, the graph construction process plays an important role in the overall model performance. However, constructing a high quality and task-specific graph requires a good amount of domain expertise and human effort.
Moreover, graph construction in NLP is often more art than science, informed solely by insight of the practitioner, and involves many trials and errors. Even though a few of existing works already explored dynamic graph construction, most GNN applications in NLP still heavily relied on domain expertise for static graph construction.

The exploration of dynamic graph construction for NLP, is still at its early stage and faces several challenges: first of all, most works on dynamic graph construction focused only on homogeneous graph construction~\citep{chen2020reinforcement,chen2020graphflow,chen2020iterative,liu2021retrieval,liu2019contextualized}, and dynamic graph construction for heterogeneous graphs~\citep{yun2019graph,zhao2021heterogeneous} is much less explored especially for NLP tasks. 
Compared to homogeneous graphs, heterogeneous graphs are capable of carrying on richer information on node types and edge types, and occur frequently in many NLP problems. Dynamic graph construction for heterogeneous graphs is also supposed to be more challenging because more types of information (e.g., node types, edge types) are expected to be learned from data.

Second, most existing dynamic graph construction techniques rely on some form of pair-wise node similarity computation whose time complexity is at least $O(n^2)$ where $n$ is the number of graph nodes. This results in scalability issues when scaling to large graphs such as KGs.
Recently, a scalable graph structure learning approach with linear time and memory complexity (in terms of the number of nodes) was proposed by adopting the anchor-based approximation technique to avoid explicit pair-wise node similarity computation~\citep{chen2020iterative}. 

Finally, various efficient transformers~\citep{tsai2019transformer,katharopoulos2020transformers,choromanski2021rethinking,peng2021random,shen2021efficient,DBLP:journals/corr/abs-2006-04768} were also developed which could inspire the research in scalable dynamic graph construction considering their close connections;
(3) As observed in some previous works, dynamic graph construction does not clearly outperform static graph construction in some NLP applications. There is still room for improvement for dynamic graph construction techniques in terms of downstream task performance.
Other interesting future directions in this line include dynamically learning edge directions for graphs, and combining static and dynamic graph construction for better performance.

\subsection{GNNs vs Transformers for NLP}
While GNNs have achieved promising results in a large variety of NLP fields, Transformers have received much more attentions due to excellent performance in many NLP applications. However, as we pointed out in the previous section, Transformers are type of special GNNs, which operated on a fully connected dynamic graph constructed by employing self-attention mechanism. Since both of them have their clear advantages over each other, there are several interesting direction worth exploiting here.

\paragraph{Combining GNNs with Transformers for NLP} The most beautiful thing about Transformers is its simple use of its elegant model architecture directly on original text sequence, which separates the graph data modeling inside transformer model from the inputs (and from the end user). However, the downside of this model choice is that the Transformers cannot directly operate on more complex data like graph-structured data directly. In contrast, GNNs are more generic model architecture directly operating on graph data, which however are needed created by the end user with either domain specific knowledge or other graph modeling techniques. Currently, Graph Transformers are most popular models that adapt structure-aware self-attention mechanism to combine the advantages of Transformers and GNNs. However, these approaches are purely replying on attention mechanism to utilize the original graph topology information, which may not the best way to explore the original graph input information, especially when graph inputs are multi-relational and heterogeneous graphs. 

\paragraph{Pre-training GNNs for NLP} One of the most important trends in NLP over the past few years is to develop large-scale pre-trained models \citep{devlin2018bert, brown2020language} most of which are based on Transformer architectures. Recently, there are also many research efforts on pre-training GNNs on graphs~\citep{hu2019strategies, qiu2020gcc} using self-supervised learning methods. However, there are very few attempts to pre-train GNNs for NLP~\citep{DBLP:conf/emnlp/HeZXJLYX20, DBLP:conf/coling/SunSQGHHZ20, DBLP:conf/emnlp/ChenSYW20}, which may exploit more types of data sources than Transformers since GNNs can take both graph structured data (e.g., from KGs) and unstructured data (e.g., from free-form text).

\subsection{Graph-to-Graph for NLP}

Since the natural language or information knowledge can be naturally formalized as graphs with a set of nodes and their relationships, many generation tasks in the domain of NLP can be formalized as a graph transformation problem, which can further be solved by the graph-to-graph models. In this way, the semantic structural information of both the input and output sentences can be fully utilized and captured. For example, the graph-to-graph transformation for AMR parsing is promising and yet unexplored.
Because AMR are naturally structured objects (e.g. tree structures), currently, many semantic AMR parsing methods are based on deep graph generative models. These methods~\citep{flanigan2014discriminative, zhang-etal-2019-amr,cai-lam-2020-amr} represent the semantics of a sentence as a semantic graph (i.e., a sub-graph of a knowledge base) and treat semantic parsing as a semantic graph matching/generation process. These end-to-end deep graph generation techniques for semantic parsing show powerful ability in automatically capturing semantic information. However, current graph-based semantic parsing models only represent either the output AMR or the input sentence as a graph, without jointly utilize the interactive relationship in both input and output, and thus can not consider the complex relationship among the topology pattern of AMR logits and input dependency/constituency parsing.

While utilizing the graph-to-graph model for NLP tasks, there are several general challenges that deserve to be explored and solved in this domain: (1) Difficulty in building an end-to-end graph transformation framework which can integrate several sub-tasks jointly. For example, it is important to jointly tackle name entity recognition and relation extraction jointly in the information extraction; (2) The different concepts of input and output graphs. In NLP tasks, we usually formalize the dependency tree as the input graph and the output graph is usually has different concept from the input graph. Thus, both the node set, and topology of the input and output graphs are different. For example, in AMR parsing, the nodes in output graph are AMR logits, while the nodes in input graphs are word tokens; (3) Difficulty in addressing the graph sparsity issue in NLP tasks. For example, AMR parsing is a much harder task in that the target vocabulary size is much larger, while the size of dataset is much smaller; (4) Unpaired graph-to-graph transformation. The annotations are relatively expensive to produce and thus corpora have on the order of tens of thousands of sentences. Utilizing the unpaired sample pairs is also important yet challenging.

\subsection{Knowledge Graph in NLP}
Knowledge graph has become an important component in many NLP tasks, such as question answering, natural language generation, knowledge graph completion and alignment. It can be either incorporated as an auxiliary information besides the input text to provide more knowledge (e.g., KG augmentation), or as an target object to be learnt or extracted from (e.g., KGE and KGC). 

\paragraph{Knowledge Graph Augmentation}
For many NLP tasks such as QA and NLG, it is increasingly common to find that the input data not only contain Query text or source text, but also incorporate KG as auxiliary information for additional knowledge. There are several challenges with KG-augmented tasks: (1) There may exist ambiguity when aligning entities in the input text to entities in KG in some text-to-text generation tasks. For example, a person name appearing in the input text can correspond to multiple entries in KG; (2) Since the scale of KG 
is often large, it takes considerable effort to extract useful information from the KG and build knowledge subgraphs which can be directly used by every sample as an auxiliary input. Generally, detailed and task-specific rules need to be designed for KG node selection; (3) Entity alignment and knowledge subgraph construction may involve inevitable errors propagating to the downstream tasks. To make better use of KGs, the techniques for pre-processing KG, such as entity alignment and related entity/node selection, need to be further explored and improved.

\paragraph{Knowledge Graph Embedding and Completion}
GNN-based KGE and KGC approaches consider incorporating the neighborhood of a head entity or a tail entity. There is a trade-off between \textit{all triples training on the same large KG \citep{shang2019end}} and \textit{each triple training on a separate knowledge subgraph constructed from the original KG \citep{teru2020inductive, xie2020reinceptione}}. The former one provides more computational efficiency while the latter one has more powerful model expressiveness.
Future research will focus on jointly reasoning text and KG by applying such methods and paying attention to the entities mentioned in the text \citep{bansal2019a2n}. Logic rules play an important role to determine whether a triple is valid or not, which may be useful for KGE and KGC \citep{xie2020reinceptione}.

\paragraph{Knowledge Graph Alignment}
Most of the existing GNN-based KG alignment models also face three critical challenges to be further explored: (1) Different KGs usually have heterogeneous schemas, and may mislead the representation learning , which makes it is difficult to integrate knowledge from different KGs \citep{cao2019multi, wu2019jointly}; (2) The data in KG is usually incomplete \citep{sun2020knowledge} which needs pre-processing; (3) The seed alignments are limited \citep{li2019semi}. How to iteratively discover new entity alignments in the GNN-based framework is a future direction \citep{wang2018cross, li2019semi}.

\subsection{Multi-relational Graph Neural Networks}
Multi-relational graphs, which adopt unified nodes but relation-specific edges, are widely observed and explored in many NLP tasks. As we discussed in Section~\ref{sec:graph_representation_learning_heterogeneous}, most multi-relational GNNs, which are capable of exploiting multi-relational graphs, are extended from conventional homogeneous GNNs. Technically, most of them either apply relation-specific parameters during neighbor aggregation or split the heterogeneous graphs to homogeneous sub-graphs~\citep{schlichtkrull2018modeling,beck-etal-2018-graph}. Although impressive progresses have been made, there is still a challenge in handling over-parameterization problem due to the diverse relations existing in the graph. Although several tricks such as parameter-sharing (e.g, see Directed-GCN~\citep{marcheggiani2017encoding}) and matrix-decomposition (e.g., see R-GCN~\citep{schlichtkrull2018modeling}) are widely used to enhance the models' generalization ability to address this issue, they still have limitations, such as resulting in the potential loss of the models' expression ability. There is a hard trade-off between the over-parameterization and powerful model expression ability.

It is worth noting that various graph transformers have been introduced to exploit the multi-relational graphs~\citep{yao2020heterogeneous,wang2020amr}. However, the challenge exists in how to fully take advantage of the strong inductive bias (i.e., the graph topology) of the graphs by the transformers which are naturally free of that. 
Currently, most of them simply regard the self-attention's map as a fully-directed graph. On top of that, researchers either apply sparsing mechanisms~\citep{yao2020heterogeneous} or allow remote connection~\citep{shaw-etal-2018-self,cai2020graph} according to the given graph. How to develop an effective and general architecture for multi-relational graphs (or heterogeneous graphs) needs further exploration.

\section{Conclusions}
\label{sec:Conclusions}

In this article, we conduct a comprehensive overview of various graph neural networks evolving various NLP problems. Specifically, we first provide the preliminary knowledge of typical GNN models including graph filters and graph poolings. Then we propose a new taxonomy that systematically organizes GNNs for NLP approaches along three dimensions, namely, graph construction, graph representation learning, and the overall encoder-decoder models. Given these specific techniques at each stage of the NLP application pipelines, we discuss a wide range of NLP applications from the perspective of graph construction, graph representation learning, and special techniques. Finally, the general challenges and future directions in this line are provided to further unleash the great potential of GNNs in the NLP field.

\bibliographystyle{ACM-Reference-Format}
\bibliography{graph4nlp}

\end{document}